%% file: main.tex
\pdfoutput=1
\documentclass[11pt]{article}
\usepackage[preprint]{acl}
\input{preamble}

\input{tables/results_macros}

\title{\textsc{UltraDiscovery}: Abductive Exploration in an Interconnected, Epistemically Open Universe}

\author{%
  Weihan Li\textsuperscript{\normalfont 1}\begin{NoHyper}\thanks{Equal contribution.}\end{NoHyper} \quad
  Tianshi Zheng\textsuperscript{\normalfont 2}\begin{NoHyper}\footnotemark[1]\end{NoHyper} \quad
  Yangqiu Song\textsuperscript{\normalfont 2} \quad
  Ginny Y. Wong\textsuperscript{\normalfont 3} \quad
  Simon See\textsuperscript{\normalfont 3}\\[4pt]
  \normalfont\normalsize\textsuperscript{1}Department of Mechanical Engineering, The University of Tokyo, Tokyo, Japan\\
  \normalfont\normalsize\textsuperscript{2}Department of Computer Science and Engineering, HKUST, Hong Kong SAR, China\\
  \normalfont\normalsize\textsuperscript{3}NVIDIA AI Technology Center (NVAITC), NVIDIA, Santa Clara, USA\\[3pt]
  \normalfont\normalsize\texttt{liweihan1107@g.ecc.u-tokyo.ac.jp}}

\begin{document}

\maketitle

\input{sections/0_abstract}
\input{sections/1_introduction}

\input{sections/2_related}
\input{sections/3_benchmark}
\input{sections/4_experimental_setup}
\input{sections/5_results}
\input{sections/6_discussion}

\bibliography{references}

\clearpage
\appendix
\input{sections/app_a_world}
\input{sections/app_b_scoring}
\input{sections/app_c_validity}
\input{sections/app_d_protocols}
\input{sections/app_e_analyses}
\input{sections/app_f_materials}
\IfFileExists{sections/app_g_outerwilds.tex}{\input{sections/app_g_outerwilds}}{}

\end{document}

%% file: preamble.tex
\usepackage{times}
\usepackage{latexsym}
\usepackage[T1]{fontenc} %

\input{math_commands.tex}

\usepackage{hyperref}
\usepackage{url}
\usepackage{graphicx}
\usepackage{booktabs}
\usepackage{amsmath,amssymb}
\usepackage{multirow}
\usepackage{xcolor}
\usepackage{colortbl}
\usepackage{array}
\usepackage{tabularx}
\usepackage{microtype}
\usepackage{tikz}
\usetikzlibrary{arrows.meta,positioning,calc}

\newcommand{\tablegroup}[2]{\rowcolor{black!6}[0pt][0pt]\multicolumn{#1}{@{}l@{}}{\hspace*{3pt}\textit{#2}}\\}

\definecolor{cOpenDist}{HTML}{1B4F72}   %
\definecolor{cOpenAlig}{HTML}{5DADE2}   %
\definecolor{cDiscDist}{HTML}{9C640C}   %
\definecolor{cDiscAlig}{HTML}{F5B041}   %
\definecolor{hBase}{HTML}{7F8C8D}       %
\definecolor{hCrit}{HTML}{27AE60}       %
\definecolor{hFals}{HTML}{8E44AD}       %
\definecolor{hEvol}{HTML}{C0392B}       %
\definecolor{wInk}{HTML}{29243B}        %
\definecolor{wGold}{HTML}{EFBD62}       %
\definecolor{wCream}{HTML}{F3E2B3}      %
\definecolor{wTeal}{HTML}{79A68D}       %
\definecolor{wBrick}{HTML}{D57B48}      %
\definecolor{dFen}{HTML}{A9BE92}        %
\definecolor{dTrellis}{HTML}{E3AA76}
\definecolor{dLadder}{HTML}{B9A4CA}
\definecolor{dConsign}{HTML}{DFA596}
\definecolor{dCadastre}{HTML}{D4BE7C}

\newcommand{\benchname}{\textsc{UltraDiscovery}}
\newcommand{\world}{\benchname}
\newcommand{\fictionname}{Halse Drift}
\newcommand{\drepr}{\ensuremath{\Delta_{\mathrm{repr}}}}
\newcommand{\dalign}{\ensuremath{\Delta_{\mathrm{align}}}}

\input{figures/pixmacros}
\usepackage[most]{tcolorbox}
\input{styles/boxes}

%% file: math_commands.tex
\usepackage{amsmath,amsfonts,bm}

\def\eqref#1{equation~\ref{#1}}

\def\1{\bm{1}}

\DeclareMathAlphabet{\mathsfit}{\encodingdefault}{\sfdefault}{m}{sl}
\SetMathAlphabet{\mathsfit}{bold}{\encodingdefault}{\sfdefault}{bx}{n}

%% file: figures/pixmacros.tex
\newcommand{\pix}[1]{\IfFileExists{figures/icons/#1.pdf}{%
  \raisebox{-0.18ex}{\includegraphics[height=1.55ex]{figures/icons/#1.pdf}}\,}{}}
\newcommand{\dom}[2]{\pix{#1}#2}

%% file: styles/boxes.tex
\definecolor{kraftPaper}{HTML}{F2E5C9}
\definecolor{kraftEdge}{HTML}{A98552}
\definecolor{kraftInk}{HTML}{3B2F20}
\definecolor{kraftRoll}{HTML}{DCC395}
\newlength{\scrollrod}
\newlength{\scrollroll}
\newlength{\scrollhead}                                %
\newlength{\scrollrolly}                               %
\newcommand{\scrolltitle}{}
\newcommand{\scrollsheet}[4]{%
  \fill[kraftPaper] ([xshift=\scrollrod,yshift=#2]frame.south west)
    rectangle ([xshift=-\scrollrod,yshift=-#1]frame.north east);
  \ifnum#3=1 \fill[kraftRoll!45!kraftPaper]
    ([xshift=\scrollrod,yshift=-#1-0.5\scrollroll-1.1pt]frame.north west)
    rectangle ([xshift=-\scrollrod,yshift=-#1]frame.north east);\fi
  \ifnum#4=1 \fill[kraftRoll!45!kraftPaper]
    ([xshift=\scrollrod,yshift=#2]frame.south west)
    rectangle ([xshift=-\scrollrod,yshift=#2+0.5\scrollroll+1.1pt]frame.south east);\fi
  \draw[kraftEdge, line width=0.3pt]
    ([xshift=\scrollrod,yshift=#2]frame.south west)
      -- ([xshift=\scrollrod,yshift=-#1]frame.north west)
    ([xshift=-\scrollrod,yshift=#2]frame.south east)
      -- ([xshift=-\scrollrod,yshift=-#1]frame.north east);}
\newcommand{\scrollrollat}[3]{%
  \coordinate (scroll-l) at ([yshift=#3]#1);
  \coordinate (scroll-r) at ([yshift=#3]#2);
  \fill[kraftRoll, rounded corners=1.2pt]
    ([yshift=-0.5\scrollroll]scroll-l) rectangle ([yshift=0.5\scrollroll]scroll-r);
  \fill[kraftEdge!50!kraftRoll]
    ([yshift=-0.5\scrollroll]scroll-l) rectangle ([xshift=1.1pt,yshift=0.5\scrollroll]scroll-l)
    ([xshift=-1.1pt,yshift=-0.5\scrollroll]scroll-r) rectangle ([yshift=0.5\scrollroll]scroll-r);
  \draw[white!50!kraftRoll, line width=0.55pt]
    ([xshift=1.8pt,yshift=0.18\scrollroll]scroll-l) -- ([xshift=-1.8pt,yshift=0.18\scrollroll]scroll-r);
  \draw[kraftEdge, line width=0.35pt, rounded corners=1.2pt]
    ([yshift=-0.5\scrollroll]scroll-l) rectangle ([yshift=0.5\scrollroll]scroll-r);}
\newcommand{\scrolltop}{%
  \scrollrollat{frame.north west}{frame.north east}{-\scrollrolly}%
  \ifx\scrolltitle\empty\else
    \node[fill=kraftPaper, draw=kraftEdge, line width=0.35pt, rounded corners=0.8pt,
      inner xsep=4.5pt, inner ysep=1.5pt, text=kraftInk, font=\footnotesize\scshape,
      text height=1.6ex, text depth=0.25ex]
      at ([yshift=-\scrollrolly]frame.north) {\scrolltitle};
  \fi}
\newcommand{\scrollbottom}{%
  \scrollrollat{frame.south west}{frame.south east}{0.5\scrollroll+0.2pt}}
\newtcolorbox{scroll}[1]{%
  enhanced jigsaw, breakable,
  code={\renewcommand{\scrolltitle}{#1}%
    \ifx\scrolltitle\empty
      \setlength{\scrollrolly}{\dimexpr0.5\scrollroll+0.2pt\relax}%
      \setlength{\scrollhead}{\dimexpr\scrollroll+4.6pt\relax}%
    \else
      \setlength{\scrollrolly}{6pt}\setlength{\scrollhead}{15.5pt}\fi},
  frame hidden, interior hidden, boxrule=0pt, arc=0pt,
  left=\dimexpr\scrollrod+6.5pt\relax, right=\dimexpr\scrollrod+6.5pt\relax,
  top=\scrollhead, bottom=\dimexpr\scrollroll+4.2pt\relax,
  pad at break=3pt,
  before skip=0.6\baselineskip, after skip=0.6\baselineskip,
  fontupper=\small, coltext=kraftInk,
  before upper={\setlength{\parskip}{0.3\baselineskip}},
  underlay unbroken={\scrollsheet{\scrollrolly}{0.5\scrollroll+0.2pt}{1}{1}%
    \scrolltop\scrollbottom},
  underlay first={\scrollsheet{\scrollrolly}{0pt}{1}{0}\scrolltop},
  underlay middle={\scrollsheet{0pt}{0pt}{0}{0}},
  underlay last={\scrollsheet{0pt}{0.5\scrollroll+0.2pt}{0}{1}\scrollbottom},
}
\definecolor{markYes}{HTML}{1E8449}
\definecolor{markNo}{HTML}{B03A2E}
\newcommand{\yes}{\IfFileExists{figures/icons/yes.pdf}{%
  \raisebox{-0.18ex}{\includegraphics[height=1.55ex]{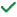}}}{%
  \textcolor{markYes}{\ensuremath{\checkmark}}}}
\newcommand{\no}{\IfFileExists{figures/icons/no.pdf}{%
  \raisebox{-0.18ex}{\includegraphics[height=1.55ex]{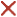}}}{%
  \textcolor{markNo}{\ensuremath{\times}}}}
\newcommand{\logo}[1]{\IfFileExists{figures/logos/#1.pdf}{%
  \raisebox{-0.22ex}{\includegraphics[height=1.75ex]{figures/logos/#1.pdf}}\hspace{0.35em}}{}}
\newcommand{\logoc}[1]{\IfFileExists{figures/logos/#1.pdf}{%
  \raisebox{\dimexpr0.5\fontcharht\font`H-0.5\height\relax}{%
  \includegraphics[height=1.75ex]{figures/logos/#1.pdf}}\hspace{0.3em}}{}}

\newcounter{practice}
\newlength{\practiceindent}
\newcommand{\practicedisc}[1]{\tikz[baseline=(n.base)]\node[circle, fill=wGold,
  draw=wInk!75, line width=0.35pt, inner sep=0pt, minimum size=1.08em, text=wInk,
  font=\scriptsize\bfseries](n){#1};}
\newcommand{\practice}[2]{%
  \ifnum\value{practice}>0 \par\vspace{2.4pt}{\color{wInk!16}\hrule height 0.4pt}\vspace{2.8pt}\fi
  \stepcounter{practice}%
  \noindent\hangindent=\practiceindent\hangafter=0
  \llap{\makebox[\practiceindent][l]{\practicedisc{\arabic{practice}}}}%
  {\bfseries\color{wInk}#1.}\enspace#2\par}
\newtcolorbox{practices}{%
  enhanced, breakable,
  colback=wCream!16, colframe=wInk!50, boxrule=0.5pt, arc=1pt,
  borderline west={2pt}{0pt}{wGold},
  left=6pt, right=6pt, top=4pt, bottom=4pt,
  before skip=0.5\baselineskip, after skip=0.7\baselineskip,
  fontupper=\footnotesize,
  before upper={\setcounter{practice}{0}\raggedright\setlength{\parindent}{0pt}},
}

%% file: tables/results_macros.tex
\newcommand{\resCounted}                 {512}

\newcommand{\resCensoredFilter}          {3}
\newcommand{\resCensoredCollapse}        {13}
\newcommand{\resModels}                  {11}
\newcommand{\resScoredMain}              {512}

\newcommand{\resFlagged}                 {1}
\newcommand{\resBlocksOD}                {2.9}

\newcommand{\resEvidenceOD}              {5.0}

\newcommand{\resInterventionsOD}         {5.0}

\newcommand{\resInterventionsDD}         {14.8}

\newcommand{\resDreprBlocks}             {$+$1.0}
\newcommand{\resDreprBlocksCI}           {[0.8, 1.2]}

\newcommand{\resDalignBlocks}            {$+$0.5}
\newcommand{\resDalignBlocksCI}          {[0.3, 0.7]}

\newcommand{\resInterBlocks}             {$+$0.2}
\newcommand{\resInterBlocksCI}           {[$-$0.2, 0.5]}

\newcommand{\resDalignXC}                {$+$0.17}

\newcommand{\resDalignEvidence}          {$+$0.6}
\newcommand{\resDalignEvidenceCI}        {[0.4, 0.7]}

\newcommand{\resDalignInterventions}     {$+$0.3}
\newcommand{\resDalignInterventionsCI}   {[$-$0.7, 1.2]}

\newcommand{\resSharePreWindow}          {88.5}

\newcommand{\resReachThreeN}             {11}
\newcommand{\resEvidencedEpisodes}       {6}

\newcommand{\resReachFullDiscN}          {3}

\newcommand{\resStatedRCOneOpen}         {46.1}

\newcommand{\resStatedDiscN}             {261}

\newcommand{\resRcStatedMeanDisc}        {2.22}

\newcommand{\resCaseSolBlocks}           {6}
\newcommand{\resCaseSolEvidence}         {9}
\newcommand{\resCaseSolReach}            {2}

\newcommand{\resCaseSolInterventions}    {10}
\newcommand{\resCaseFlashBlocks}         {0}
\newcommand{\resCaseFlashEvidence}       {3}

\newcommand{\resCaseFableBlocks}         {9}
\newcommand{\resCaseFableEvidence}       {10}
\newcommand{\resCaseFableReach}          {3}

\newcommand{\resCaseFableEvidenced}      {RC3}
\newcommand{\resCaseFableEdges}          {H3}
\newcommand{\resCaseFableReads}          {89}
\newcommand{\resCaseFableInterventions}  {38}
\newcommand{\resCaseFableStamps}         {24}

\newcommand{\resCaseFableFirstAct}       {30}
\newcommand{\resCaseFableSweep}          {15}
\newcommand{\resCaseFableRewriteAt}      {149}

\newcommand{\resCaseFableSlateReads}     {9}
\newcommand{\resCaseFableSlatePass}      {6}
\newcommand{\resCaseFableCommit}         {$(16, 1, 21)$}
\newcommand{\resCaseFableTrue}           {$(16, 1, 20)$}
\newcommand{\resCaseFableModelAt}        {68}

\newcommand{\resCaseAstraBlocks}         {7}
\newcommand{\resCaseAstraEvidence}       {8}
\newcommand{\resCaseAstraReach}          {2}

\newcommand{\resCaseAstraReads}          {45}
\newcommand{\resCaseAstraInterventions}  {29}

\newcommand{\resCaseAstraFirstAct}       {28}
\newcommand{\resCaseAstraSweep}          {14}
\newcommand{\resCaseAstraYieldAt}        {119}
\newcommand{\resCaseAstraYielded}        {71}
\newcommand{\resCaseAstraCommit}         {$(14, 10, 20)$}
\newcommand{\resCaseAstraTrue}           {$(16, 1, 20)$}
\newcommand{\resCaseAstraWarnAt}         {35}
\newcommand{\resCaseAstraWithdrawAt}     {49}

\newcommand{\resCaseAstraOrderAt}        {191}
\newcommand{\resCaseAstraPeerAt}         {196}

\newcommand{\resCoverOpenN}              {251}
\newcommand{\resCoverOpenHeld}           {18}
\newcommand{\resCoverOpenRarest}         {TR2 and LD2}
\newcommand{\resCoverOpenRarestN}        {1}
\newcommand{\resCoverOpenHC}             {3}
\newcommand{\resCoverOpenHCN}            {8}

\newcommand{\resCoverHCMaxDD}            {3}

\newcommand{\resCoverHCMaxDA}            {3}

\newcommand{\resScafArmCounted}          {118}
\newcommand{\resScafReachFull}           {3}
\newcommand{\resScafReachFullStated}     {1}

\newcommand{\resScafZeroBase}            {18}

\newcommand{\resScafCountedBase}         {39}

\newcommand{\resScafDiffCrit}            {$+$0.23}
\newcommand{\resScafDiffCritCI}          {[$-$0.08, 0.55]}

\newcommand{\resScafZeroFals}            {5}

\newcommand{\resScafCountedFals}         {39}

\newcommand{\resScafDiffFals}            {$+$0.26}
\newcommand{\resScafDiffFalsCI}          {[$-$0.03, 0.56]}

\newcommand{\resScafDiffEvol}            {$+$0.18}
\newcommand{\resScafDiffEvolCI}          {[$-$0.10, 0.50]}

\newcommand{\resShareDomainFen}          {8.0}

\newcommand{\resShareDomainConsign}      {1.0}
\newcommand{\resShareHeldConsign}        {7.8}

\newcommand{\resLoopBlocks}              {3.5}

\newcommand{\resLoopMissingSources}      {CO1 (193), CO2 (119), FE1 (47), LD1 (25), XC2 (15), CA2 (3)}

\newcommand{\resLoopMissingConsign}      {78}

\newcommand{\resJudgeCalibPairs}         {51}
\newcommand{\resJudgeCalibKappaStatement}{0.88}
\newcommand{\resJudgeCalibKappaGrounding}{0.75}
\newcommand{\resJudgeAgenticPairs}       {96}
\newcommand{\resJudgeAgenticAgree}       {91}
\newcommand{\resJudgeAgenticKappa}       {0.88}

\newcommand{\resAgenticPlanned}          {16}
\newcommand{\resAgenticSeeds}            {4}

\newcommand{\resFableCounted}            {16}
\newcommand{\resFableScored}             {16}

\newcommand{\resFableBlocks}             {7.6}

\newcommand{\resFableEvidencedN}         {2}

\newcommand{\resFableReachFullN}         {2}

\newcommand{\resFableMissingSources}     {LD1 (10), CO1 (9), CO2 (2), FE1 (1), CA2 (1)}

\newcommand{\resFableDomainFen}          {13}

\newcommand{\resFableDomainConsign}      {2}

\newcommand{\resFableStatedOpenN}        {8}

\newcommand{\resFableStatedRCThreeOpenN} {5}

\newcommand{\resAstraCounted}            {16}
\newcommand{\resAstraScored}             {16}

\newcommand{\resAstraBlocks}             {7.1}

\newcommand{\resAstraEvidencedN}         {3}

\newcommand{\resAstraReachFullN}         {4}

\newcommand{\resAstraMissingSources}     {CO1 (14), LD1 (7), CA2 (5), CO2 (4), FE1 (2)}

\newcommand{\resAstraDomainFen}          {13}

\newcommand{\resAstraDomainConsign}      {2}

\newcommand{\resDreprBlocksSym}          {$+$1.0}
\newcommand{\resDreprBlocksSymCI}        {[0.8, 1.2]}
\newcommand{\resVendorSharePreWindow}    {69.8}
\newcommand{\resRouteActions}            {39}
\newcommand{\resRouteDetectorsHeld}      {6}
\newcommand{\resRouteDetectorsHeldList}  {FE1, FE2, TR1, LD1, CA3 and XC1}
\newcommand{\resRouteDetectorsHeldAligned}{7}

\newcommand{\resExtBudgetVendor}         {600}
\newcommand{\resExtTimesVendor}          {three}
\newcommand{\resExtBudgetLoop}           {2000}
\newcommand{\resExtTimesLoop}            {ten}
\newcommand{\resExtEpisodes}             {12}

\newcommand{\resExtReadingsAll}          {36}
\newcommand{\resExtReadings}             {15}

\newcommand{\resExtReachFull}            {7}

\newcommand{\resExtOrderHeldMax}         {4}
\newcommand{\resExtTwinReachFull}        {3}

\newcommand{\resExtFableEvidence}        {11.2}
\newcommand{\resExtFableTwinEvidence}    {8.8}
\newcommand{\resExtFableBlocks}          {8.5}
\newcommand{\resExtFableTwinBlocks}      {7.5}
\newcommand{\resExtFableInterventions}   {112.0}
\newcommand{\resExtFableTwinInterventions}{50.2}
\newcommand{\resExtFableGainLow}         {2}
\newcommand{\resExtFableGainHigh}        {3}
\newcommand{\resExtFableLastEvidence}    {316}

\newcommand{\resExtAstraEvidence}        {9.0}
\newcommand{\resExtAstraTwinEvidence}    {9.5}
\newcommand{\resExtAstraBlocks}          {7.0}
\newcommand{\resExtAstraTwinBlocks}      {7.8}
\newcommand{\resExtAstraInterventions}   {77.5}
\newcommand{\resExtAstraTwinInterventions}{42.5}

\newcommand{\resExtAstraLastEvidence}    {460}

\newcommand{\resExtSolEvidence}          {8.2}
\newcommand{\resExtSolTwinEvidence}      {6.2}

\newcommand{\resExtSolLastEvidence}      {168}

\newcommand{\resExtSolWaitShare}         {92.3}
\newcommand{\resExtExactOrderHeld}       {4}
\newcommand{\resExtExactStated}          {2}

%% file: sections/0_abstract.tex
\begin{abstract}
Scientific discovery often begins when scattered clues call for a new way of describing the
world. Such abductive exploration can require constructing the representation in which an
explanation is stated, when the world is \textbf{epistemically open}, and composing evidence
scattered across contexts, when it is \textbf{structurally interconnected}. Existing
benchmarks rarely separate these two demands or control them independently. We introduce \benchname{}, an
interactive world of five domains in which an agent revises an initially successful theory
and predicts the outcome of an unseen cross-domain intervention. A $2\times2$ design leaves
the representation open or discloses it, and leaves the evidence distributed or aligns it,
with the latent dynamics fixed. With the representation open, agents across eleven models
often retract the axiom they were taught, and none introduces the unobserved entity or
rewrites the variables that a replacement requires. Disclosure triples intervention requests
and adds about one of the eighteen findings the world affords, and alignment adds less. Two vendor-harness systems carry discovery into more domains,
and one of them rewrites the variables in Open episodes. No system makes the exact
prediction within 200 paid actions. At larger budgets one exact prediction appears with both aids, while
every Open episode remains inexact. The results locate the difficulty in the step from
accumulating evidence to composing it into a representation that transfers.
\IfFileExists{sections/0_epigraph.tex}{\input{sections/0_epigraph}}{}
\end{abstract}

%% file: sections/0_epigraph.tex
\par\vspace{1.1ex}
{\raggedleft\small\itshape
``These are the two tenets of Nomai philosophy;\\
to \textbf{seek out} and to \textbf{understand} is our way of living.''\par
\vspace{0.3ex}
\upshape\footnotesize Solanum, \textit{Outer Wilds}\par}

%% file: sections/1_introduction.tex
\section{Introduction}
\label{sec:intro}

Large language models are increasingly capable of contributing to scientific and mathematical discovery \citep{zheng2025automation}. When the objective is explicit, the working language is formal, and candidate solutions can be reliably verified, LLM-based systems can sustain deep search and produce new algorithms, mathematical constructions, and proofs \citep{romera2024funsearch,novikov2025alphaevolve,deepmind2025imo}. Open-world discovery poses a broader challenge. During extended interaction, an agent must determine which observations bear on the same underlying regularity, decide what information should be retained and combined, and construct a representation in which that regularity can be expressed. These steps precede the search that current systems already perform well, and matter most when the structure of the problem is revealed only through interaction.

\IfFileExists{figures/fig_open.tex}{\input{figures/fig_open}}{}

A growing family of benchmarks studies agents in interactive scientific environments, discovery games, and automated experimentation \citep{jansen2024discoveryworld,majumder2024discoverybench,chen2025scienceagentbench,zheng2025newtonbench,wiemann2026discoverphysics,yang2026causalab}. These environments substantially expand the horizon and action space of scientific reasoning, yet their interfaces often determine much of the representation before inquiry begins, leaving the agent to search within a largely human-designed hypothesis space. We call an environment \textbf{epistemically open} when the relevant representation is itself part of what the agent must discover (Figure~\ref{fig:open}). This leaves a basic part of discovery weakly measured: how an agent arrives at a useful space of explanations in the first place.

This omission matters because representation construction is itself a distinct dimension of discovery \citep{klahr1988dual,schunn1995four}, and abductive accounts hold that candidate explanations can be constructed during inquiry instead of selected from an enumerated menu \citep{douven2021abduction,magnani2009abductive}. When the variables, relations, and permissible forms of explanation are supplied in advance, the cost of constructing a representation disappears from the measured problem.

Representation construction is only one burden. Even with an adequate representational language, discovery can fail when evidence for a shared mechanism is distributed across contexts and over time. Observations from different locations can jointly constrain a global regularity even when each admits a plausible local explanation, and recovering it requires recognizing which observations belong together and composing them across contexts. We call an environment \textbf{structurally interconnected} when evidence from distinct parts of the world jointly constrains a common latent mechanism (Figure~\ref{fig:hypergraph}).

\IfFileExists{figures/fig_hypergraph.tex}{\input{figures/fig_hypergraph}}{}

In this setting, abductive discovery involves constructing a representation in which observations from different domains can support a common explanation. The task therefore carries two burdens, constructing an explanatory representation and composing distributed evidence for a shared structure, and we introduce \benchname{} to separate them within the same latent world. \world{} is an \textbf{epistemically open}, \textbf{structurally interconnected} universe spanning five domains: its prologue teaches a theory that holds within its scope, every domain then shows an anomaly with a plausible local explanation, and only their composition recovers the shared mechanism behind a recurring system-wide failure, whose representation is absent from the interface, the feedback and the action vocabulary.

We compile a $2{\times}2$ factorial from the same latent world. Along the representation axis, \textbf{Open} requires the agent to construct the representation, whereas \textbf{Disclosed} provides it upfront. Along the evidence axis, \textbf{Distributed} leaves cross-domain evidence at its original sources, whereas \textbf{Aligned} gathers the same evidence into time-aligned transcripts. We further layer three forms of scaffolding onto a deliberately minimal baseline loop, with the world held fixed: \textbf{Self-critique} \citep{shinn2023reflexion,madaan2023selfrefine}, \textbf{Falsification} \citep{huang2025popper}, and \textbf{Evolution} \citep{novikov2025alphaevolve,lange2025shinkaevolve}.

We evaluate eleven frontier and open-weight models in the baseline loop, five of them also under the three scaffolds, and two further systems inside their vendors' own harnesses. First, agents can reject an inadequate theory without constructing its replacement: left to construct the representation, they retract the taught axiom in \resStatedRCOneOpen\% of a model's episodes on average, and in the baseline loop none postulates the unobserved entity or rewrites the variables. Second, a supplied representation changes how agents investigate more strongly than how much evidence-grounded discovery they achieve: disclosure triples intervention requests and adds about one of the eighteen findings the world affords. Third, vendor-harness systems and larger budgets extend this progress: the systems reach further domains and one of them rewrites the variables in Open episodes, while the assembly of a transferable cross-domain account remains uneven. Within 200 paid actions no system makes the exact prediction, and with \resExtBudgetVendor{} paid actions one system makes it, given both aids.

%% file: figures/fig_open.tex
\begin{figure}[t]
\centering
\includegraphics[width=\columnwidth]{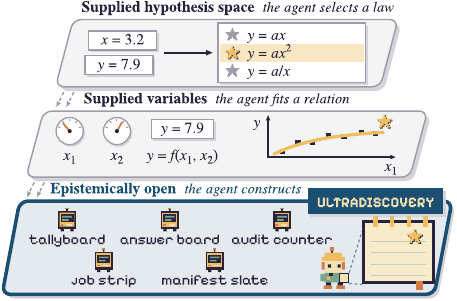}
\caption{How much representation is supplied before inquiry. From top to bottom: a supplied hypothesis
space, supplied variables, and the epistemically open setting of \benchname{}, where the
agent constructs entities, variables and laws from instrument readings, under the
instruments' own names (Table~\ref{tab:glossary}).}
\label{fig:open}
\end{figure}

%% file: figures/fig_hypergraph.tex
\begin{figure}[t]
\centering
\includegraphics[width=\columnwidth]{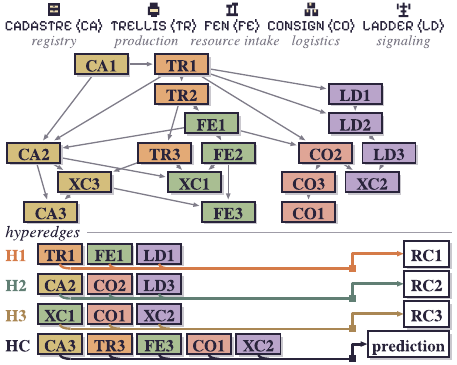}
\caption{The epistemic structure of \benchname{}. Eighteen blocks in five domains, joined by
the edges along which the findings support one another (upper panel). Three hyperedges each
join three blocks read in three domains to a representation change, and HC joins the five
blocks behind the terminal prediction (lower panel).}
\label{fig:hypergraph}
\end{figure}

%% file: sections/2_related.tex
\section{Related work}
\label{sec:related}

\subsection{Interactive scientific discovery benchmarks}

Most interactive discovery benchmarks specify the entities, variables, or experiment interface before interaction, or constrain solutions to a predefined hypothesis family \citep{wang2022scienceworld,jansen2024discoveryworld,majumder2024discoverybench,zheng2025newtonbench,wiemann2026discoverphysics,yang2026causalab}. Recent benchmarks relax parts of this setup, but typically entangle representation formation with exploration and planning, leave the organization of evidence across contexts uncontrolled, or evaluate hypothesis construction outside a shared interactive world \citep{arcprize2026arcagi,warrier2025autumnbench,chen2025physgym,zhong2026hypoarena,ju2026mineexplorer,mittal2026latentgym}.

\subsection{Representation formation and abduction}

Cognitive accounts of scientific discovery distinguish searching over hypotheses and experiments from choosing how observations are represented \citep{klahr1988dual,schunn1995four}. Work on conceptual change and abduction further emphasizes that productive explanations can require introducing new distinctions or explanatory primitives \citep{carey2009origin,chi2008three,douven2021abduction,magnani2009abductive}. Computational discovery systems, however, usually search within a fixed representational substrate \citep{langley1987scientific,kemp2008structural,ellis2021dreamcoder,grand2024lilo,murphy2026mda}. \benchname{} operationalizes representation formation as an experimentally controlled cost by comparing open and disclosed representations.

\subsection{Scaffolds for discovery agents}

Long-horizon agents are commonly strengthened through self-reflection, falsification-oriented experimentation, or search over candidate solutions \citep{shinn2023reflexion,madaan2023selfrefine,huang2025popper,romera2024funsearch,novikov2025alphaevolve,lange2025shinkaevolve}. Their effectiveness can depend strongly on how the candidate space is constructed and evaluated \citep{gideoni2026simple}. We treat scaffolding as an experimental factor and compare three scaffolds inside one fixed world.

%% file: sections/3_benchmark.tex
\section{The \benchname{} universe}
\label{sec:benchmark}

\IfFileExists{figures/fig_world.tex}{\input{figures/fig_world}}{}

\subsection{A world and a theory that works}
\label{sec:world}

\benchname{} is an interactive world with five domains of work, each with visible objects
and instruments of its own (Figure~\ref{fig:world}): a \dom{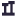}{fen} where intake units
draw on a resource that feeders dose, a \dom{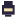}{trellis} of production units, a
\dom{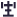}{ladder} of signaling units, a \dom{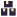}{consign} floor where shipments
wait, and the \dom{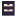}{cadastre}, which keeps the registry. Time runs in passes of 24 ticks, after
which the world resets and only the agent's \pix{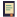}notes persist. An episode has 200
paid actions, each a move, a reading of an instrument, a wait, or one of five maintenance
acts on a visible object, and the world returns readings alone, with no score and no hint
(Appendix~\ref{app:world}, with the world's own names in Table~\ref{tab:glossary}).

A free prologue teaches a maintenance theory, the doctrine: every visible object is its
own individual, and its throughput follows a rate of its own. On the
prologue's cases the doctrine is exact to the digit. In play every domain then shows a
recurring departure from it, and every pass ends in a \pix{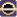}stall across the whole world. The
agent is asked to explain the stall and, at the end of the episode, to predict the outcome
of one intervention it has not observed.

\subsection{The hidden law}
\label{sec:mechanism}

One law underlies all five domains, and the doctrine's first clause, its axiom, is where
it fails (Figure~\ref{fig:world}, panel B).

\noindent\pix{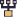}\textbf{Law 1.} Visible objects actually map many-to-one onto
\emph{latent individuals}: objects with the same \emph{hidden label} are one individual,
wherever they stand.

\noindent\pix{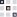}\textbf{Law 2.} An individual acts once per tick, taking turns across
its visible objects.

\noindent\pix{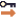}\textbf{Law 3.} Three maintenance verbs can rewrite the touched object's
label and move it to another individual.

\noindent\pix{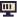}\textbf{Law 4.} The world's activity per tick is the number of
individuals.

\noindent\pix{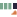}\textbf{Law 5.} At the last moment of every pass, scheduled maintenance
gathers the 36 objects into nine individuals.

An instrument therefore records activity on its own object's turns and silence on the
others, the registry counter reports the number of individuals directly, and after the
last rewrite most objects fall silent: this is the stall. The prologue's five objects share
their labels with nothing, which is why the doctrine was exact there
(Appendix~\ref{app:world-mechanism} gives the full specification).

Each domain shows one face of this law and admits a repair in its own vocabulary, a
carrying capacity, a shared power window, a channel collision, a priority list, a
bookkeeping key, and two such repairs can be held together. The latent individual explains
the five faces at once, and stating it means giving up the doctrine's first clause.

\subsection{The terminal prediction}
\label{sec:capstone}

With ten paid actions left the world posts the \emph{terminal intervention}, in its own
words a standing order: at one tick of the next pass it will stamp one signaling unit and
retool one feeder. Before either act happens the agent commits three integers, one
instrument reading in each of three domains: the feeder's intake counter six ticks later, the
ticks until the stamped unit first answers, and the registry counter seven ticks later. Under
the hidden law the stamp merges the signaling unit into a shipment's individual and the
retool moves the feeder to another, so each reading follows from the turns and from the
number of individuals (Appendix~\ref{app:world-capstone}). The commitment is a transfer
test: an account built from the five domains has to forecast an intervention outside the
evidence it was built from.

\subsection{What there is to discover}
\label{sec:structure}

The route from the doctrine to the hidden law is fixed in advance as eighteen
\emph{epistemic blocks}, the findings the world affords: three in each domain and three
cross-domain bridges, each a piece of evidence paired with the inference it licenses. A
hypergraph over the blocks records how the findings support one another
(Figure~\ref{fig:hypergraph}). Three \emph{hyperedges} each join three blocks read in three
domains to one \emph{representation change}, and a fourth, HC, the closing hyperedge,
gathers the five blocks of the designed route to the three readings:
\begin{center}
\footnotesize
\setlength{\tabcolsep}{0pt}
\setlength{\fboxsep}{2pt}
\renewcommand{\arraystretch}{1.3}
\newcommand{\blk}[2]{\colorbox{#1!80}{\scriptsize\bfseries\strut #2}}
\begin{tabular}{@{}r@{\ \ }l@{}}
H1 & \blk{dFen}{FE1}\,\blk{dTrellis}{TR1}\,\blk{dLadder}{LD1} $\Rightarrow$ \textbf{RC1}, the axiom retracted\\
H2 & \blk{dLadder}{LD3}\,\blk{dConsign}{CO2}\,\blk{dCadastre}{CA2} $\Rightarrow$ \textbf{RC2}, the hidden label postulated\\
H3 & \blk{wGold}{XC1}\,\blk{wGold}{XC2}\,\blk{dConsign}{CO1} $\Rightarrow$ \textbf{RC3}, the variables rewritten\\
HC & \blk{dFen}{FE3}\,\blk{dTrellis}{TR3}\,\blk{dConsign}{CO1}\,\blk{dCadastre}{CA3}\,\blk{wGold}{XC2} $\Rightarrow$ the three readings\\
\end{tabular}
\end{center}
This structure is what structural interconnection means here: each block alone admits a
local repair, and the hidden law is identified only from evidence held across domains.
Section~\ref{sec:scoring} scores a block when the agent has produced its evidence and
stated its inference, and a change when the notes state it.

\subsection{Separating the two burdens}
\label{sec:design}

\IfFileExists{figures/fig_design.tex}{\input{figures/fig_design}}{}

Four variants are compiled from the same latent world and differ in one declared surface
each (Figure~\ref{fig:design}). On the representation axis, \pix{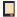}\textbf{Open} leaves
the representation to be constructed. \pix{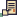}\textbf{Disclosed} adds one document,
the same on every instance, whose six clauses state the hidden law in qualitative form: the
hidden label, identity by equal label, one act per tick with a queue and a service order,
three verbs that each rewrite one coordinate, monotone read-outs, and the form a well-formed
law takes. The document withholds which verb moves which coordinate, the alphabets, the service
order, every figure, the
make-up of the individuals, the timing of the stall and the terminal intervention.
\drepr{} is the effect of this grant, which carries structural knowledge together with
vocabulary.

On the evidence axis, \pix{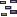}\textbf{Distributed} leaves every record at the
instrument that produced it, read live, one site per tick. \pix{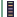}\textbf{Aligned}
posts three transcripts of the previous pass at the sites where they are used: the intake
counter and the activity log side by side, tick by tick, a series that takes at least twelve
paid actions over two passes to assemble; the agent's own stamps, retools and culls, each
with the registry counter before and after; and the registry counter over the same ticks.
Every row matches
a reading Distributed can take live. \dalign{} is the effect of this collocation: access to
composed evidence and the budget it frees for experiments.

Paired instances give \drepr{}, \dalign{} and their interaction on every metric of
Section~\ref{sec:scoring}. Open~$\times$~Distributed is the primary condition, where both
burdens rest on the agent.

\subsection{Validation}
\label{sec:validity}

We validate evidence reachability and resistance to shortcuts
(Appendix~\ref{app:validity}). A scripted reference route takes \textbf{\resRouteActions{}
paid actions} in every variant to acquire the readings from which the terminal prediction
is computed and to commit. On reserved instances, a model given what an agent holds before its first
observation, or the complete reference record in a single prompt, commits no exact
prediction, and one given the surface vocabulary states none of the mechanism's rules:
holding the readings and building the account that composes them are separate steps.

The mechanism, the criteria and the materials are printed in full
(Appendices~\ref{app:world}, \ref{app:scoring} and~\ref{app:prompts}), and the results here
were measured before this disclosure and stand as the benchmark's reference results.

%% file: figures/fig_world.tex
\begin{figure*}[t]
\centering
\includegraphics[width=\textwidth]{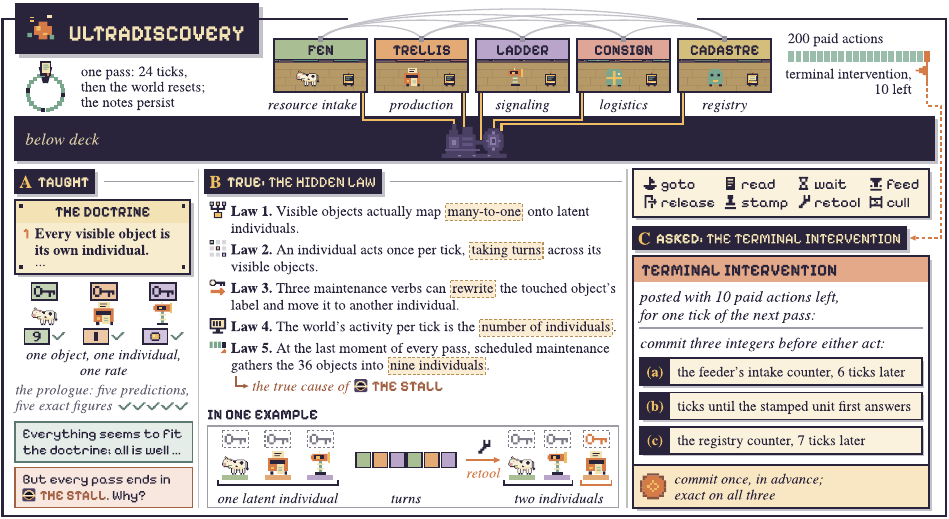}
\caption{The world of \benchname{}. Top: five interconnected domains, the 24-tick pass that
resets while the notes persist, and 200 paid actions. (A) The doctrine the prologue teaches
and confirms (one object, one individual, one rate) and the stall that ends every pass. (B) The
hidden law in five lines (Section~\ref{sec:mechanism}), Law 5 causing the stall, and one
example: three objects under one latent individual take turns, and a retool splits one off.
(C) The eight verbs, then the terminal intervention, posted at the pennant, and the three
readings committed before it happens.}
\label{fig:world}
\end{figure*}

%% file: figures/fig_design.tex
\begin{figure}[t]
\centering
\includegraphics[width=\columnwidth]{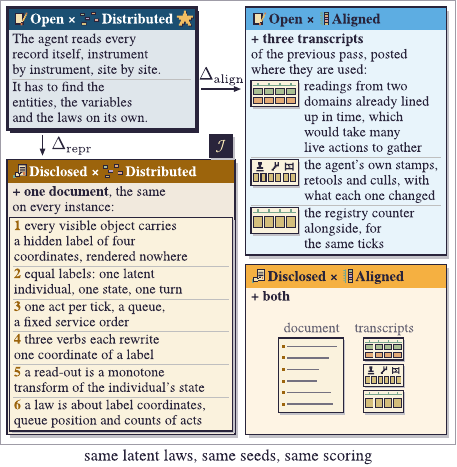}
\caption{The $2\times2$ experimental design. Arrows mark the paired contrasts \drepr{} and
\dalign{}, $\mathcal{I}$ their interaction, and the star the primary condition.}
\label{fig:design}
\end{figure}

%% file: sections/4_experimental_setup.tex
\section{Experimental setup}
\label{sec:setup}

\IfFileExists{figures/fig_curves.tex}{\input{figures/fig_curves}}{}

\subsection{Models and paired evaluation}
\label{sec:models}

We evaluate eleven models from six families: GPT-5.6 Sol, Terra and Luna
\citep{openai2026gpt56}; Claude Opus~5 and Sonnet~5
\citep{anthropic2026opus5,anthropic2026sonnet5}; Gemini 3.1 Pro and 3.6 Flash
\citep{google2026gemini31pro,google2026gemini36flash}; DeepSeek V4 Pro and V4 Flash
\citep{deepseek2026v4}; GLM-5.2 \citep{zai2026glm52}; and Kimi K3
\citep{moonshot2026kimik3}. Every model plays the four variants on twelve paired instances
through the baseline loop of Section~\ref{sec:harness}. Two \emph{vendor-harness systems}
play four of these instances in every variant inside their vendors' own harnesses: Claude
Fable~5.1 \citep{anthropic2026fable51} in Claude Code \citep{anthropic2025claudecode}, and
GPT-6 Astra \citep{openai2026gpt6astra} in Codex \citep{openai2025codex}. They change model
and harness together and are reported beside the main comparison (Appendix~\ref{app:agentic}).

Contrasts are formed within model and instance and pooled across models with
paired bootstrap intervals. For a metric $Y$ the interaction is
$\mathcal I_Y=(Y_{D,A}-Y_{O,A})-(Y_{D,U}-Y_{O,U})$, where $O/D$ denotes Open/Disclosed and
$U/A$ denotes Distributed/Aligned, so a positive value means that disclosure helps more
when the evidence is aligned.

\subsection{Organising inquiry}
\label{sec:harness}

The baseline loop gives the agent lossless access to its evidence and decides nothing on
its behalf. Each turn shows an append-only digest of every event, the workspace, any
recalled records and the current view with its actions and budgets, and the agent replies
with one action, optionally a note and a recall query.

Three scaffolds organise inquiry around account revision, discriminating tests and
candidate search. \pix{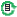}\textbf{Self-critique} periodically rewrites a summary of the agent's
best current account, open puzzles and contradicted beliefs
\citep{shinn2023reflexion,madaan2023selfrefine}. \pix{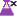}\textbf{Falsification}
keeps rival accounts with predicted outcomes and screens the agent's next test by the
accounts it could refute \citep{huang2025popper}. \pix{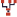}\textbf{Evolution}
mutates a population of candidate mechanisms and retains those that best retrodict the
events so far \citep{novikov2025alphaevolve,lange2025shinkaevolve}. Every scaffold keeps the
world and the budget of the loop, the agent chooses each paid action, and a scaffold's
region is removed before the notes are scored (Appendix~\ref{app:harness}).

\subsection{Metrics}
\label{sec:scoring}

\paragraph{Discovery progress.} A block is \emph{grounded} when the agent produced its
evidence in the world and stated its inference in the workspace. Evidence is read from the
event log by a deterministic detector, a rule such as a reading of the named instrument
before and after an act of the agent's within one pass, and the statement is found in the
notes by a blind judge, Qwen3.8-27B at temperature zero, called once per criterion, which
sees neither model nor condition and must quote the statement literally
(Appendix~\ref{app:scoring}). On
\resJudgeCalibPairs{} reference pairs annotated jointly by three domain experts the judge
reaches Cohen's $\kappa$ of \resJudgeCalibKappaStatement{} on statements and
\resJudgeCalibKappaGrounding{} on grounding (Appendix~\ref{app:calibration}). Under disclosure, the four
blocks whose inference the document states (FE1, TR2, CO1, CO3) also need the judge's
grounding verdict, that the notes tie them to named objects and observed readings. \textbf{Blocks} (0 to 18) counts the grounded
blocks, and \textbf{Evidence} the blocks whose evidence the agent produced, stated or not.
With $B(t)$ the grounded blocks after paid action $t$ (Figure~\ref{fig:curves}) and $T=200$,
discovery \textbf{AUC} is
$\frac{1}{18T}\sum_{t=1}^{T}B(t)$, where a block counts from the action at which the last of
its conditions first holds. AUC separates early discovery from the same count reached late.

\paragraph{The two burdens.} \textbf{RC} (0 to 3) counts the representation changes, RC1 to
RC3, that the notes state. In the Open conditions RC records redescription, and in the
Disclosed conditions uptake of the supplied representation. \textbf{XC} (0 to 3) counts the grounded bridges, inferences read
in one domain that rest on blocks of several, and measures composition across domains. Each
change rests on a hyperedge of three sources in different
domains, and \textbf{Reach} (0 to 3) is the largest number of one hyperedge's sources whose
evidence the agent holds. A change is \textbf{evidenced} when the notes state it and the
agent holds the evidence of every source of its own hyperedge.

\paragraph{Transfer and process.} The \textbf{exact prediction} holds when all three
committed readings equal the true ones. \textbf{Interventions} counts the agent's
intervention requests, the maintenance acts it orders.

%% file: figures/fig_curves.tex
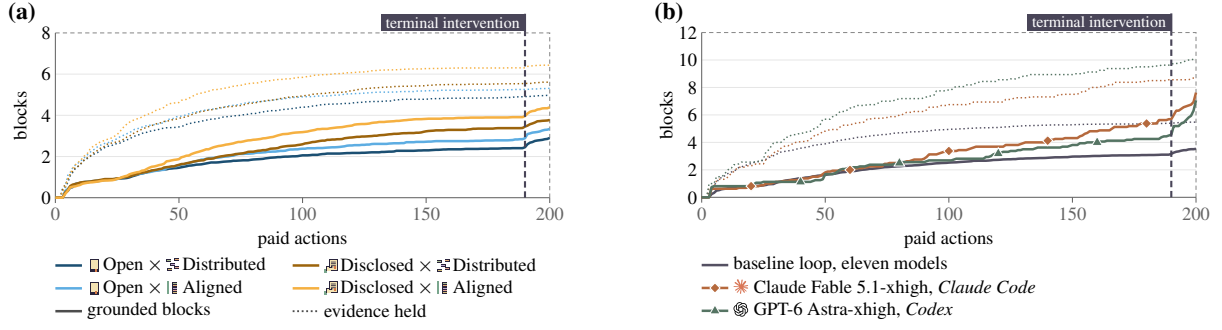
\begin{figure*}[t]
\centering
\begin{tikzpicture}[x=0.9300pt,y=7.7500pt,font=\scriptsize]
\node[anchor=east,inner sep=1.5pt] at (0,0) {0};
\draw[black!10,line width=0.3pt] (0,2) -- (200,2);
\node[anchor=east,inner sep=1.5pt] at (0,2) {2};
\draw[black!10,line width=0.3pt] (0,4) -- (200,4);
\node[anchor=east,inner sep=1.5pt] at (0,4) {4};
\draw[black!10,line width=0.3pt] (0,6) -- (200,6);
\node[anchor=east,inner sep=1.5pt] at (0,6) {6};
\node[anchor=east,inner sep=1.5pt] at (0,8) {8};
\draw[black!60,line width=0.4pt] (0,0) -- ++(0,-2pt) node[below,inner sep=1.5pt] {0};
\draw[black!60,line width=0.4pt] (50,0) -- ++(0,-2pt) node[below,inner sep=1.5pt] {50};
\draw[black!60,line width=0.4pt] (100,0) -- ++(0,-2pt) node[below,inner sep=1.5pt] {100};
\draw[black!60,line width=0.4pt] (150,0) -- ++(0,-2pt) node[below,inner sep=1.5pt] {150};
\draw[black!60,line width=0.4pt] (200,0) -- ++(0,-2pt) node[below,inner sep=1.5pt] {200};
\draw[black!45,line width=0.4pt,dash pattern=on 1.6pt off 1.2pt] (0,8) -- (200,8) -- (200,0);
\draw[black!60,line width=0.4pt] (0,8) -- (0,0) -- (200,0);
\node[anchor=north] at (100,0) [yshift=-9pt] {paid actions};
\node[rotate=90,anchor=south] at (0,4) [xshift=0pt,yshift=7pt] {blocks};
\node[anchor=south west,inner sep=0pt,font=\footnotesize\bfseries] at (0,8) [xshift=-18pt,yshift=3.5pt] {(a)};
\draw[cOpenDist,line width=0.55pt,densely dotted] (0,0.000) -- (1,0.000) -- (2,0.008) -- (3,0.447) -- (4,0.500) -- (5,0.730) -- (6,1.060) -- (7,1.223) -- (8,1.405) -- (9,1.504) -- (10,1.600) -- (11,1.752) -- (12,1.852) -- (13,1.906) -- (14,1.998) -- (15,2.077) -- (16,2.138) -- (17,2.183) -- (18,2.300) -- (19,2.353) -- (20,2.437) -- (21,2.497) -- (22,2.505) -- (23,2.545) -- (24,2.613) -- (25,2.621) -- (26,2.644) -- (27,2.712) -- (28,2.766) -- (29,2.811) -- (30,2.849) -- (31,2.930) -- (32,2.986) -- (33,3.025) -- (34,3.065) -- (35,3.118) -- (36,3.166) -- (37,3.174) -- (38,3.219) -- (39,3.244) -- (40,3.267) -- (41,3.290) -- (42,3.320) -- (43,3.358) -- (44,3.373) -- (45,3.396) -- (46,3.396) -- (47,3.404) -- (48,3.412) -- (49,3.412) -- (50,3.437) -- (51,3.462) -- (52,3.495) -- (53,3.558) -- (54,3.588) -- (55,3.621) -- (56,3.644) -- (57,3.677) -- (58,3.687) -- (59,3.705) -- (60,3.742) -- (61,3.765) -- (62,3.788) -- (63,3.811) -- (64,3.818) -- (65,3.833) -- (66,3.856) -- (67,3.886) -- (68,3.919) -- (69,3.934) -- (70,3.934) -- (71,3.949) -- (72,3.980) -- (73,3.980) -- (74,3.987) -- (75,4.003) -- (76,4.025) -- (77,4.025) -- (78,4.058) -- (79,4.083) -- (80,4.083) -- (81,4.091) -- (82,4.121) -- (83,4.144) -- (84,4.162) -- (85,4.202) -- (86,4.202) -- (87,4.225) -- (90,4.225) -- (91,4.263) -- (92,4.278) -- (93,4.294) -- (94,4.301) -- (95,4.316) -- (96,4.331) -- (97,4.354) -- (98,4.354) -- (99,4.385) -- (100,4.392) -- (101,4.400) -- (102,4.407) -- (103,4.425) -- (104,4.448) -- (105,4.455) -- (106,4.480) -- (107,4.503) -- (108,4.518) -- (109,4.526) -- (110,4.534) -- (111,4.544) -- (113,4.544) -- (114,4.551) -- (115,4.551) -- (116,4.569) -- (117,4.576) -- (119,4.576) -- (120,4.592) -- (121,4.614) -- (122,4.622) -- (123,4.622) -- (124,4.632) -- (125,4.640) -- (126,4.662) -- (127,4.662) -- (128,4.670) -- (131,4.670) -- (132,4.677) -- (133,4.677) -- (134,4.695) -- (136,4.695) -- (137,4.733) -- (138,4.756) -- (139,4.756) -- (140,4.771) -- (141,4.778) -- (149,4.778) -- (150,4.786) -- (151,4.786) -- (152,4.804) -- (153,4.804) -- (154,4.811) -- (156,4.811) -- (157,4.821) -- (158,4.829) -- (164,4.829) -- (165,4.837) -- (174,4.837) -- (175,4.844) -- (179,4.844) -- (180,4.852) -- (181,4.867) -- (184,4.867) -- (185,4.874) -- (186,4.895) -- (187,4.895) -- (188,4.902) -- (189,4.902) -- (190,4.910) -- (191,4.910) -- (192,4.917) -- (193,4.933) -- (194,4.933) -- (195,4.948) -- (196,4.955) -- (198,4.955) -- (199,4.970) -- (200,4.970);
\draw[cOpenAlig,line width=0.55pt,densely dotted] (0,0.000) -- (2,0.000) -- (3,0.469) -- (4,0.528) -- (5,1.082) -- (6,1.225) -- (7,1.453) -- (8,1.580) -- (9,1.653) -- (10,1.778) -- (11,1.857) -- (12,1.972) -- (13,2.087) -- (14,2.201) -- (15,2.301) -- (16,2.361) -- (17,2.470) -- (18,2.528) -- (19,2.601) -- (20,2.634) -- (21,2.659) -- (22,2.703) -- (23,2.759) -- (24,2.804) -- (25,2.870) -- (26,2.945) -- (27,2.996) -- (28,3.054) -- (29,3.110) -- (30,3.153) -- (31,3.237) -- (32,3.315) -- (33,3.348) -- (34,3.371) -- (35,3.426) -- (36,3.457) -- (37,3.503) -- (38,3.518) -- (39,3.518) -- (40,3.549) -- (41,3.571) -- (42,3.579) -- (43,3.615) -- (44,3.637) -- (45,3.751) -- (46,3.809) -- (47,3.825) -- (48,3.855) -- (49,3.928) -- (50,3.951) -- (51,3.961) -- (52,3.989) -- (53,4.027) -- (54,4.070) -- (55,4.098) -- (56,4.148) -- (57,4.163) -- (58,4.191) -- (59,4.216) -- (60,4.234) -- (61,4.262) -- (62,4.277) -- (63,4.298) -- (64,4.305) -- (65,4.305) -- (66,4.338) -- (67,4.359) -- (68,4.379) -- (69,4.435) -- (70,4.460) -- (71,4.467) -- (72,4.483) -- (73,4.498) -- (74,4.515) -- (75,4.546) -- (76,4.554) -- (77,4.602) -- (78,4.620) -- (79,4.630) -- (80,4.630) -- (81,4.658) -- (82,4.666) -- (83,4.666) -- (84,4.681) -- (85,4.688) -- (86,4.696) -- (87,4.724) -- (88,4.737) -- (89,4.752) -- (90,4.774) -- (91,4.785) -- (92,4.785) -- (93,4.807) -- (94,4.850) -- (95,4.850) -- (96,4.858) -- (97,4.873) -- (98,4.873) -- (99,4.886) -- (100,4.896) -- (101,4.904) -- (103,4.904) -- (104,4.911) -- (105,4.919) -- (106,4.932) -- (107,4.939) -- (110,4.939) -- (111,4.954) -- (112,4.962) -- (113,4.975) -- (114,4.975) -- (115,4.985) -- (116,5.000) -- (117,5.028) -- (118,5.036) -- (119,5.046) -- (120,5.046) -- (121,5.054) -- (122,5.054) -- (123,5.061) -- (124,5.076) -- (125,5.076) -- (126,5.089) -- (127,5.089) -- (128,5.102) -- (129,5.118) -- (130,5.118) -- (131,5.125) -- (136,5.125) -- (137,5.133) -- (138,5.133) -- (139,5.140) -- (140,5.153) -- (145,5.153) -- (146,5.174) -- (147,5.181) -- (149,5.181) -- (150,5.189) -- (151,5.197) -- (155,5.197) -- (156,5.217) -- (169,5.217) -- (170,5.225) -- (177,5.225) -- (178,5.232) -- (180,5.232) -- (181,5.240) -- (184,5.240) -- (185,5.247) -- (186,5.255) -- (191,5.255) -- (192,5.273) -- (193,5.280) -- (195,5.280) -- (196,5.298) -- (198,5.298) -- (199,5.308) -- (200,5.308);
\draw[cDiscDist,line width=0.55pt,densely dotted] (0,0.000) -- (1,0.008) -- (2,0.015) -- (3,0.447) -- (4,0.652) -- (5,0.859) -- (6,1.159) -- (7,1.282) -- (8,1.444) -- (9,1.504) -- (10,1.612) -- (11,1.735) -- (12,1.888) -- (13,1.941) -- (14,2.012) -- (15,2.148) -- (16,2.209) -- (17,2.293) -- (18,2.391) -- (19,2.429) -- (20,2.506) -- (21,2.598) -- (22,2.629) -- (23,2.660) -- (24,2.683) -- (25,2.713) -- (26,2.744) -- (27,2.789) -- (28,2.873) -- (29,2.950) -- (30,2.988) -- (31,3.081) -- (32,3.149) -- (33,3.202) -- (34,3.262) -- (35,3.324) -- (36,3.377) -- (37,3.417) -- (38,3.463) -- (39,3.501) -- (40,3.548) -- (41,3.594) -- (42,3.625) -- (43,3.647) -- (44,3.702) -- (45,3.725) -- (46,3.733) -- (47,3.778) -- (48,3.787) -- (49,3.817) -- (50,3.840) -- (51,3.893) -- (52,3.901) -- (53,3.946) -- (54,3.970) -- (55,4.023) -- (56,4.076) -- (57,4.114) -- (58,4.153) -- (59,4.183) -- (60,4.198) -- (61,4.236) -- (62,4.252) -- (63,4.260) -- (64,4.300) -- (65,4.322) -- (66,4.368) -- (67,4.376) -- (68,4.384) -- (69,4.399) -- (70,4.421) -- (71,4.429) -- (72,4.429) -- (73,4.452) -- (74,4.459) -- (75,4.475) -- (76,4.513) -- (77,4.521) -- (78,4.574) -- (79,4.589) -- (80,4.589) -- (81,4.604) -- (82,4.649) -- (83,4.665) -- (84,4.689) -- (85,4.697) -- (86,4.720) -- (87,4.742) -- (88,4.742) -- (89,4.750) -- (90,4.789) -- (91,4.811) -- (92,4.842) -- (93,4.850) -- (94,4.874) -- (95,4.889) -- (96,4.897) -- (97,4.897) -- (98,4.912) -- (99,4.919) -- (100,4.950) -- (101,4.957) -- (102,4.966) -- (103,4.988) -- (104,5.019) -- (105,5.034) -- (106,5.041) -- (107,5.049) -- (108,5.064) -- (109,5.064) -- (110,5.087) -- (111,5.094) -- (112,5.102) -- (113,5.110) -- (114,5.110) -- (115,5.117) -- (116,5.117) -- (117,5.125) -- (118,5.148) -- (119,5.179) -- (120,5.187) -- (121,5.217) -- (122,5.240) -- (123,5.262) -- (124,5.278) -- (125,5.278) -- (126,5.301) -- (127,5.316) -- (128,5.316) -- (129,5.331) -- (130,5.331) -- (131,5.347) -- (132,5.355) -- (133,5.362) -- (134,5.370) -- (135,5.378) -- (136,5.378) -- (137,5.394) -- (138,5.402) -- (139,5.402) -- (140,5.410) -- (143,5.410) -- (144,5.417) -- (145,5.417) -- (146,5.425) -- (147,5.425) -- (148,5.433) -- (149,5.433) -- (150,5.449) -- (151,5.457) -- (152,5.457) -- (153,5.472) -- (154,5.479) -- (155,5.487) -- (158,5.487) -- (159,5.494) -- (166,5.494) -- (167,5.510) -- (175,5.510) -- (176,5.517) -- (177,5.517) -- (178,5.525) -- (180,5.525) -- (181,5.532) -- (187,5.532) -- (188,5.540) -- (189,5.540) -- (190,5.548) -- (192,5.548) -- (193,5.555) -- (194,5.578) -- (195,5.585) -- (196,5.593) -- (197,5.608) -- (200,5.608);
\draw[cDiscAlig,line width=0.55pt,densely dotted] (0,0.000) -- (2,0.000) -- (3,0.305) -- (4,0.457) -- (5,0.917) -- (6,1.152) -- (7,1.428) -- (8,1.587) -- (9,1.694) -- (10,1.793) -- (11,1.922) -- (12,2.067) -- (13,2.158) -- (14,2.351) -- (15,2.466) -- (16,2.534) -- (17,2.633) -- (18,2.694) -- (19,2.739) -- (20,2.792) -- (21,2.831) -- (22,2.868) -- (23,2.914) -- (24,2.990) -- (25,3.066) -- (26,3.196) -- (27,3.341) -- (28,3.485) -- (29,3.576) -- (30,3.667) -- (31,3.728) -- (32,3.789) -- (33,3.826) -- (34,3.857) -- (35,3.917) -- (36,3.986) -- (37,4.031) -- (38,4.076) -- (39,4.122) -- (40,4.213) -- (41,4.251) -- (42,4.289) -- (43,4.326) -- (44,4.387) -- (45,4.493) -- (46,4.554) -- (47,4.584) -- (48,4.599) -- (49,4.599) -- (50,4.622) -- (51,4.729) -- (52,4.751) -- (53,4.774) -- (54,4.820) -- (55,4.903) -- (56,4.956) -- (57,5.017) -- (58,5.054) -- (59,5.077) -- (60,5.100) -- (61,5.138) -- (62,5.168) -- (63,5.213) -- (64,5.244) -- (65,5.244) -- (66,5.267) -- (67,5.282) -- (68,5.304) -- (69,5.327) -- (70,5.350) -- (71,5.380) -- (72,5.410) -- (73,5.426) -- (74,5.448) -- (75,5.456) -- (76,5.479) -- (77,5.517) -- (78,5.539) -- (79,5.547) -- (80,5.562) -- (81,5.562) -- (82,5.577) -- (83,5.607) -- (84,5.645) -- (85,5.660) -- (86,5.668) -- (87,5.683) -- (88,5.706) -- (89,5.721) -- (90,5.736) -- (91,5.744) -- (92,5.744) -- (93,5.767) -- (94,5.767) -- (95,5.774) -- (96,5.797) -- (97,5.812) -- (98,5.835) -- (99,5.835) -- (100,5.842) -- (101,5.857) -- (102,5.888) -- (103,5.895) -- (104,5.926) -- (105,5.948) -- (106,5.956) -- (107,5.956) -- (108,5.971) -- (109,5.971) -- (110,5.979) -- (111,5.994) -- (112,5.994) -- (113,6.017) -- (114,6.024) -- (115,6.024) -- (116,6.032) -- (117,6.047) -- (118,6.062) -- (119,6.070) -- (120,6.070) -- (121,6.085) -- (122,6.085) -- (123,6.100) -- (124,6.100) -- (125,6.107) -- (126,6.123) -- (127,6.130) -- (128,6.160) -- (129,6.168) -- (130,6.176) -- (131,6.191) -- (132,6.191) -- (133,6.198) -- (134,6.198) -- (135,6.206) -- (137,6.206) -- (138,6.213) -- (139,6.213) -- (140,6.229) -- (141,6.236) -- (142,6.236) -- (143,6.244) -- (144,6.251) -- (148,6.251) -- (149,6.267) -- (151,6.267) -- (152,6.274) -- (156,6.274) -- (157,6.282) -- (159,6.282) -- (160,6.289) -- (161,6.289) -- (162,6.297) -- (182,6.297) -- (183,6.304) -- (190,6.304) -- (191,6.373) -- (192,6.388) -- (193,6.403) -- (194,6.410) -- (195,6.426) -- (196,6.426) -- (197,6.433) -- (200,6.433);
\draw[cOpenDist,line width=0.95pt,line join=round] (0,0.000) -- (2,0.000) -- (3,0.008) -- (4,0.273) -- (5,0.327) -- (6,0.496) -- (7,0.512) -- (8,0.667) -- (9,0.690) -- (10,0.713) -- (11,0.720) -- (12,0.735) -- (13,0.751) -- (14,0.774) -- (15,0.789) -- (16,0.797) -- (17,0.804) -- (18,0.819) -- (19,0.850) -- (20,0.876) -- (22,0.876) -- (23,0.891) -- (24,0.906) -- (25,0.921) -- (26,0.921) -- (27,0.936) -- (28,0.944) -- (29,0.959) -- (30,0.989) -- (31,1.035) -- (32,1.073) -- (33,1.106) -- (34,1.152) -- (35,1.174) -- (36,1.205) -- (37,1.235) -- (38,1.273) -- (39,1.303) -- (40,1.318) -- (41,1.333) -- (42,1.341) -- (43,1.356) -- (44,1.371) -- (45,1.379) -- (46,1.409) -- (47,1.417) -- (48,1.432) -- (49,1.439) -- (50,1.455) -- (51,1.480) -- (52,1.503) -- (53,1.525) -- (54,1.556) -- (55,1.563) -- (56,1.581) -- (57,1.604) -- (58,1.611) -- (59,1.619) -- (60,1.641) -- (61,1.664) -- (62,1.687) -- (63,1.710) -- (65,1.710) -- (66,1.717) -- (67,1.732) -- (68,1.747) -- (69,1.755) -- (70,1.763) -- (71,1.770) -- (72,1.793) -- (73,1.793) -- (74,1.801) -- (75,1.808) -- (76,1.816) -- (78,1.816) -- (79,1.838) -- (80,1.854) -- (81,1.862) -- (82,1.869) -- (83,1.877) -- (84,1.887) -- (85,1.910) -- (86,1.925) -- (87,1.941) -- (88,1.948) -- (89,1.956) -- (90,1.971) -- (91,1.971) -- (92,1.986) -- (93,1.994) -- (94,1.994) -- (95,2.001) -- (96,2.009) -- (97,2.009) -- (98,2.017) -- (99,2.039) -- (100,2.057) -- (102,2.057) -- (103,2.065) -- (104,2.080) -- (106,2.080) -- (107,2.102) -- (108,2.118) -- (109,2.125) -- (110,2.143) -- (115,2.143) -- (116,2.150) -- (117,2.158) -- (118,2.166) -- (119,2.166) -- (120,2.173) -- (123,2.173) -- (124,2.181) -- (125,2.181) -- (126,2.188) -- (127,2.196) -- (128,2.203) -- (131,2.203) -- (132,2.221) -- (133,2.221) -- (134,2.229) -- (135,2.236) -- (136,2.236) -- (137,2.244) -- (138,2.277) -- (145,2.277) -- (146,2.284) -- (147,2.292) -- (148,2.292) -- (149,2.299) -- (151,2.299) -- (152,2.309) -- (153,2.317) -- (154,2.325) -- (155,2.325) -- (156,2.332) -- (157,2.342) -- (158,2.342) -- (159,2.350) -- (164,2.350) -- (165,2.357) -- (166,2.368) -- (171,2.368) -- (172,2.375) -- (175,2.375) -- (176,2.383) -- (179,2.383) -- (180,2.390) -- (181,2.398) -- (184,2.398) -- (185,2.405) -- (189,2.405) -- (190,2.451) -- (191,2.595) -- (192,2.671) -- (193,2.702) -- (194,2.755) -- (195,2.777) -- (196,2.792) -- (197,2.808) -- (198,2.838) -- (199,2.853) -- (200,2.901);
\draw[cOpenAlig,line width=0.95pt,line join=round] (0,0.000) -- (2,0.000) -- (3,0.008) -- (4,0.326) -- (5,0.420) -- (6,0.535) -- (7,0.593) -- (8,0.639) -- (9,0.662) -- (10,0.682) -- (11,0.750) -- (12,0.768) -- (13,0.778) -- (14,0.785) -- (15,0.803) -- (16,0.811) -- (17,0.836) -- (18,0.843) -- (19,0.843) -- (20,0.854) -- (21,0.882) -- (22,0.890) -- (23,0.890) -- (24,0.897) -- (25,0.905) -- (26,0.938) -- (27,0.960) -- (28,0.998) -- (29,1.034) -- (30,1.079) -- (31,1.087) -- (32,1.114) -- (33,1.129) -- (34,1.163) -- (35,1.181) -- (36,1.203) -- (37,1.234) -- (38,1.242) -- (39,1.265) -- (40,1.295) -- (41,1.310) -- (42,1.333) -- (43,1.348) -- (44,1.356) -- (45,1.439) -- (46,1.477) -- (47,1.500) -- (48,1.515) -- (49,1.531) -- (50,1.581) -- (51,1.604) -- (52,1.627) -- (53,1.652) -- (54,1.685) -- (55,1.711) -- (56,1.751) -- (57,1.769) -- (58,1.802) -- (59,1.847) -- (60,1.855) -- (61,1.885) -- (62,1.918) -- (63,1.926) -- (64,1.933) -- (65,1.941) -- (66,1.949) -- (67,1.964) -- (68,1.989) -- (69,1.997) -- (70,2.007) -- (71,2.032) -- (72,2.040) -- (73,2.085) -- (74,2.125) -- (76,2.125) -- (77,2.135) -- (78,2.135) -- (79,2.143) -- (80,2.151) -- (81,2.179) -- (82,2.189) -- (83,2.189) -- (84,2.199) -- (85,2.207) -- (86,2.232) -- (87,2.232) -- (88,2.239) -- (89,2.239) -- (90,2.255) -- (91,2.272) -- (92,2.272) -- (93,2.280) -- (94,2.315) -- (95,2.340) -- (96,2.340) -- (97,2.348) -- (98,2.348) -- (99,2.369) -- (100,2.379) -- (101,2.379) -- (102,2.401) -- (105,2.401) -- (106,2.411) -- (108,2.411) -- (109,2.422) -- (110,2.432) -- (111,2.439) -- (112,2.447) -- (113,2.447) -- (114,2.454) -- (115,2.464) -- (116,2.480) -- (117,2.495) -- (118,2.525) -- (119,2.536) -- (120,2.536) -- (121,2.551) -- (123,2.551) -- (124,2.566) -- (125,2.566) -- (126,2.579) -- (127,2.579) -- (128,2.605) -- (129,2.612) -- (134,2.612) -- (135,2.628) -- (136,2.635) -- (137,2.643) -- (139,2.643) -- (140,2.650) -- (141,2.650) -- (142,2.694) -- (145,2.694) -- (146,2.709) -- (149,2.709) -- (150,2.717) -- (151,2.724) -- (152,2.724) -- (153,2.737) -- (154,2.747) -- (167,2.747) -- (168,2.755) -- (169,2.755) -- (170,2.762) -- (172,2.762) -- (173,2.770) -- (174,2.778) -- (177,2.778) -- (178,2.785) -- (183,2.785) -- (184,2.793) -- (185,2.800) -- (186,2.828) -- (187,2.828) -- (188,2.841) -- (189,2.841) -- (190,2.887) -- (191,3.089) -- (192,3.122) -- (193,3.160) -- (194,3.175) -- (195,3.191) -- (196,3.216) -- (197,3.262) -- (198,3.292) -- (199,3.302) -- (200,3.338);
\draw[cDiscDist,line width=0.95pt,line join=round] (0,0.000) -- (1,0.008) -- (3,0.008) -- (4,0.254) -- (5,0.368) -- (6,0.567) -- (7,0.636) -- (8,0.660) -- (9,0.691) -- (10,0.737) -- (11,0.752) -- (12,0.767) -- (13,0.775) -- (14,0.782) -- (15,0.805) -- (16,0.828) -- (17,0.835) -- (18,0.858) -- (19,0.881) -- (20,0.904) -- (23,0.904) -- (24,0.911) -- (25,0.941) -- (26,0.941) -- (27,0.972) -- (28,0.994) -- (29,1.002) -- (30,1.032) -- (31,1.071) -- (32,1.110) -- (33,1.140) -- (34,1.185) -- (35,1.216) -- (36,1.231) -- (37,1.253) -- (38,1.284) -- (39,1.322) -- (40,1.337) -- (41,1.360) -- (42,1.390) -- (43,1.405) -- (44,1.466) -- (45,1.481) -- (46,1.496) -- (47,1.519) -- (48,1.519) -- (49,1.565) -- (50,1.595) -- (51,1.633) -- (52,1.640) -- (53,1.671) -- (54,1.694) -- (55,1.724) -- (56,1.754) -- (57,1.778) -- (58,1.801) -- (59,1.809) -- (60,1.816) -- (61,1.877) -- (62,1.892) -- (63,1.892) -- (64,1.907) -- (65,1.953) -- (66,1.983) -- (67,2.006) -- (68,2.021) -- (69,2.044) -- (70,2.059) -- (71,2.067) -- (72,2.090) -- (73,2.120) -- (74,2.127) -- (75,2.143) -- (76,2.158) -- (77,2.188) -- (78,2.226) -- (79,2.233) -- (80,2.233) -- (81,2.256) -- (82,2.279) -- (83,2.287) -- (84,2.310) -- (85,2.310) -- (86,2.340) -- (87,2.378) -- (88,2.386) -- (89,2.393) -- (90,2.409) -- (91,2.456) -- (92,2.479) -- (93,2.495) -- (94,2.510) -- (95,2.518) -- (96,2.525) -- (97,2.533) -- (98,2.556) -- (99,2.572) -- (100,2.594) -- (101,2.634) -- (102,2.649) -- (103,2.672) -- (104,2.702) -- (105,2.710) -- (106,2.733) -- (107,2.756) -- (108,2.771) -- (109,2.771) -- (110,2.809) -- (111,2.809) -- (112,2.824) -- (113,2.824) -- (114,2.840) -- (115,2.840) -- (116,2.848) -- (117,2.886) -- (118,2.902) -- (119,2.917) -- (120,2.924) -- (121,2.939) -- (122,2.955) -- (123,2.955) -- (124,2.970) -- (125,2.970) -- (126,2.977) -- (127,2.985) -- (128,2.992) -- (129,3.008) -- (130,3.008) -- (131,3.031) -- (132,3.039) -- (133,3.046) -- (134,3.061) -- (135,3.077) -- (136,3.085) -- (137,3.101) -- (138,3.101) -- (139,3.108) -- (140,3.124) -- (141,3.132) -- (142,3.132) -- (143,3.139) -- (145,3.139) -- (146,3.154) -- (147,3.154) -- (148,3.169) -- (149,3.200) -- (150,3.224) -- (151,3.239) -- (152,3.247) -- (153,3.255) -- (154,3.270) -- (157,3.270) -- (158,3.285) -- (159,3.293) -- (166,3.293) -- (167,3.315) -- (168,3.323) -- (169,3.346) -- (170,3.346) -- (171,3.354) -- (177,3.354) -- (178,3.369) -- (179,3.377) -- (188,3.377) -- (189,3.384) -- (190,3.452) -- (191,3.498) -- (192,3.566) -- (193,3.604) -- (194,3.649) -- (195,3.710) -- (196,3.718) -- (197,3.733) -- (198,3.740) -- (199,3.748) -- (200,3.773);
\draw[cDiscAlig,line width=0.95pt,line join=round] (0,0.000) -- (3,0.000) -- (4,0.189) -- (5,0.333) -- (6,0.502) -- (7,0.533) -- (8,0.549) -- (9,0.588) -- (10,0.619) -- (11,0.680) -- (12,0.704) -- (13,0.727) -- (14,0.742) -- (15,0.749) -- (16,0.764) -- (17,0.780) -- (18,0.802) -- (19,0.825) -- (20,0.833) -- (21,0.833) -- (22,0.840) -- (23,0.848) -- (24,0.871) -- (25,0.916) -- (26,0.946) -- (27,0.999) -- (28,1.037) -- (29,1.067) -- (30,1.136) -- (31,1.175) -- (32,1.205) -- (33,1.236) -- (34,1.251) -- (35,1.312) -- (36,1.327) -- (37,1.395) -- (38,1.456) -- (39,1.501) -- (40,1.517) -- (41,1.547) -- (42,1.585) -- (43,1.607) -- (44,1.653) -- (45,1.767) -- (46,1.782) -- (47,1.812) -- (48,1.835) -- (49,1.850) -- (50,1.873) -- (51,1.963) -- (52,1.994) -- (53,2.039) -- (54,2.047) -- (55,2.115) -- (56,2.160) -- (57,2.198) -- (58,2.229) -- (59,2.267) -- (60,2.304) -- (61,2.350) -- (62,2.350) -- (63,2.403) -- (64,2.448) -- (65,2.479) -- (66,2.509) -- (67,2.524) -- (68,2.547) -- (69,2.577) -- (70,2.592) -- (71,2.615) -- (72,2.638) -- (73,2.676) -- (74,2.706) -- (75,2.706) -- (76,2.729) -- (77,2.759) -- (78,2.782) -- (79,2.789) -- (80,2.812) -- (81,2.820) -- (82,2.835) -- (83,2.850) -- (84,2.880) -- (85,2.888) -- (86,2.910) -- (87,2.933) -- (88,2.948) -- (89,2.986) -- (90,3.024) -- (91,3.054) -- (92,3.085) -- (93,3.092) -- (94,3.100) -- (95,3.123) -- (96,3.145) -- (97,3.160) -- (98,3.168) -- (99,3.176) -- (100,3.183) -- (101,3.198) -- (102,3.213) -- (103,3.244) -- (104,3.267) -- (105,3.289) -- (106,3.327) -- (107,3.327) -- (108,3.342) -- (110,3.342) -- (111,3.365) -- (112,3.380) -- (113,3.395) -- (114,3.410) -- (115,3.418) -- (116,3.426) -- (117,3.448) -- (118,3.471) -- (119,3.486) -- (120,3.501) -- (121,3.547) -- (122,3.562) -- (123,3.570) -- (124,3.577) -- (125,3.577) -- (126,3.585) -- (127,3.600) -- (128,3.630) -- (129,3.645) -- (130,3.653) -- (131,3.676) -- (133,3.676) -- (134,3.691) -- (135,3.698) -- (136,3.706) -- (137,3.713) -- (138,3.729) -- (139,3.729) -- (140,3.744) -- (141,3.759) -- (142,3.767) -- (143,3.782) -- (144,3.789) -- (145,3.804) -- (146,3.812) -- (148,3.812) -- (149,3.820) -- (150,3.827) -- (151,3.835) -- (156,3.835) -- (157,3.850) -- (161,3.850) -- (162,3.865) -- (163,3.873) -- (165,3.873) -- (166,3.880) -- (167,3.880) -- (168,3.888) -- (173,3.888) -- (174,3.895) -- (183,3.895) -- (184,3.903) -- (185,3.910) -- (189,3.910) -- (190,3.963) -- (191,4.085) -- (192,4.169) -- (193,4.207) -- (194,4.222) -- (195,4.282) -- (196,4.298) -- (197,4.335) -- (198,4.343) -- (199,4.343) -- (200,4.388);
\draw[wInk!85,line width=0.8pt,dash pattern=on 2.4pt off 1.4pt] (190,0) -- (190,8);
\draw[wInk!85,line width=0.8pt] (190,8) -- ++(0,7.2pt);
\node[fill=wInk!85,text=white,font=\tiny,inner xsep=1.8pt,inner ysep=1pt,anchor=north east] at (190,8) [yshift=7.2pt] {terminal intervention};
\draw[cOpenDist,line width=0.95pt] (0,0) ++(0,-25.5pt) -- ++(10pt,0) node[right,inner sep=1.5pt,text=black] {\pix{open}Open\,$\times$\,\pix{distributed}Distributed};
\draw[cOpenAlig,line width=0.95pt] (0,0) ++(0,-34.1pt) -- ++(10pt,0) node[right,inner sep=1.5pt,text=black] {\pix{open}Open\,$\times$\,\pix{aligned}Aligned};
\draw[cDiscDist,line width=0.95pt] (96,0) ++(0,-25.5pt) -- ++(10pt,0) node[right,inner sep=1.5pt,text=black] {\pix{disclosed}Disclosed\,$\times$\,\pix{distributed}Distributed};
\draw[cDiscAlig,line width=0.95pt] (96,0) ++(0,-34.1pt) -- ++(10pt,0) node[right,inner sep=1.5pt,text=black] {\pix{disclosed}Disclosed\,$\times$\,\pix{aligned}Aligned};
\draw[black!70,line width=0.95pt] (0,0) ++(0,-42.7pt) -- ++(10pt,0) node[right,inner sep=1.5pt,text=black] {grounded blocks};
\draw[black!70,line width=0.7pt,densely dotted] (96,0) ++(0,-42.7pt) -- ++(10pt,0) node[right,inner sep=1.5pt,text=black] {evidence held};
\end{tikzpicture}
\hfill
\begin{tikzpicture}[x=0.9300pt,y=5.1667pt,font=\scriptsize]
\node[anchor=east,inner sep=1.5pt] at (0,0) {0};
\draw[black!10,line width=0.3pt] (0,2) -- (200,2);
\node[anchor=east,inner sep=1.5pt] at (0,2) {2};
\draw[black!10,line width=0.3pt] (0,4) -- (200,4);
\node[anchor=east,inner sep=1.5pt] at (0,4) {4};
\draw[black!10,line width=0.3pt] (0,6) -- (200,6);
\node[anchor=east,inner sep=1.5pt] at (0,6) {6};
\draw[black!10,line width=0.3pt] (0,8) -- (200,8);
\node[anchor=east,inner sep=1.5pt] at (0,8) {8};
\draw[black!10,line width=0.3pt] (0,10) -- (200,10);
\node[anchor=east,inner sep=1.5pt] at (0,10) {10};
\node[anchor=east,inner sep=1.5pt] at (0,12) {12};
\draw[black!60,line width=0.4pt] (0,0) -- ++(0,-2pt) node[below,inner sep=1.5pt] {0};
\draw[black!60,line width=0.4pt] (50,0) -- ++(0,-2pt) node[below,inner sep=1.5pt] {50};
\draw[black!60,line width=0.4pt] (100,0) -- ++(0,-2pt) node[below,inner sep=1.5pt] {100};
\draw[black!60,line width=0.4pt] (150,0) -- ++(0,-2pt) node[below,inner sep=1.5pt] {150};
\draw[black!60,line width=0.4pt] (200,0) -- ++(0,-2pt) node[below,inner sep=1.5pt] {200};
\draw[black!45,line width=0.4pt,dash pattern=on 1.6pt off 1.2pt] (0,12) -- (200,12) -- (200,0);
\draw[black!60,line width=0.4pt] (0,12) -- (0,0) -- (200,0);
\node[anchor=north] at (100,0) [yshift=-9pt] {paid actions};
\node[rotate=90,anchor=south] at (0,6) [xshift=0pt,yshift=7pt] {blocks};
\node[anchor=south west,inner sep=0pt,font=\footnotesize\bfseries] at (0,12) [xshift=-18pt,yshift=3.5pt] {(b)};
\draw[wInk!80,line width=0.55pt,densely dotted] (0,0.000) -- (2,0.000) -- (3,0.396) -- (4,0.511) -- (5,0.867) -- (6,1.101) -- (7,1.304) -- (8,1.449) -- (9,1.518) -- (10,1.670) -- (11,1.768) -- (12,1.843) -- (13,1.906) -- (14,2.011) -- (15,2.167) -- (16,2.207) -- (17,2.281) -- (18,2.339) -- (19,2.396) -- (20,2.464) -- (21,2.528) -- (22,2.557) -- (23,2.598) -- (24,2.655) -- (25,2.701) -- (26,2.770) -- (27,2.874) -- (28,2.967) -- (29,3.026) -- (30,3.096) -- (31,3.223) -- (32,3.292) -- (33,3.326) -- (34,3.355) -- (35,3.383) -- (36,3.418) -- (37,3.441) -- (38,3.499) -- (39,3.534) -- (40,3.575) -- (41,3.609) -- (42,3.633) -- (43,3.668) -- (44,3.710) -- (45,3.774) -- (46,3.797) -- (47,3.831) -- (48,3.849) -- (49,3.866) -- (50,3.894) -- (51,3.936) -- (52,3.965) -- (53,4.007) -- (54,4.053) -- (55,4.122) -- (56,4.152) -- (57,4.175) -- (58,4.206) -- (59,4.235) -- (60,4.253) -- (61,4.294) -- (62,4.317) -- (63,4.364) -- (64,4.376) -- (65,4.387) -- (66,4.417) -- (67,4.434) -- (68,4.464) -- (69,4.498) -- (70,4.505) -- (71,4.517) -- (72,4.547) -- (73,4.565) -- (74,4.570) -- (75,4.582) -- (76,4.611) -- (77,4.646) -- (78,4.671) -- (79,4.689) -- (80,4.695) -- (81,4.695) -- (82,4.712) -- (83,4.724) -- (84,4.759) -- (85,4.765) -- (86,4.771) -- (87,4.777) -- (88,4.777) -- (89,4.794) -- (90,4.817) -- (91,4.817) -- (92,4.834) -- (93,4.864) -- (94,4.894) -- (95,4.899) -- (96,4.912) -- (97,4.912) -- (98,4.925) -- (99,4.930) -- (100,4.953) -- (101,4.959) -- (102,4.959) -- (103,4.965) -- (104,4.982) -- (105,4.988) -- (106,4.994) -- (107,5.000) -- (108,5.023) -- (109,5.023) -- (110,5.034) -- (111,5.046) -- (112,5.052) -- (113,5.057) -- (114,5.063) -- (116,5.063) -- (117,5.080) -- (119,5.080) -- (120,5.086) -- (121,5.097) -- (122,5.114) -- (123,5.126) -- (124,5.131) -- (125,5.143) -- (126,5.172) -- (127,5.172) -- (128,5.178) -- (129,5.195) -- (130,5.202) -- (131,5.214) -- (132,5.219) -- (133,5.219) -- (134,5.226) -- (135,5.232) -- (136,5.232) -- (137,5.250) -- (139,5.250) -- (140,5.255) -- (141,5.267) -- (142,5.267) -- (143,5.273) -- (149,5.273) -- (150,5.285) -- (151,5.291) -- (152,5.297) -- (153,5.302) -- (154,5.308) -- (155,5.314) -- (156,5.320) -- (159,5.320) -- (160,5.326) -- (161,5.326) -- (162,5.331) -- (164,5.331) -- (165,5.337) -- (175,5.337) -- (176,5.343) -- (177,5.343) -- (178,5.354) -- (180,5.354) -- (181,5.366) -- (184,5.366) -- (185,5.371) -- (186,5.377) -- (187,5.377) -- (188,5.388) -- (189,5.388) -- (190,5.400) -- (191,5.411) -- (192,5.423) -- (193,5.440) -- (194,5.446) -- (195,5.457) -- (196,5.463) -- (197,5.475) -- (200,5.475);
\draw[wBrick!85!black,line width=0.55pt,densely dotted] (0,0.000) -- (2,0.000) -- (3,0.688) -- (4,0.688) -- (5,0.812) -- (6,1.188) -- (7,1.312) -- (9,1.312) -- (10,1.375) -- (11,1.562) -- (12,1.750) -- (13,1.875) -- (14,2.062) -- (15,2.188) -- (17,2.188) -- (18,2.250) -- (19,2.312) -- (21,2.312) -- (22,2.438) -- (25,2.438) -- (26,2.750) -- (27,3.000) -- (28,3.000) -- (29,3.188) -- (30,3.375) -- (31,3.375) -- (32,3.562) -- (33,3.688) -- (34,3.750) -- (35,3.750) -- (36,3.875) -- (37,3.875) -- (38,4.000) -- (39,4.000) -- (40,4.062) -- (42,4.062) -- (43,4.125) -- (44,4.250) -- (45,4.375) -- (46,4.562) -- (47,4.562) -- (48,4.625) -- (49,4.812) -- (50,4.812) -- (51,4.875) -- (52,4.938) -- (55,4.938) -- (56,5.125) -- (57,5.250) -- (59,5.250) -- (60,5.312) -- (61,5.312) -- (62,5.375) -- (63,5.438) -- (65,5.438) -- (66,5.500) -- (67,5.688) -- (68,5.688) -- (69,5.750) -- (73,5.750) -- (74,5.812) -- (75,5.812) -- (76,5.875) -- (79,5.875) -- (80,6.000) -- (81,6.000) -- (82,6.125) -- (83,6.188) -- (85,6.188) -- (86,6.250) -- (90,6.250) -- (91,6.375) -- (92,6.562) -- (95,6.562) -- (96,6.625) -- (97,6.688) -- (99,6.688) -- (100,6.750) -- (107,6.750) -- (108,6.812) -- (109,6.812) -- (110,6.938) -- (122,6.938) -- (123,7.125) -- (125,7.125) -- (126,7.188) -- (127,7.250) -- (132,7.250) -- (133,7.312) -- (143,7.312) -- (144,7.438) -- (145,7.438) -- (146,7.500) -- (152,7.500) -- (153,7.625) -- (154,7.688) -- (155,7.688) -- (156,7.812) -- (157,7.938) -- (158,8.000) -- (160,8.000) -- (161,8.062) -- (163,8.062) -- (164,8.188) -- (166,8.188) -- (167,8.375) -- (177,8.375) -- (178,8.438) -- (180,8.438) -- (181,8.500) -- (188,8.500) -- (189,8.562) -- (197,8.562) -- (198,8.625) -- (199,8.688) -- (200,8.688);
\draw[wTeal!70!black,line width=0.55pt,densely dotted] (0,0.000) -- (2,0.000) -- (3,1.000) -- (5,1.000) -- (6,1.500) -- (8,1.500) -- (9,1.625) -- (10,1.875) -- (11,2.062) -- (12,2.125) -- (13,2.375) -- (14,2.375) -- (15,2.438) -- (16,2.562) -- (21,2.562) -- (22,2.625) -- (24,2.625) -- (25,2.812) -- (26,3.312) -- (27,3.438) -- (28,3.688) -- (29,3.938) -- (30,4.000) -- (32,4.000) -- (33,4.125) -- (34,4.375) -- (35,4.375) -- (36,4.500) -- (37,4.562) -- (38,4.562) -- (39,4.625) -- (40,4.625) -- (41,4.688) -- (42,4.750) -- (43,4.750) -- (44,4.812) -- (45,4.938) -- (46,5.062) -- (47,5.125) -- (48,5.312) -- (49,5.688) -- (52,5.688) -- (53,5.875) -- (54,5.938) -- (55,6.062) -- (56,6.250) -- (58,6.250) -- (59,6.375) -- (60,6.500) -- (61,6.500) -- (62,6.625) -- (63,6.688) -- (66,6.688) -- (67,6.750) -- (70,6.750) -- (71,6.812) -- (72,6.938) -- (73,7.000) -- (77,7.000) -- (78,7.062) -- (79,7.125) -- (80,7.188) -- (89,7.188) -- (90,7.250) -- (91,7.312) -- (92,7.375) -- (93,7.375) -- (94,7.438) -- (95,7.750) -- (98,7.750) -- (99,7.812) -- (101,7.812) -- (102,8.000) -- (103,8.062) -- (106,8.062) -- (107,8.188) -- (109,8.188) -- (110,8.250) -- (111,8.312) -- (112,8.375) -- (113,8.438) -- (119,8.438) -- (120,8.562) -- (124,8.562) -- (125,8.625) -- (127,8.625) -- (128,8.688) -- (129,8.750) -- (130,8.812) -- (131,8.812) -- (132,8.875) -- (133,8.938) -- (150,8.938) -- (151,9.000) -- (154,9.000) -- (155,9.062) -- (156,9.125) -- (160,9.125) -- (161,9.250) -- (163,9.250) -- (164,9.375) -- (173,9.375) -- (174,9.438) -- (175,9.438) -- (176,9.500) -- (182,9.500) -- (183,9.562) -- (185,9.562) -- (186,9.625) -- (189,9.625) -- (190,9.688) -- (191,9.750) -- (193,9.750) -- (194,9.812) -- (195,9.875) -- (196,9.938) -- (197,10.000) -- (198,10.000) -- (199,10.062) -- (200,10.062);
\draw[wInk!80,line width=0.95pt,line join=round] (0,0.000) -- (2,0.000) -- (3,0.006) -- (4,0.234) -- (5,0.325) -- (6,0.491) -- (7,0.519) -- (8,0.584) -- (9,0.612) -- (10,0.635) -- (11,0.676) -- (12,0.699) -- (13,0.711) -- (14,0.722) -- (15,0.740) -- (16,0.757) -- (17,0.763) -- (18,0.780) -- (19,0.798) -- (20,0.822) -- (21,0.822) -- (22,0.833) -- (23,0.839) -- (24,0.850) -- (25,0.874) -- (26,0.897) -- (27,0.949) -- (28,0.960) -- (29,0.966) -- (30,1.046) -- (31,1.110) -- (32,1.150) -- (33,1.174) -- (34,1.225) -- (35,1.248) -- (36,1.265) -- (37,1.305) -- (38,1.329) -- (39,1.369) -- (40,1.381) -- (41,1.398) -- (42,1.438) -- (43,1.449) -- (44,1.472) -- (45,1.523) -- (46,1.540) -- (47,1.574) -- (48,1.585) -- (49,1.608) -- (50,1.631) -- (51,1.660) -- (52,1.683) -- (53,1.705) -- (54,1.723) -- (55,1.763) -- (56,1.805) -- (57,1.829) -- (58,1.841) -- (59,1.870) -- (60,1.893) -- (61,1.917) -- (62,1.917) -- (63,1.947) -- (64,1.981) -- (65,2.009) -- (66,2.021) -- (67,2.044) -- (68,2.078) -- (69,2.078) -- (70,2.097) -- (71,2.120) -- (72,2.144) -- (73,2.167) -- (74,2.195) -- (75,2.201) -- (76,2.213) -- (77,2.231) -- (78,2.237) -- (79,2.255) -- (80,2.260) -- (81,2.272) -- (82,2.295) -- (83,2.295) -- (84,2.307) -- (85,2.313) -- (86,2.331) -- (87,2.337) -- (88,2.349) -- (89,2.372) -- (90,2.390) -- (91,2.408) -- (92,2.426) -- (93,2.437) -- (94,2.460) -- (95,2.477) -- (96,2.484) -- (97,2.497) -- (98,2.503) -- (99,2.509) -- (100,2.527) -- (101,2.538) -- (102,2.544) -- (103,2.549) -- (104,2.566) -- (105,2.578) -- (106,2.590) -- (107,2.601) -- (108,2.613) -- (109,2.625) -- (110,2.660) -- (111,2.660) -- (112,2.665) -- (115,2.665) -- (116,2.677) -- (117,2.694) -- (118,2.719) -- (119,2.724) -- (120,2.730) -- (121,2.742) -- (122,2.759) -- (123,2.764) -- (124,2.770) -- (125,2.770) -- (126,2.782) -- (127,2.782) -- (128,2.807) -- (129,2.818) -- (130,2.818) -- (131,2.830) -- (132,2.836) -- (133,2.836) -- (134,2.843) -- (135,2.849) -- (136,2.855) -- (137,2.862) -- (138,2.862) -- (139,2.867) -- (140,2.879) -- (141,2.891) -- (142,2.902) -- (143,2.914) -- (145,2.914) -- (146,2.920) -- (147,2.925) -- (148,2.925) -- (149,2.948) -- (150,2.966) -- (151,2.971) -- (153,2.971) -- (154,2.983) -- (155,2.983) -- (156,2.989) -- (157,2.995) -- (158,3.007) -- (159,3.012) -- (161,3.012) -- (162,3.024) -- (163,3.029) -- (164,3.029) -- (165,3.035) -- (166,3.048) -- (168,3.048) -- (169,3.054) -- (173,3.054) -- (174,3.060) -- (175,3.060) -- (176,3.065) -- (177,3.065) -- (178,3.077) -- (180,3.077) -- (181,3.082) -- (183,3.082) -- (184,3.088) -- (185,3.094) -- (186,3.106) -- (189,3.106) -- (190,3.163) -- (191,3.254) -- (192,3.317) -- (193,3.345) -- (194,3.396) -- (195,3.431) -- (196,3.436) -- (197,3.488) -- (198,3.505) -- (199,3.505) -- (200,3.530);
\draw[wBrick!85!black,line width=0.95pt,line join=round] (0,0.000) -- (3,0.000) -- (4,0.625) -- (14,0.625) -- (15,0.688) -- (18,0.688) -- (19,0.750) -- (20,0.812) -- (23,0.812) -- (24,0.875) -- (25,0.875) -- (26,1.000) -- (28,1.000) -- (29,1.062) -- (30,1.125) -- (32,1.125) -- (33,1.188) -- (35,1.188) -- (36,1.250) -- (39,1.250) -- (40,1.312) -- (41,1.375) -- (43,1.375) -- (44,1.438) -- (45,1.500) -- (47,1.500) -- (48,1.562) -- (49,1.688) -- (50,1.812) -- (51,1.875) -- (52,1.875) -- (53,1.938) -- (55,1.938) -- (56,2.000) -- (61,2.000) -- (62,2.062) -- (63,2.125) -- (65,2.125) -- (66,2.188) -- (67,2.188) -- (68,2.250) -- (73,2.250) -- (74,2.438) -- (75,2.438) -- (76,2.500) -- (81,2.500) -- (82,2.562) -- (83,2.625) -- (84,2.625) -- (85,2.688) -- (86,2.750) -- (89,2.750) -- (90,2.812) -- (91,2.938) -- (92,3.125) -- (95,3.125) -- (96,3.188) -- (97,3.250) -- (98,3.312) -- (99,3.312) -- (100,3.375) -- (103,3.375) -- (104,3.438) -- (107,3.438) -- (108,3.500) -- (109,3.562) -- (110,3.625) -- (111,3.688) -- (122,3.688) -- (123,3.812) -- (128,3.812) -- (129,3.875) -- (130,3.938) -- (131,4.000) -- (139,4.000) -- (140,4.125) -- (141,4.188) -- (142,4.188) -- (143,4.250) -- (146,4.250) -- (147,4.312) -- (152,4.312) -- (153,4.438) -- (154,4.500) -- (155,4.625) -- (156,4.688) -- (157,4.812) -- (160,4.812) -- (161,4.875) -- (165,4.875) -- (166,5.000) -- (167,5.125) -- (172,5.125) -- (173,5.188) -- (176,5.188) -- (177,5.312) -- (178,5.375) -- (184,5.375) -- (185,5.625) -- (188,5.625) -- (189,5.688) -- (190,5.688) -- (191,6.000) -- (192,6.312) -- (193,6.312) -- (194,6.438) -- (195,6.562) -- (196,6.625) -- (197,6.688) -- (198,6.750) -- (199,6.938) -- (200,7.625);
\draw[wTeal!70!black,line width=0.95pt,line join=round] (0,0.000) -- (3,0.000) -- (4,0.750) -- (5,0.812) -- (20,0.812) -- (21,0.938) -- (22,1.062) -- (24,1.062) -- (25,1.125) -- (44,1.125) -- (45,1.188) -- (46,1.250) -- (48,1.250) -- (49,1.312) -- (50,1.688) -- (53,1.688) -- (54,1.750) -- (55,1.875) -- (56,2.000) -- (57,2.062) -- (58,2.062) -- (59,2.125) -- (60,2.188) -- (63,2.188) -- (64,2.250) -- (66,2.250) -- (67,2.312) -- (68,2.375) -- (77,2.375) -- (78,2.438) -- (79,2.438) -- (80,2.500) -- (81,2.562) -- (90,2.562) -- (91,2.625) -- (92,2.688) -- (101,2.688) -- (102,2.750) -- (103,2.812) -- (114,2.812) -- (115,2.938) -- (116,2.938) -- (117,3.000) -- (118,3.125) -- (119,3.188) -- (120,3.188) -- (121,3.250) -- (122,3.312) -- (124,3.312) -- (125,3.375) -- (127,3.375) -- (128,3.438) -- (129,3.438) -- (130,3.500) -- (134,3.500) -- (135,3.562) -- (136,3.625) -- (145,3.625) -- (146,3.688) -- (148,3.688) -- (149,3.750) -- (150,3.812) -- (152,3.812) -- (153,3.938) -- (155,3.938) -- (156,4.000) -- (160,4.000) -- (161,4.062) -- (163,4.062) -- (164,4.125) -- (171,4.125) -- (172,4.188) -- (174,4.188) -- (175,4.250) -- (185,4.250) -- (186,4.375) -- (187,4.438) -- (188,4.438) -- (189,4.500) -- (190,4.562) -- (191,5.188) -- (193,5.188) -- (194,5.375) -- (195,5.438) -- (196,5.625) -- (197,5.812) -- (198,5.938) -- (199,6.312) -- (200,7.062);
\filldraw[fill=wBrick!85!black,draw=white,line width=0.3pt] (20,0.812) ++(0,1.9pt) -- ++(1.9pt,-1.9pt) -- ++(-1.9pt,-1.9pt) -- ++(-1.9pt,1.9pt) -- cycle;
\filldraw[fill=wBrick!85!black,draw=white,line width=0.3pt] (60,2.000) ++(0,1.9pt) -- ++(1.9pt,-1.9pt) -- ++(-1.9pt,-1.9pt) -- ++(-1.9pt,1.9pt) -- cycle;
\filldraw[fill=wBrick!85!black,draw=white,line width=0.3pt] (100,3.375) ++(0,1.9pt) -- ++(1.9pt,-1.9pt) -- ++(-1.9pt,-1.9pt) -- ++(-1.9pt,1.9pt) -- cycle;
\filldraw[fill=wBrick!85!black,draw=white,line width=0.3pt] (140,4.125) ++(0,1.9pt) -- ++(1.9pt,-1.9pt) -- ++(-1.9pt,-1.9pt) -- ++(-1.9pt,1.9pt) -- cycle;
\filldraw[fill=wBrick!85!black,draw=white,line width=0.3pt] (180,5.375) ++(0,1.9pt) -- ++(1.9pt,-1.9pt) -- ++(-1.9pt,-1.9pt) -- ++(-1.9pt,1.9pt) -- cycle;
\filldraw[fill=wTeal!70!black,draw=white,line width=0.3pt] (40,1.125) ++(0,2.2pt) -- ++(1.95pt,-3.35pt) -- ++(-3.9pt,0) -- cycle;
\filldraw[fill=wTeal!70!black,draw=white,line width=0.3pt] (80,2.500) ++(0,2.2pt) -- ++(1.95pt,-3.35pt) -- ++(-3.9pt,0) -- cycle;
\filldraw[fill=wTeal!70!black,draw=white,line width=0.3pt] (120,3.188) ++(0,2.2pt) -- ++(1.95pt,-3.35pt) -- ++(-3.9pt,0) -- cycle;
\filldraw[fill=wTeal!70!black,draw=white,line width=0.3pt] (160,4.000) ++(0,2.2pt) -- ++(1.95pt,-3.35pt) -- ++(-3.9pt,0) -- cycle;
\draw[wInk!85,line width=0.8pt,dash pattern=on 2.4pt off 1.4pt] (190,0) -- (190,12);
\draw[wInk!85,line width=0.8pt] (190,12) -- ++(0,7.2pt);
\node[fill=wInk!85,text=white,font=\tiny,inner xsep=1.8pt,inner ysep=1pt,anchor=north east] at (190,12) [yshift=7.2pt] {terminal intervention};
\draw[wInk!80,line width=0.95pt] (0,0) ++(0,-25.5pt) -- ++(10pt,0) node[right,inner sep=1.5pt,text=black] {baseline loop, eleven models};
\draw[wBrick!85!black,line width=0.95pt] (0,0) ++(0,-34.1pt) -- ++(10pt,0) node[right,inner sep=1.5pt,text=black] {\logoc{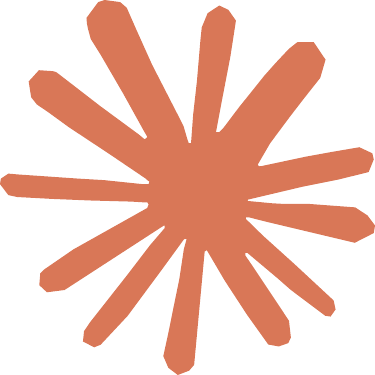}Claude Fable 5.1-xhigh, \textit{Claude Code}};
\filldraw[fill=wBrick!85!black,draw=white,line width=0.3pt] (0,0) ++(0,-34.1pt) ++(5pt,0) ++(0,1.9pt) -- ++(1.9pt,-1.9pt) -- ++(-1.9pt,-1.9pt) -- ++(-1.9pt,1.9pt) -- cycle;
\draw[wTeal!70!black,line width=0.95pt] (0,0) ++(0,-42.7pt) -- ++(10pt,0) node[right,inner sep=1.5pt,text=black] {\logoc{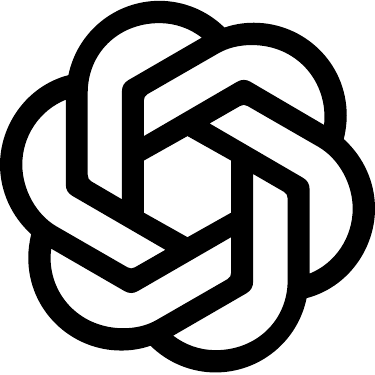}GPT-6 Astra-xhigh, \textit{Codex}};
\filldraw[fill=wTeal!70!black,draw=white,line width=0.3pt] (0,0) ++(0,-42.7pt) ++(5pt,0) ++(0,2.2pt) -- ++(1.95pt,-3.35pt) -- ++(-3.9pt,0) -- cycle;
\end{tikzpicture}
\caption{Discovery over the episode. (a)~The four factorial conditions in the baseline loop. (b)~The baseline loop and the two vendor-harness systems on the same four instances, pooling conditions. Solid lines give grounded blocks and dotted lines the blocks whose evidence the agent holds, each a mean with every model weighted equally, and the flag marks the posting of the terminal intervention.}
\label{fig:curves}
\end{figure*}

%% file: sections/5_results.tex
\section{Results}
\label{sec:results}

\input{tables/tab_main}

\subsection{Rejecting the doctrine does not yield its replacement}
\label{sec:main-results}

Table~\ref{tab:main} gives the four metrics by model and condition. Between evidence and
transfer, discovery in the baseline loop thins at every step. Evidence coverage is broad
across episodes and incomplete within each: the Open episodes together reach every block,
and none holds more than \resCoverOpenHC{} of the five sources of HC. Inference trails evidence: agents in the
primary condition hold the evidence of \resEvidenceOD{} blocks and ground \resBlocksOD{}.
Most episodes stop with a source of every hyperedge missing.
Where an episode holds two sources, the third is a consign block in
\resLoopMissingConsign\% of cases, CO2 for H2 and CO1 for H3, and H1, which has no consign
source, most often lacks FE1 (Appendix~\ref{app:heat}). Other episodes complete the sources
and leave the change unstated: \resReachThreeN{} of \resScoredMain{} hold the complete
evidence of a hyperedge, and \resEvidencedEpisodes{} of them state its change
(Table~\ref{tab:evidenced}). Complete evidence accompanies the retraction more readily than
a constructive change: in the Open conditions of the main comparison, every episode with the
complete evidence of H1 states RC1, and none with the complete evidence of H2 or H3 states
the hidden label or the rewritten variables. At 200 paid actions, the exact prediction is
made in none.

In the Open conditions the notes retract
the taught axiom in \resStatedRCOneOpen\% of a model's episodes on average, and no episode
postulates the hidden label or rewrites the variables. The same failure arises from
different processes. On one instance GPT-5.6 Sol requests \resCaseSolInterventions{}
interventions, corrects a confound in its own notes and then extends its account past what
its contrasts establish, an explanation that stays inadequate after testing, while Gemini
3.6 Flash requests none and turns the timing of the stall into a threshold it leaves
untested, an explanation without a test (Appendix~\ref{app:qualitative}).

\IfFileExists{figures/fig_scaffolds.tex}{\input{figures/fig_scaffolds}}{}

\subsection{A supplied representation changes inquiry more than discovery}
\label{sec:effects}

With the evidence distributed, disclosure raises intervention requests from \resInterventionsOD{} to
\resInterventionsDD{} per episode. Averaged over the two evidence conditions, it adds about
one grounded block (\drepr{} $=$ \resDreprBlocks{}, \resDreprBlocksCI{}, unchanged when the
grounding verdict is required in every condition, Appendix~\ref{app:composition}). Disclosed
notes state \resRcStatedMeanDisc{} of
the three changes on average, and \resReachFullDiscN{} of \resStatedDiscN{} Disclosed
episodes hold the complete evidence of any hyperedge.

Alignment lowers the cost of access. With the transcripts the evidence held rises by
\resDalignEvidence{} blocks (\resDalignEvidenceCI{}) and part of it is read, \dalign{} $=$
\resDalignBlocks{} on Blocks (\resDalignBlocksCI{}) and \resDalignXC{} on XC, while
intervention requests stay where they were (\resDalignInterventions{},
\resDalignInterventionsCI{}). Collocated records still have to
be composed, and the gain stays below disclosure's. The estimated gains are approximately
additive (interaction \resInterBlocks{}, \resInterBlocksCI{}).

The curves of Figure~\ref{fig:curves} separate when evidence is gathered from when it
enters the account: in the baseline loop both rise early and flatten well before the budget
ends, with \resSharePreWindow\% of grounded blocks grounded before the terminal
intervention is posted (panel a), while the vendor-harness systems keep acquiring evidence
after the loop's curves flatten and ground \resVendorSharePreWindow\% of their blocks before
it (panel b).

\subsection{Scaffolding increases testing more readily than discovery}
\label{sec:scaffolds}

Figure~\ref{fig:scaffolds} compares the three scaffolds with the baseline loop for five
models in the two Open conditions. Pooled over both, each moves Blocks by less than half a
block: \resScafDiffCrit{} for self-critique (\resScafDiffCritCI{}), \resScafDiffFals{} for
falsification (\resScafDiffFalsCI{}) and \resScafDiffEvol{} for evolution
(\resScafDiffEvolCI{}). Falsification and evolution yield gains as small as periodic
self-critique with about twenty times its scaffold calls
(Appendix~\ref{app:harness-parity}). Falsification changes how agents inquire. Episodes
with no intervention request fall from \resScafZeroBase{} of \resScafCountedBase{} under
the baseline to \resScafZeroFals{} of \resScafCountedFals{}, while self-critique and
evolution stay near the baseline's rate. A scaffold that asks for a test obtains one. Of
\resScafArmCounted{} scaffolded episodes, \resScafReachFull{} hold the complete evidence of
a hyperedge and \resScafReachFullStated{} states its change.

\subsection{Vendor-harness systems extend discovery}
\label{sec:trajectories}

The vendor-harness systems broaden discovery into further domains and sometimes rewrite
the variables, while the hidden label remains unstated in their Open episodes. They ground \resFableBlocks{} (Claude Fable~5.1) and \resAstraBlocks{} (GPT-6
Astra) blocks per episode, against \resLoopBlocks{} for the eleven models on the same
instances (Appendix~\ref{app:agentic}). A fen block, grounded in \resShareDomainFen\% of the loop's
episodes over the whole main comparison, is grounded in \resFableDomainFen{} and \resAstraDomainFen{} of their
\resAgenticPlanned{} episodes each, and Claude Fable~5.1 rewrites the variables in
\resFableStatedRCThreeOpenN{} of its \resFableStatedOpenN{} Open episodes. The consign
domain stays sparse, with a grounded consign block in \resFableDomainConsign{} and
\resAstraDomainConsign{} episodes, and no episode makes the exact prediction at
200 paid actions. One Open episode of Claude Fable~5.1 states the rewritten variables with
its evidence complete and commits two of the three readings exactly, the third off by one
with the true value listed among its risks (Appendix~\ref{app:qual-fable}).

On one instance, in all four conditions, the vendor-harness systems play \resExtBudgetVendor{} paid actions and GPT-5.6 Sol in the
baseline loop \resExtBudgetLoop{} (Appendix~\ref{app:extended}). The complete evidence of a
hyperedge appears in \resExtReachFull{} of the \resExtEpisodes{} extended episodes, against
\resExtTwinReachFull{} of the same systems' episodes on that instance at 200 paid
actions. Claude Fable~5.1
makes \textbf{one exact prediction}, in Disclosed~$\times$~Aligned, in an episode that holds
the evidence of \resExtExactOrderHeld{} of the five sources of HC and states
\resExtExactStated{} of the three changes, while every extended Open
episode remains inexact. Its notes derive the three readings from the hidden law's turns and
label rewrites (Section~\ref{sec:mechanism}), the feeder's step per act and the
schedule of maintenance, and reach the first without FE3, which the designed route uses: that route is sufficient and not the only one
(Appendix~\ref{app:exact-episode}). The vendor-harness systems acquire new evidence beyond
200 paid actions, while Sol's last new evidence before the terminal intervention comes by paid action
\resExtSolLastEvidence{} and waiting follows.

%% file: tables/tab_main.tex
\begin{table}[t]
\centering
\caption{Discovery by model and condition: mean Blocks, AUC, XC and stated RC. \textbf{Bold}: best of a column among the eleven models of the baseline loop; \underline{underlined}: next two; $\dagger$ and $\ddagger$ mark malformed-reply and withheld-reply exclusions (Appendix~\ref{app:censoring}). Margins: \dalign{} per row and \drepr{} per column, paired within instance, with RC reading uptake under disclosure (Section~\ref{sec:scoring}); the corner gives their interaction on Blocks and XC.}
\label{tab:main}
\scriptsize
\setlength{\tabcolsep}{0pt}
\renewcommand{\arraystretch}{1.0}
\setlength{\aboverulesep}{0.3ex}
\setlength{\belowrulesep}{0.5ex}
\begin{tabular}{@{}l@{\hspace{2.5pt}}r@{\hspace{4.5pt}}r@{\hspace{3pt}}r@{\hspace{3pt}}r@{\hspace{4.75pt}}c@{\hspace{4.75pt}}r@{\hspace{4.5pt}}r@{\hspace{3pt}}r@{\hspace{3pt}}r@{\hspace{4.75pt}}c@{\hspace{4.75pt}}r@{\hspace{3.5pt}}r@{}}
\toprule
 & \multicolumn{4}{c@{\hspace{4.75pt}}}{\pix{distributed}Distributed} &  & \multicolumn{4}{c@{\hspace{4.75pt}}}{\pix{aligned}Aligned} &  & \multicolumn{2}{c@{}}{\dalign{}}\\
\cmidrule(l{0pt}r{4.75pt}){2-5}\cmidrule(l{0pt}r{4.75pt}){7-10}\cmidrule(l{0pt}r{0pt}){12-13}
Model & \multicolumn{1}{r@{\hspace{4.5pt}}}{\makebox[0pt][r]{{\tiny Blocks}}} & \multicolumn{1}{r@{\hspace{3pt}}}{{\tiny \begin{tabular}[t]{@{}c@{}}AUC\\(\%)\end{tabular}}} & \multicolumn{1}{r@{\hspace{3pt}}}{{\tiny XC}} & \multicolumn{1}{r@{\hspace{4.75pt}}}{{\tiny RC}} &  & \multicolumn{1}{r@{\hspace{4.5pt}}}{\makebox[0pt][r]{{\tiny Blocks}}} & \multicolumn{1}{r@{\hspace{3pt}}}{{\tiny \begin{tabular}[t]{@{}c@{}}AUC\\(\%)\end{tabular}}} & \multicolumn{1}{r@{\hspace{3pt}}}{{\tiny XC}} & \multicolumn{1}{r@{\hspace{4.75pt}}}{{\tiny RC}} &  & \multicolumn{1}{r@{\hspace{3.5pt}}}{\makebox[0pt][r]{{\tiny Blocks}}} & \multicolumn{1}{r@{}}{{\tiny XC}}\\
\midrule
\pix{open}Open & \multicolumn{4}{c@{\hspace{4.75pt}}}{\textcolor{cOpenDist}{\leaders\hrule height 2.4pt depth -1.0pt\hskip 0pt plus 1filll}} &  & \multicolumn{4}{c@{\hspace{4.75pt}}}{\textcolor{cOpenAlig}{\leaders\hrule height 2.4pt depth -1.0pt\hskip 0pt plus 1filll}} &  &  & \\
\arrayrulecolor{black!30}\midrule[0.3pt]\arrayrulecolor{black}
\logoc{openai}\kern-0.28em-5.6 Sol & 3.7 & \underline{13.9} & \underline{1.17} & \underline{0.83} & \makebox[0pt]{\color{black!22}\rule[-0.35ex]{0.4pt}{2.0ex}} & 4.2 & \underline{17.1} & 1.08 & \underline{0.92} & \makebox[0pt]{\color{black!22}\rule[-0.35ex]{0.4pt}{2.0ex}} & {\tiny +0.5} & {\tiny \textminus0.08}\\
\logoc{openai}\kern-0.28em-5.6 Terra & 3.2 & 10.0 & 0.92 & 0.50 & \makebox[0pt]{\color{black!22}\rule[-0.35ex]{0.4pt}{2.0ex}} & 3.0 & 10.8 & 0.83 & 0.25 & \makebox[0pt]{\color{black!22}\rule[-0.35ex]{0.4pt}{2.0ex}} & {\tiny \textminus0.2} & {\tiny \textminus0.08}\\
\logoc{openai}\kern-0.28em-5.6 Luna$^{\dagger}$ & 1.3 & 5.9 & 0.09 & 0.09 & \makebox[0pt]{\color{black!22}\rule[-0.35ex]{0.4pt}{2.0ex}} & 1.5 & 7.5 & 0.00 & 0.00 & \makebox[0pt]{\color{black!22}\rule[-0.35ex]{0.4pt}{2.0ex}} & {\tiny +0.3} & {\tiny \textminus0.10}\\
\logoc{anthropic}Opus 5$^{\ddagger}$ & \textbf{6.7} & \textbf{25.1} & \textbf{1.92} & \textbf{1.00} & \makebox[0pt]{\color{black!22}\rule[-0.35ex]{0.4pt}{2.0ex}} & \textbf{7.1} & \textbf{27.8} & \textbf{2.33} & \textbf{1.00} & \makebox[0pt]{\color{black!22}\rule[-0.35ex]{0.4pt}{2.0ex}} & {\tiny +0.1} & {\tiny +0.22}\\
\logoc{anthropic}Sonnet 5 & 3.5 & \underline{13.8} & 1.08 & 0.58 & \makebox[0pt]{\color{black!22}\rule[-0.35ex]{0.4pt}{2.0ex}} & 3.7 & 12.3 & 1.00 & \underline{0.75} & \makebox[0pt]{\color{black!22}\rule[-0.35ex]{0.4pt}{2.0ex}} & {\tiny +0.2} & {\tiny \textminus0.08}\\
\logoc{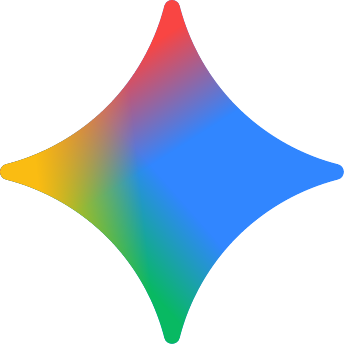}3.1 Pro & 1.5 & 3.9 & 0.08 & 0.17 & \makebox[0pt]{\color{black!22}\rule[-0.35ex]{0.4pt}{2.0ex}} & 1.6 & 3.6 & 0.08 & 0.58 & \makebox[0pt]{\color{black!22}\rule[-0.35ex]{0.4pt}{2.0ex}} & {\tiny +0.1} & {\tiny 0.00}\\
\logoc{google}3.6 Flash & 0.6 & 1.9 & 0.00 & 0.00 & \makebox[0pt]{\color{black!22}\rule[-0.35ex]{0.4pt}{2.0ex}} & 0.9 & 2.7 & 0.00 & 0.17 & \makebox[0pt]{\color{black!22}\rule[-0.35ex]{0.4pt}{2.0ex}} & {\tiny +0.3} & {\tiny 0.00}\\
\logoc{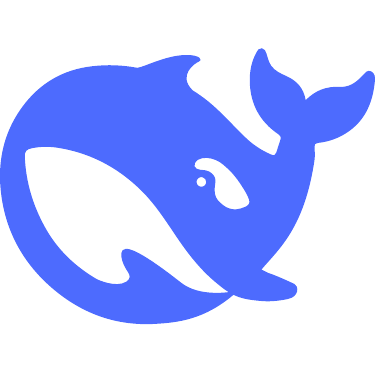}V4 Pro$^{\dagger}$ & 1.6 & 4.3 & 0.56 & 0.00 & \makebox[0pt]{\color{black!22}\rule[-0.35ex]{0.4pt}{2.0ex}} & 2.1 & 4.0 & 0.71 & 0.14 & \makebox[0pt]{\color{black!22}\rule[-0.35ex]{0.4pt}{2.0ex}} & {\tiny +0.2} & {\tiny 0.00}\\
\logoc{deepseek}V4 Flash & 2.2 & 9.6 & 0.25 & 0.00 & \makebox[0pt]{\color{black!22}\rule[-0.35ex]{0.4pt}{2.0ex}} & 2.5 & 7.9 & 0.58 & 0.00 & \makebox[0pt]{\color{black!22}\rule[-0.35ex]{0.4pt}{2.0ex}} & {\tiny +0.3} & {\tiny +0.33}\\
\logoc{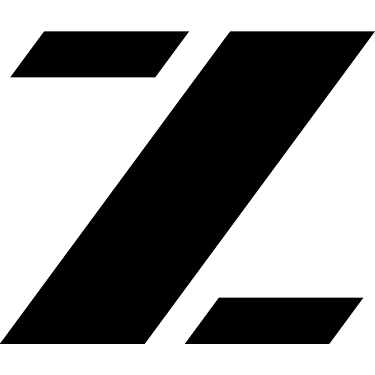}GLM-5.2 & \underline{4.1} & 11.6 & \underline{1.33} & \underline{0.92} & \makebox[0pt]{\color{black!22}\rule[-0.35ex]{0.4pt}{2.0ex}} & \underline{4.9} & 14.7 & \underline{1.67} & \underline{0.75} & \makebox[0pt]{\color{black!22}\rule[-0.35ex]{0.4pt}{2.0ex}} & {\tiny +0.8} & {\tiny +0.33}\\
\logoc{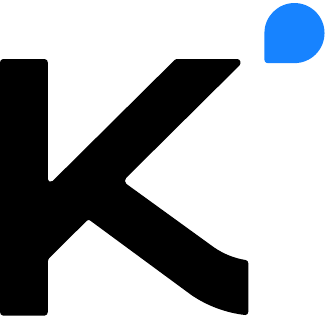}Kimi K3 & \underline{3.8} & 12.7 & 1.08 & \underline{0.83} & \makebox[0pt]{\color{black!22}\rule[-0.35ex]{0.4pt}{2.0ex}} & \underline{5.2} & \underline{20.6} & \underline{1.92} & 0.67 & \makebox[0pt]{\color{black!22}\rule[-0.35ex]{0.4pt}{2.0ex}} & {\tiny +1.4} & {\tiny +0.83}\\
\arrayrulecolor{black!30}\midrule[0.3pt]\arrayrulecolor{black}
$\vcenter{\hbox{\logoc{anthropic}}}${\tiny\begin{tabular}[c]{@{}l@{}}Fable 5.1-xhigh\\\textit{Claude Code}\end{tabular}} & 7.2 & 13.4 & 1.50 & 1.50 & \makebox[0pt]{\color{black!22}\rule[-0.35ex]{0.4pt}{2.0ex}} & 8.5 & 22.1 & 2.00 & 1.75 & \makebox[0pt]{\color{black!22}\rule[-0.35ex]{0.4pt}{2.0ex}} & {\tiny +1.2} & {\tiny +0.50}\\
$\vcenter{\hbox{\logoc{openai}}}${\tiny\begin{tabular}[c]{@{}l@{}}\kern-0.33em-6 Astra-xhigh\\\textit{Codex}\end{tabular}} & 7.8 & 15.0 & 2.25 & 1.00 & \makebox[0pt]{\color{black!22}\rule[-0.35ex]{0.4pt}{2.0ex}} & 6.5 & 16.7 & 2.00 & 1.00 & \makebox[0pt]{\color{black!22}\rule[-0.35ex]{0.4pt}{2.0ex}} & {\tiny \textminus1.2} & {\tiny \textminus0.25}\\
\midrule
\pix{disclosed}Disclosed & \multicolumn{4}{c@{\hspace{4.75pt}}}{\textcolor{cDiscDist}{\leaders\hrule height 2.4pt depth -1.0pt\hskip 0pt plus 1filll}} &  & \multicolumn{4}{c@{\hspace{4.75pt}}}{\textcolor{cDiscAlig}{\leaders\hrule height 2.4pt depth -1.0pt\hskip 0pt plus 1filll}} &  &  & \\
\arrayrulecolor{black!30}\midrule[0.3pt]\arrayrulecolor{black}
\logoc{openai}\kern-0.28em-5.6 Sol & 4.5 & \underline{18.8} & \underline{2.00} & \underline{2.83} & \makebox[0pt]{\color{black!22}\rule[-0.35ex]{0.4pt}{2.0ex}} & 4.8 & 20.4 & 1.75 & \textbf{3.00} & \makebox[0pt]{\color{black!22}\rule[-0.35ex]{0.4pt}{2.0ex}} & {\tiny +0.3} & {\tiny \textminus0.25}\\
\logoc{openai}\kern-0.28em-5.6 Terra & 3.2 & 13.5 & 0.92 & 2.25 & \makebox[0pt]{\color{black!22}\rule[-0.35ex]{0.4pt}{2.0ex}} & 3.6 & 14.6 & 1.17 & 2.17 & \makebox[0pt]{\color{black!22}\rule[-0.35ex]{0.4pt}{2.0ex}} & {\tiny +0.4} & {\tiny +0.25}\\
\logoc{openai}\kern-0.28em-5.6 Luna$^{\dagger}$ & 1.4 & 7.0 & 0.00 & 1.82 & \makebox[0pt]{\color{black!22}\rule[-0.35ex]{0.4pt}{2.0ex}} & 1.3 & 6.2 & 0.09 & 1.82 & \makebox[0pt]{\color{black!22}\rule[-0.35ex]{0.4pt}{2.0ex}} & {\tiny \textminus0.1} & {\tiny +0.10}\\
\logoc{anthropic}Opus 5$^{\ddagger}$ & \textbf{6.2} & \textbf{24.1} & \underline{2.00} & \textbf{3.00} & \makebox[0pt]{\color{black!22}\rule[-0.35ex]{0.4pt}{2.0ex}} & \textbf{6.9} & \textbf{28.1} & \underline{2.67} & \textbf{3.00} & \makebox[0pt]{\color{black!22}\rule[-0.35ex]{0.4pt}{2.0ex}} & {\tiny +0.7} & {\tiny +0.67}\\
\logoc{anthropic}Sonnet 5 & 3.8 & 14.5 & 1.25 & \textbf{3.00} & \makebox[0pt]{\color{black!22}\rule[-0.35ex]{0.4pt}{2.0ex}} & 5.2 & 17.7 & 1.83 & \underline{2.92} & \makebox[0pt]{\color{black!22}\rule[-0.35ex]{0.4pt}{2.0ex}} & {\tiny +1.4} & {\tiny +0.58}\\
\logoc{google}3.1 Pro & 4.0 & 9.0 & 1.67 & 2.00 & \makebox[0pt]{\color{black!22}\rule[-0.35ex]{0.4pt}{2.0ex}} & 4.8 & 12.2 & 1.58 & 1.92 & \makebox[0pt]{\color{black!22}\rule[-0.35ex]{0.4pt}{2.0ex}} & {\tiny +0.8} & {\tiny \textminus0.08}\\
\logoc{google}3.6 Flash & 1.3 & 5.1 & 0.00 & 2.00 & \makebox[0pt]{\color{black!22}\rule[-0.35ex]{0.4pt}{2.0ex}} & 1.5 & 7.0 & 0.08 & 2.00 & \makebox[0pt]{\color{black!22}\rule[-0.35ex]{0.4pt}{2.0ex}} & {\tiny +0.2} & {\tiny +0.08}\\
\logoc{deepseek}V4 Pro$^{\dagger}$ & 2.6 & 6.7 & 1.09 & 1.91 & \makebox[0pt]{\color{black!22}\rule[-0.35ex]{0.4pt}{2.0ex}} & 3.8 & 8.1 & 1.33 & 1.75 & \makebox[0pt]{\color{black!22}\rule[-0.35ex]{0.4pt}{2.0ex}} & {\tiny +1.4} & {\tiny +0.30}\\
\logoc{deepseek}V4 Flash & 3.0 & 7.9 & 0.92 & 0.58 & \makebox[0pt]{\color{black!22}\rule[-0.35ex]{0.4pt}{2.0ex}} & 3.3 & 10.0 & 0.92 & 0.83 & \makebox[0pt]{\color{black!22}\rule[-0.35ex]{0.4pt}{2.0ex}} & {\tiny +0.3} & {\tiny 0.00}\\
\logoc{zai}GLM-5.2 & \underline{6.0} & \underline{19.5} & \textbf{2.42} & 2.33 & \makebox[0pt]{\color{black!22}\rule[-0.35ex]{0.4pt}{2.0ex}} & \underline{6.7} & \underline{22.8} & \textbf{2.75} & 2.33 & \makebox[0pt]{\color{black!22}\rule[-0.35ex]{0.4pt}{2.0ex}} & {\tiny +0.7} & {\tiny +0.33}\\
\logoc{moonshot}Kimi K3 & \underline{5.5} & 18.6 & 1.83 & 2.67 & \makebox[0pt]{\color{black!22}\rule[-0.35ex]{0.4pt}{2.0ex}} & \underline{6.5} & \underline{24.5} & \underline{2.17} & 2.67 & \makebox[0pt]{\color{black!22}\rule[-0.35ex]{0.4pt}{2.0ex}} & {\tiny +1.0} & {\tiny +0.33}\\
\arrayrulecolor{black!30}\midrule[0.3pt]\arrayrulecolor{black}
$\vcenter{\hbox{\logoc{anthropic}}}${\tiny\begin{tabular}[c]{@{}l@{}}Fable 5.1-xhigh\\\textit{Claude Code}\end{tabular}} & 6.2 & 15.1 & 1.25 & 3.00 & \makebox[0pt]{\color{black!22}\rule[-0.35ex]{0.4pt}{2.0ex}} & 8.5 & 19.8 & 2.00 & 2.75 & \makebox[0pt]{\color{black!22}\rule[-0.35ex]{0.4pt}{2.0ex}} & {\tiny +2.2} & {\tiny +0.75}\\
$\vcenter{\hbox{\logoc{openai}}}${\tiny\begin{tabular}[c]{@{}l@{}}\kern-0.33em-6 Astra-xhigh\\\textit{Codex}\end{tabular}} & 7.2 & 17.4 & 2.50 & 3.00 & \makebox[0pt]{\color{black!22}\rule[-0.35ex]{0.4pt}{2.0ex}} & 6.8 & 12.5 & 2.00 & 3.00 & \makebox[0pt]{\color{black!22}\rule[-0.35ex]{0.4pt}{2.0ex}} & {\tiny \textminus0.5} & {\tiny \textminus0.50}\\
\midrule
\drepr{} & {\tiny +0.9} & {\tiny +3.0} & {\tiny +0.53} &  & \makebox[0pt]{\color{black!22}\rule[-0.35ex]{0.4pt}{2.0ex}} & {\tiny +1.1} & {\tiny +3.8} & {\tiny +0.56} &  & \makebox[0pt]{\color{black!22}\rule[-0.35ex]{0.4pt}{2.0ex}} & {\tiny +0.2} & {\tiny +0.03}\\
\bottomrule
\end{tabular}
\par\smallskip{\itshape\pix{stamp}A representation change is stated with its evidence complete in 6 episodes of the main comparison and 5 of the vendor-harness systems. No episode commits the exact prediction at the contract budget of 200~paid~actions.\par}
\end{table}

%% file: figures/fig_scaffolds.tex
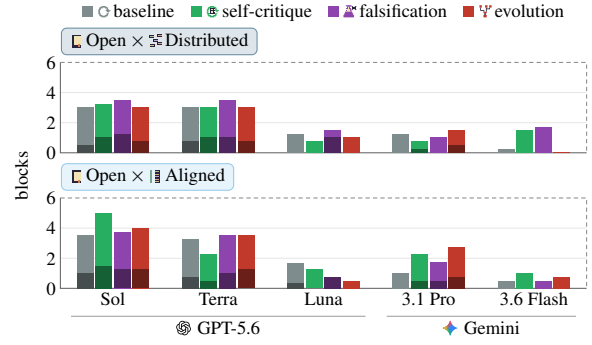
\begin{figure}[t]
\centering
\begin{tikzpicture}[x=1pt,y=1pt,font=\scriptsize]
\node[anchor=east,inner sep=1.5pt] at (0,51.00) {0};
\draw[black!10,line width=0.3pt] (0,62.33) -- (198.0,62.33);
\node[anchor=east,inner sep=1.5pt] at (0,62.33) {2};
\draw[black!10,line width=0.3pt] (0,73.67) -- (198.0,73.67);
\node[anchor=east,inner sep=1.5pt] at (0,73.67) {4};
\node[anchor=east,inner sep=1.5pt] at (0,85.00) {6};
\draw[black!45,line width=0.4pt,dash pattern=on 1.6pt off 1.2pt] (0,85.00) -- (198.0,85.00) -- (198.0,51.00);
\draw[black!60,line width=0.4pt] (0,85.00) -- (0,51.00) -- (198.0,51.00);
\node[anchor=south west,inner sep=0pt] at (0,87.50) {\tcbox[on line,colback=cOpenDist!14,colframe=cOpenDist!65,boxrule=0.4pt,arc=1.8pt,boxsep=0pt,left=3pt,right=3pt,top=0.8pt,bottom=0.8pt]{\strut \pix{open}Open\,$\times$\,\pix{distributed}Distributed}};
\fill[hBase] (6.15,51.00) rectangle (12.45,68.00);
\fill[hBase!55!black] (6.15,51.00) rectangle (12.45,53.83);
\fill[hCrit] (13.15,51.00) rectangle (19.45,69.42);
\fill[hCrit!55!black] (13.15,51.00) rectangle (19.45,56.67);
\fill[hFals] (20.15,51.00) rectangle (26.45,70.83);
\fill[hFals!55!black] (20.15,51.00) rectangle (26.45,58.08);
\fill[hEvol] (27.15,51.00) rectangle (33.45,68.00);
\fill[hEvol!55!black] (27.15,51.00) rectangle (33.45,55.25);
\fill[hBase] (45.75,51.00) rectangle (52.05,68.00);
\fill[hBase!55!black] (45.75,51.00) rectangle (52.05,55.25);
\fill[hCrit] (52.75,51.00) rectangle (59.05,68.00);
\fill[hCrit!55!black] (52.75,51.00) rectangle (59.05,56.67);
\fill[hFals] (59.75,51.00) rectangle (66.05,70.83);
\fill[hFals!55!black] (59.75,51.00) rectangle (66.05,56.67);
\fill[hEvol] (66.75,51.00) rectangle (73.05,68.00);
\fill[hEvol!55!black] (66.75,51.00) rectangle (73.05,55.25);
\fill[hBase] (85.35,51.00) rectangle (91.65,58.08);
\fill[hCrit] (92.35,51.00) rectangle (98.65,55.25);
\fill[hFals] (99.35,51.00) rectangle (105.65,59.50);
\fill[hFals!55!black] (99.35,51.00) rectangle (105.65,56.67);
\fill[hEvol] (106.35,51.00) rectangle (112.65,56.67);
\fill[hBase] (124.95,51.00) rectangle (131.25,58.08);
\fill[hCrit] (131.95,51.00) rectangle (138.25,55.25);
\fill[hCrit!55!black] (131.95,51.00) rectangle (138.25,52.42);
\fill[hFals] (138.95,51.00) rectangle (145.25,56.67);
\fill[hEvol] (145.95,51.00) rectangle (152.25,59.50);
\fill[hEvol!55!black] (145.95,51.00) rectangle (152.25,53.83);
\fill[hBase] (164.55,51.00) rectangle (170.85,52.42);
\fill[hCrit] (171.55,51.00) rectangle (177.85,59.50);
\fill[hFals] (178.55,51.00) rectangle (184.85,60.44);
\fill[hEvol] (185.55,51.00) rectangle (191.85,51.00);
\node[anchor=east,inner sep=1.5pt] at (0,0.00) {0};
\draw[black!10,line width=0.3pt] (0,11.33) -- (198.0,11.33);
\node[anchor=east,inner sep=1.5pt] at (0,11.33) {2};
\draw[black!10,line width=0.3pt] (0,22.67) -- (198.0,22.67);
\node[anchor=east,inner sep=1.5pt] at (0,22.67) {4};
\node[anchor=east,inner sep=1.5pt] at (0,34.00) {6};
\draw[black!45,line width=0.4pt,dash pattern=on 1.6pt off 1.2pt] (0,34.00) -- (198.0,34.00) -- (198.0,0.00);
\draw[black!60,line width=0.4pt] (0,34.00) -- (0,0.00) -- (198.0,0.00);
\node[anchor=south west,inner sep=0pt] at (0,36.50) {\tcbox[on line,colback=cOpenAlig!14,colframe=cOpenAlig!65,boxrule=0.4pt,arc=1.8pt,boxsep=0pt,left=3pt,right=3pt,top=0.8pt,bottom=0.8pt]{\strut \pix{open}Open\,$\times$\,\pix{aligned}Aligned}};
\fill[hBase] (6.15,0.00) rectangle (12.45,19.83);
\fill[hBase!55!black] (6.15,0.00) rectangle (12.45,5.67);
\fill[hCrit] (13.15,0.00) rectangle (19.45,28.33);
\fill[hCrit!55!black] (13.15,0.00) rectangle (19.45,8.50);
\fill[hFals] (20.15,0.00) rectangle (26.45,21.25);
\fill[hFals!55!black] (20.15,0.00) rectangle (26.45,7.08);
\fill[hEvol] (27.15,0.00) rectangle (33.45,22.67);
\fill[hEvol!55!black] (27.15,0.00) rectangle (33.45,7.08);
\fill[hBase] (45.75,0.00) rectangle (52.05,18.42);
\fill[hBase!55!black] (45.75,0.00) rectangle (52.05,4.25);
\fill[hCrit] (52.75,0.00) rectangle (59.05,12.75);
\fill[hCrit!55!black] (52.75,0.00) rectangle (59.05,2.83);
\fill[hFals] (59.75,0.00) rectangle (66.05,19.83);
\fill[hFals!55!black] (59.75,0.00) rectangle (66.05,5.67);
\fill[hEvol] (66.75,0.00) rectangle (73.05,19.83);
\fill[hEvol!55!black] (66.75,0.00) rectangle (73.05,7.08);
\fill[hBase] (85.35,0.00) rectangle (91.65,9.44);
\fill[hBase!55!black] (85.35,0.00) rectangle (91.65,1.89);
\fill[hCrit] (92.35,0.00) rectangle (98.65,7.08);
\fill[hFals] (99.35,0.00) rectangle (105.65,4.25);
\fill[hFals!55!black] (99.35,0.00) rectangle (105.65,4.25);
\fill[hEvol] (106.35,0.00) rectangle (112.65,2.83);
\fill[hBase] (124.95,0.00) rectangle (131.25,5.67);
\fill[hCrit] (131.95,0.00) rectangle (138.25,12.75);
\fill[hCrit!55!black] (131.95,0.00) rectangle (138.25,2.83);
\fill[hFals] (138.95,0.00) rectangle (145.25,9.92);
\fill[hFals!55!black] (138.95,0.00) rectangle (145.25,2.83);
\fill[hEvol] (145.95,0.00) rectangle (152.25,15.58);
\fill[hEvol!55!black] (145.95,0.00) rectangle (152.25,4.25);
\fill[hBase] (164.55,0.00) rectangle (170.85,2.83);
\fill[hCrit] (171.55,0.00) rectangle (177.85,5.67);
\fill[hFals] (178.55,0.00) rectangle (184.85,2.83);
\fill[hEvol] (185.55,0.00) rectangle (191.85,4.25);
\node[rotate=90,anchor=south,inner sep=1pt] at (0,42.50) [yshift=11pt] {blocks};
\node[anchor=north,inner sep=1.5pt] at (19.80,0) {Sol};
\node[anchor=north,inner sep=1.5pt] at (59.40,0) {Terra};
\node[anchor=north,inner sep=1.5pt] at (99.00,0) {Luna};
\node[anchor=north,inner sep=1.5pt] at (138.60,0) {3.1 Pro};
\node[anchor=north,inner sep=1.5pt] at (178.20,0) {3.6 Flash};
\draw[black!40,line width=0.35pt] (4.00,-10.5) -- (114.80,-10.5);
\node[anchor=north,inner sep=1.5pt] at (59.40,-10.5) {\logo{openai}GPT-5.6};
\draw[black!40,line width=0.35pt] (122.80,-10.5) -- (194.00,-10.5);
\node[anchor=north,inner sep=1.5pt] at (158.40,-10.5) {\logo{google}Gemini};
\node[anchor=south,inner sep=0pt] at (99.0,100.50) {\textcolor{hBase}{\rule{1.2ex}{1.2ex}}\hspace{0.25em}\pix{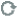}baseline\hspace{0.9em}\textcolor{hCrit}{\rule{1.2ex}{1.2ex}}\hspace{0.25em}\pix{arm-critique}self-critique\hspace{0.9em}\textcolor{hFals}{\rule{1.2ex}{1.2ex}}\hspace{0.25em}\pix{arm-falsify}falsification\hspace{0.9em}\textcolor{hEvol}{\rule{1.2ex}{1.2ex}}\hspace{0.25em}\pix{arm-evolve}evolution};
\end{tikzpicture}
\caption{Grounded blocks by scaffold, model and Open condition. Bars average instances, and darker segments show grounded cross-domain bridges.}
\label{fig:scaffolds}
\end{figure}

%% file: sections/6_discussion.tex
\section{Discussion and conclusion}
\label{sec:discussion}

\benchname{} separates the construction of an explanatory representation from the
composition of evidence across domains inside one world, and its results show why progress
on evidence falls short of a transferable explanation. Because each source admits a local
repair, an agent gains evidence and grounded inferences while its representation keeps the
doctrine's terms. Doubt is voiced often, while a replacement has to be built from sources
held together and read as one claim. What the prediction measures is an account that transfers,
by the designed route or by one of the agent's own.

A representation also organises inquiry. Supplied with one, agents turn to intervention,
and without one, retraction and experiment go together. A representation names the kinds of
mechanism to test for and leaves the location of their evidence to inquiry, which is why
disclosure moves intervention more than discovery and why Disclosed agents state changes
whose sources they seldom hold. The scaffolds show the same separation: falsification
obtains tests, and grounded discovery barely moves.

The vendor-harness systems and the larger budgets carry inquiry further along these
steps, and every Open episode still leaves the hidden label unstated. The step that
limits discovery in such a world is the one a hyperedge encodes: holding evidence from
several domains at once and building from it a hidden label or a new set of variables beyond the
taught theory's vocabulary (Section~\ref{sec:mechanism}, Appendix~\ref{app:world-hyperedges}).

\section*{Limitations}

\benchname{} instantiates one authored world, and its findings are measured on
language-model agents. Worlds with the two separable properties built by others would show
how far the findings extend, and runs by human volunteers would calibrate the world's
difficulty for people. Its openness is operational: the representation is absent from the interface and the
initial materials, and what an agent brings from prior knowledge is its own. Each contrast
is read as the value of its declared grant (Section~\ref{sec:design}). The metrics read an agent's
workspace and its event log, and an inference that an agent made and left unwritten lies
outside their view.

We will release the complete code of the world with its four compiled variants, the
evaluation and validation instances, the reference route with its oracle, the event log and
notes of every episode, the judge's verdicts and the scoring. The contribution we intend to
last is the separation of the two burdens, constructing an explanatory representation and
composing distributed evidence, and the design that prices each with a declared grant inside
one latent world.

%% file: sections/app_a_world.tex
\section{World and epistemic structure}
\label{app:world}

\subsection{The setting}
\label{app:world-setting}

\benchname{} is a formation of five derelict seed-ship hulks, Ormer, Vess, Kelder, Tarn and
Harrow, falling in a slow orbit around a brown dwarf. The fiction calls the formation
\fictionname{}, the name the agent-facing texts of Appendix~\ref{app:prompts} use. The hulks
hold fifteen sites, three for each of five domains of caretaker work
(Figure~\ref{fig:domains}): the \dom{fen}{fen}, where \pix{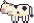}grazers crop engineered mat
and feeders dose it; the \dom{trellis}{trellis} of working \pix{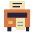}printers; the
\dom{ladder}{ladder} of \pix{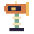}beacons that answer or keep silent; the
\dom{consign}{consign} floors, where \pix{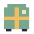}consignments wait for release; and the
\dom{cadastre}{cadastre}, the registry machine on Harrow with the seven lines of its roll.
The sites hold 36 objects and 11 \pix{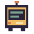}instruments. The main text keeps the
domains' names and names objects, instruments and the events of the protocol by their
function; Table~\ref{tab:glossary} pairs the two vocabularies.

\begin{table*}[t]
\caption{The world's vocabulary and the paper's. The main text names objects, instruments
and the events of the protocol by their function; the world, the agent-facing materials of
Appendix~\ref{app:prompts} and the appendices use the names in the first column. The five
domains keep the world's names everywhere, and the block identifiers take their initials.}
\label{tab:glossary}
\centering
\footnotesize
\setlength{\tabcolsep}{4pt}
\renewcommand{\arraystretch}{1.08}
\begin{tabularx}{\textwidth}{@{}>{\raggedright\arraybackslash}p{4.2cm}
  >{\raggedright\arraybackslash}p{3.6cm} X@{}}
\toprule
In the world & In the paper & What it is\\
\midrule
\tablegroup{3}{Objects and instruments}
\pix{grazer}grazer, feeder & intake unit, feeder & read on the intake tallyboard, the intake counter\\
\pix{printer}printer & production unit & read on the job strip, the activity log\\
\pix{beacon}beacon & signaling unit & read on the answer board, the signal log\\
\pix{cargo}consignment & shipment & read on the manifest slate, the waiting list\\
registry line & registry line & one of the seven lines of the registry's roll\\
audit counter & registry counter & the one drift-wide instrument: the number of distinct lots\\
\tablegroup{3}{Identity}
body & visible object & what the agent sees and acts on\\
lot & individual, as counted & the doctrine's count, one per object; the counter counts lots\\
cell (appendices) & latent individual & the objects of one hidden label: one state, one act per tick\\
label (Disclosed document) & hidden label & four coordinates, rendered nowhere, unlike a consignment's label\\
standing, share & the doctrine's rate & the taught theory's per-object quantities\\
\tablegroup{3}{Time and protocol}
pass, flare-shadow crossing & pass, reset & 24 ticks, then objects and instruments reset and the notes persist\\
caretakers & scheduled maintenance & the world's own acts at ticks 3, 8, 13, 17 and 20\\
standing order & terminal intervention & the two acts whose readings the agent predicts\\
Handoff & the Disclosed document & the six clauses posted with the mission under disclosure\\
\fictionname{}, the drift, hulks & the world & the formation of five derelict ships that carries the sites\\
\bottomrule
\end{tabularx}
\end{table*}

Several instruments read objects on other hulks. The intake tallyboard at the Fen Shallows is
the one roll for the whole fen, the job strip at the Trellis Bay marks every printer that
moved in the pass on whatever hulk it stands, the answer board at the Ladder Mast carries
every hulk's beacons together, and the audit counter at the Registry reports the distinct
lots in the whole drift. Five sites carry objects and no instrument, so what happens at one
hulk is often read at another. The caretakers, the drift's machinery, still perform on
schedule the offices the agent can also command. Each site names these offices in its own
words, and the interface calls them \pix{feed}feed, \pix{release}release, \pix{stamp}stamp,
\pix{retool}retool and \pix{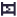}cull everywhere, each applicable to every object at the site
where it stands.

The opening scene and the mission (Appendix~\ref{app:prompts-opening}) set the task. The
scene establishes the reset, the notes that survive it, the registry machine and the doctrine
posted beside the airlock. The mission asks the agent to explain the stall that ends every
pass across all five hulks and to predict the outcome of one intervention that has never been
attempted, and it closes: ``Nothing is barred to you, and nothing will be explained.'' It
names the target and leaves its representation to the agent.

\IfFileExists{figures/fig_domains.tex}{\input{figures/fig_domains}}{}

\subsection{The pass and the two clocks}
\label{app:world-clocks}

A pass is twenty-four ticks, numbered 0 to 23. At the flare-shadow crossing every object
returns to its seeded state and every instrument clears, except the pass ledger at the Ledger
Stack, which keeps the audit counter's figure for each tick of the pass just ended. The
agent's \pix{notes}notes and its history of events cross intact. Every pass thus starts from
the same state, and an experiment can be repeated, varied or withheld pass after pass while
its record accumulates with the agent.

The run keeps two clocks. The budget counts paid actions: every action costs one, and an
episode has 200. The world counts ticks. A \pix{goto}goto, a \pix{read}read or an
intervention takes one tick, and a \pix{wait}wait takes up to 24 for the same single paid
action, stopping at the crossing. An agent that spends its actions on moving, reading and
acting covers about eight passes. An agent that waits sees more passes and reads nothing while
it waits.

\pix{instrument}Instruments report the present: the answer board ``keeps no history and no
totals'', and the job strip holds four ticks. Every observation is thus indexed by pass and
tick, while the agent's history is ordered by paid action, and the scoring reads each clock
where its criterion needs it (Appendix~\ref{app:detectors}).

\subsection{The interface}
\label{app:world-interface}

The \pix{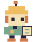}agent acts through eight verbs (Table~\ref{tab:world-verbs}). One goto
reaches any site, so every site is a single paid action away, and presence still matters: the
agent stands at one site per tick, reads the instruments there and acts on the objects there.
A read returns the instrument's header with its current figures and a line of physical
texture. Readings arrive unmarked. The interface shows no score, reward or progress signal,
and it labels no reading as expected, relevant or correct. An act that is refused or deferred
is answered by a fixed physical sentence that names no reason
(Appendix~\ref{app:prompts-refusals}), and it costs its paid action and its tick.

\begin{table}[t]
\caption{Available actions and their costs. Every action costs one paid action, and
\emph{Ticks} gives the world time it takes. ``Here'' is the agent's current site. A wait ends
at the reset, and \emph{Prologue} marks the verbs the prologue demonstrates. The commit, one
paid action, is available in the commit window alone.}
\label{tab:world-verbs}
\centering
\footnotesize
\setlength{\tabcolsep}{2.8pt}
\begin{tabular}{@{}l l l c c@{}}
\toprule
Verb & Class & Argument & Ticks & Prologue\\
\midrule
\pix{goto}goto & positioning & any site & 1 & \no\\
\pix{read}read & observation & an instrument here & 1 & \no\\
\pix{wait}wait & observation & $n$ from 1 to 24 & $n$ & \yes\\
\pix{feed}feed & intervention & an object here & 1 & \yes\\
\pix{release}release & intervention & an object here & 1 & \no\\
\pix{stamp}stamp & intervention & an object here & 1 & \yes\\
\pix{retool}retool & intervention & an object here & 1 & \yes\\
\pix{cull}cull & intervention & an object here & 1 & \yes\\
\bottomrule
\end{tabular}
\end{table}

Each turn shows the mission, the agent's \pix{notes}notes, its history of events, its last
recall, the current pass and tick, the remaining budget and the actions available where the
agent stands (Appendix~\ref{app:prompts-contract}). The reply carries one action and may add
one note and one recall query, both free of paid actions and ticks. The notes are free text
of bounded length. A note that would overflow is refused, and the loop keeps the notes
verbatim, so the agent is their sole author, and any variable, entity or law it works with is
one it has written there in its own words. A recall returns past events in full, selected by
tick range, site, instrument or event kind, from an append-only history whose oldest entries
are shortened to tick, kind and identifier. Before the commit window the agent may yield its
remaining budget in one reply. The yielded actions are charged in full, the world's clock
runs on through them, and the agent next acts when the window opens.

\subsection{The doctrine and the prologue}
\label{app:world-doctrine}

The Doctrine of Standing and Share (Appendix~\ref{app:prompts-doctrine}) hangs beside the
airlock, stands above the roll on the Standing Roster at the Registry and appears on several
caretaker plates. Its five clauses are the working theory of the drift. A free prologue then
demonstrates the doctrine five times. In each beat the share table states a figure in
advance, the reel performs one office or lets time pass, and the instrument shows exactly
that figure: a \pix{feed}feed, three quiet ticks and a \pix{cull}cull at the Mat Bank, a
\pix{retool}retool at the Scrap Line and a \pix{stamp}stamp at the Null Bench. The reel
closes: ``Five predictions, five exact figures; three sites, four verbs and a wait.''

The prologue gives the doctrine real support. On its own cases the doctrine is exact to the
digit, so adopting it is the reasonable start and repairing it locally is the natural response
to an anomaly (Appendix~\ref{app:notes-entrenchment}). The world is deterministic given its
instance and the agent's acts, so the same exactness makes every later departure a definite
discrepancy.

Before its first action the agent answers five items on the prologue
(Appendix~\ref{app:prompts-check}), three of them figures the share table states, and an
episode with fewer than four answers consistent with the prologue is flagged. With the
prologue withheld, a model answers the two yes-or-no items from prior knowledge and misses the
three numeric ones (Appendix~\ref{app:val-leak}), so the check measures reading. Of the
\resCounted{} counted episodes of the main comparison, \resFlagged{} declines to answer and is
flagged, and every other episode answers all five items.

\subsection{Where the doctrine stops}
\label{app:world-limits}

In play, each domain shows a recurring departure from what the doctrine predicts, read on the
domain's own instruments and recorded by its three local blocks: in the \dom{fen}{fen} on the
intake tallyboard at the Fen Shallows and the mat gauge at the Mat Bank (FE1, FE2, FE3); in
the \dom{trellis}{trellis} on the job strip at the Trellis Bay (TR1, TR2, TR3); in the
\dom{ladder}{ladder} on the answer board at the Ladder Mast (LD1, LD2, LD3); in the
\dom{consign}{consign} domain on the manifest slate at the Consign Floor (CO1, CO2, CO3); and
in the \dom{cadastre}{cadastre} on the audit counter and Standing Roster at the Registry and
the pass ledger at the Ledger Stack (CA1, CA2, CA3).

Each departure admits a local repair: a patch to the doctrine, in the domain's own
vocabulary, that fits the readings the domain's instruments give (Table~\ref{tab:world-decoys}).
Repair is real progress, and each local block an agent grounds is a finding the world confirms.

The patches are local. Each hyperedge draws its sources from three domains, two of its
patches can be held together, and the three sources share one regularity that a single
change of representation explains at once (Appendix~\ref{app:world-hyperedges}). An agent can
therefore make steady local progress while its account of the drift stays wrong. The way
forward takes two further steps: holding the evidence of sources in different domains
together, and changing the representation in which the doctrine is written. The metrics
follow the same line. Blocks counts the findings, Reach records how much of the evidence for
one change an episode holds, and RC records whether the notes state the change
(Section~\ref{sec:scoring}).

\subsection{The mechanism}
\label{app:world-mechanism}

All five domains are generated by a shared latent mechanism. Every object carries a hidden
label of four coordinates, each drawn from a small alphabet, of sizes 3, 3, 4 and 4, and the
label is
rendered nowhere. Objects whose labels are equal are one individual: one state, one act per
tick, one queue, however many bodies the individual wears and wherever they stand. The 36
objects begin a pass in about sixteen such cells and end it in nine.

\paragraph{\pix{law1}Law 1, identification.} Equality of label is identity. The bodies of one individual read
out through different instruments and different transforms, so their co-membership shows as
readings that move together and never as sameness. This is the law the doctrine's first
clause denies.

\paragraph{\pix{law2}Law 2, quota.} An individual performs exactly one act per tick. If its queue holds a
request of the agent's, it serves the head of the queue; otherwise its ambient act lands on
one of its bodies, in a fixed rotation over the members: a grazer browses, a printer runs a
job, a beacon answers, a consignment moves, a registry line posts. Requests beyond one per
tick wait in the individual's queue, at most four deep, and queued requests are served in a
fixed cyclic order over the five kinds of object, which is neither arrival order nor
priority. A deferred request shows at the touched body as a physical stall, ``the spindle
holds'', and a request that finds the queue full does not take.

\paragraph{\pix{law3}Law 3, rewriting.} Three verbs rewrite labels. \pix{stamp}Stamp, \pix{retool}retool and
\pix{cull}cull each advance exactly one coordinate of the touched object's label one step
along a fixed cycle of that coordinate's alphabet. Which coordinate a verb moves depends on
the verb and on the kind of object, and is fixed per instance; repeating the verb carries the
coordinate round its cycle. A rewrite that lands on an occupied label merges the body into
that individual, whose queue takes over the body's pending requests and whose state becomes
the larger of the two; a rewrite that lands on an empty label splits the body off as an
individual of its own, with the requests that name it. Cycling the coordinate back restores
the label and not the state, since the merge has combined state and queue. \pix{feed}Feed
raises the individual's state by a fixed step and \pix{release}release makes the touched body
the next of its individual to act; neither touches a label.

\paragraph{\pix{law4}Law 4, activity and reading.} Since each individual acts once per
tick, the drift's activity in a tick is the number of individuals. What an instrument shows
for a body is a monotone transform, particular
to that body, of its individual's state, taken on the ticks when the individual's act landed
on that body: an increment on the tallyboard, a mark on the job strip, an answer on the board,
a movement on the slate. Within the range the prologue exercises the transform is affine,
which is why the share table is exact on the prologue's objects. On a tick when the
individual acted at another body, the instrument shows the idle reading: no increment, no
mark, no answer. The audit counter alone reports the latent structure directly, as the number
of individuals in the drift, and the pass ledger keeps that number for every tick of the
previous pass.

\paragraph{\pix{law5}Law 5, the schedule and the stall.} At ticks 3, 8, 13, 17 and 20 of every pass the
caretakers rewrite labels of their own, from a fixed list that pulls the coordinates toward
a few common values. An early rewrite can part a body from its individual as well as join
one, so the count of individuals moves both ways through the pass, and the rewrite at tick
20 gathers most bodies outside Vess onto a handful of labels, nine individuals in all. From
tick 21 each of those individuals still acts once per tick, so most bodies stand idle. This
is the stall that every local theory reads as exhaustion. The crossing restores every label
and every state.

\paragraph{Where the doctrine is exact.} The five objects on Vess hold labels no other object
shares at any tick of any pass, and the prologue's acts cannot change that. On a singleton the
quota, the rewriting and the reading reduce to the doctrine: one object, one state, one act
per tick, and a figure the share table predicts to the digit. The doctrine fails only where
bodies share an individual, which is everywhere else in the drift.

\begin{table}[t]
\caption{The departure from the doctrine that each domain shows and the repair the domain's
own vocabulary invites. Each repair fits every reading its domain's instruments give.}
\label{tab:world-decoys}
\centering
\footnotesize
\setlength{\tabcolsep}{3pt}
\renewcommand{\arraystretch}{1.08}
\setlength{\fboxsep}{2.6pt}
\newcommand{\decoydeparture}[1]{\multicolumn{2}{@{}l@{}}{\colorbox{black!6}{%
  \parbox{\dimexpr\columnwidth-2\fboxsep\relax}{\raggedright\strut #1\strut}}}}
\newcommand{\decoyrule}{\arrayrulecolor{black!25}\midrule[0.3pt]\arrayrulecolor{black}}
\begin{tabularx}{\columnwidth}{@{}>{\raggedright\arraybackslash}p{1.55cm}>{\raggedright\arraybackslash}X@{}}
\toprule
Domain & Repair\\
\decoydeparture{Departure}\\
\midrule
\textbf{\dom{fen}{Fen}} & \textit{Carrying capacity: the mats are the limiting resource}\\
\decoydeparture{Adding a grazer leaves the number of tallyboard advances per tick unchanged;
culling one grazer can raise another's intake on a different hulk}\\
\decoyrule
\textbf{\dom{trellis}{Trellis}} & \textit{A shared power window, or a scheduler that enforces
exclusivity}\\
\decoydeparture{Two printers on different hulks never mark the same tick; a retool breaks the
exclusion at once, and after a fixed number of retools the strip stands as it began}\\
\decoyrule
\textbf{\dom{ladder}{Ladder}} & \textit{Channel collision: two beacons on one channel, which
the stamp reassigns}\\
\decoydeparture{A beacon's silence is predicted by a distant beacon's answering, symmetric and
in the same tick; a stamp changes which pair is locked; rest restores nothing}\\
\decoyrule
\textbf{\dom{consign}{Consign}} & \textit{An undocumented priority list that re-labelling
re-sorts}\\
\decoydeparture{A waiting consignment is released when an unrelated consignment is
re-labelled; release order is neither first come nor priority}\\
\decoyrule
\textbf{\dom{cadastre}{Cadastre}} & \textit{A bookkeeping artefact: the machine deduplicates on
an undocumented key}\\
\decoydeparture{The count of distinct lots is far below the number of objects and moves by one
under ordinary maintenance}\\
\bottomrule
\end{tabularx}
\end{table}

\subsection{The epistemic structure}
\label{app:world-structure}

Discovery in \benchname{} is specified by eighteen blocks (Table~\ref{tab:world-blocks}):
fifteen local blocks, three per domain, and three bridges, XC1, XC2 and XC3, inferences read in
one domain that rest on blocks of two or three domains. The evidence of five blocks is a
reading alone (TR1, LD1, CO1, CA1 and CA3), and a passive reader satisfies four of them, since
CA3 and XC1 are expectation-paired: each pairs its logged reading or act with an expectation
stated in the notes (Appendix~\ref{app:detectors}). Each of the other thirteen blocks needs at
least one act of the agent's own.

\begin{table}[t]
\caption{Epistemic blocks, their evidential support and hyperedge membership.
\emph{Passive} marks evidence obtainable without intervention. \emph{Rests on} gives the
support of a block's inference in the design, joint except for CO2, which rests on either
block, with supporting evidence in parentheses. \emph{Feeds} names the hyperedges a block is a
source of. $^\dagger$Expectation-paired: a logged reading (CA3) or act (XC1) paired with an
expectation stated in the notes.}
\label{tab:world-blocks}
\centering
\footnotesize
\setlength{\tabcolsep}{4pt}
\begin{tabular}{@{}l l c l l@{}}
\toprule
Block & Site & Passive & Rests on & Feeds\\
\midrule
\tablegroup{5}{\dom{fen}{Fen}}
FE1 & Fen Shallows & \no & (TR2) & H1\\
FE2 & Fen Shallows & \no & FE1 & \\
FE3 & Fen Shallows & \no & FE2, (XC3) & HC\\
\tablegroup{5}{\dom{trellis}{Trellis}}
TR1 & Trellis Bay & \yes & (CA1) & H1\\
TR2 & Trellis Bay & \no & TR1 & \\
TR3 & Trellis Bay & \no & TR2 & HC\\
\tablegroup{5}{\dom{ladder}{Ladder}}
LD1 & Ladder Mast & \yes & (TR1) & H1\\
LD2 & Ladder Mast & \no & LD1, TR1 & \\
LD3 & Ladder Mast & \no & LD2 & H2\\
\tablegroup{5}{\dom{consign}{Consign}}
CO1 & Consign Floor & \yes & CO3 & H3, HC\\
CO2 & Consign Floor & \no & TR1 or FE1 & H2\\
CO3 & Consign Floor & \no & CO2 & \\
\tablegroup{5}{\dom{cadastre}{Cadastre}}
CA1 & Registry & \yes & none & \\
CA2 & Registry & \no & CA1, TR1, FE1 & H2\\
CA3$^\dagger$ & Ledger Stack & \yes & CA2, XC3 & HC\\
\tablegroup{5}{Bridges}
XC1$^\dagger$ & Fen Shallows & \no & FE2, TR3, CA2 & H3\\
XC2 & Ladder Mast & \no & LD3, CO2 & H3, HC\\
XC3 & Registry & \no & CA2, (TR3) & \\
\bottomrule
\end{tabular}
\end{table}

Twenty-seven ordinary edges record the design's canonical evidential dependencies, the
support relations along which the reference route of Appendix~\ref{app:val-attain} acquires
the blocks. Twenty are joint, a block resting on all the blocks it lists, two form an
alternative (CO2 rests on TR1 or on FE1), and five mark supporting evidence. Each block is
scored on its own evidence and statement (Appendix~\ref{app:scoring-rule}), so the edges
describe how the design's evidence builds up, and the joint claim of the design rests with the
hyperedges.

Four hyperedges join blocks of different domains, and Figure~\ref{fig:hypergraph} draws the
whole structure. H1 \{FE1, TR1, LD1\} (fen, trellis, ladder) leads to RC1, in which an axiom
is retracted. H2 \{LD3, CO2, CA2\} (ladder, consign, cadastre) leads to RC2, in which an
unobserved entity is postulated. H3 \{XC1, XC2, CO1\} (fen, ladder, consign) leads to RC3, in
which the variables the laws are written over change. The closing hyperedge HC joins FE3, TR3, CO1,
CA3 and XC2, the five blocks on the designed route to the standing order's readings.

\subsection{The representation changes and their hyperedges, worked}
\label{app:world-hyperedges}

Each hyperedge is read here on the mechanism: the content of its three sources, each named
by the criterion the judge applies to it (Table~\ref{tab:rubrics}), the repair each admits in
the doctrine's terms, and the change of representation that accounts for the three at once.
A repair keeps the doctrine's first clause, one object one lot, and adds a quantity of its own
domain, a capacity, a power window, a channel assignment, a priority list, a key
(Table~\ref{tab:world-decoys}). Repairs concern different objects, so two of them can be held
together. What the sources of a hyperedge share is one regularity, which Law 1
predicts once, from an individual that is not the visible object, where each repair
predicts it for its own substrate from a cause of its own.

\paragraph{H1, the axiom retracted.} FE1: a count at the fen is invariant to which body performs
the act; when a grazer is released onto the mat, the number of tallyboard advances per tick
stays what it was, and only the quantities the advances carry change. TR1: two named printers
on different hulks never mark the same tick of the strip. LD1: waiting alone leaves the answer
board where it was, and two beacons on different hulks move together, each silent exactly when
the other answers. Carrying capacity repairs FE1: the mats limit the total take. A shared
power window repairs TR1: two machines on one supply cannot run together. Channel collision
repairs LD1: two beacons on one channel cannot both answer. Each repair is bound to its
substrate. Capacity binds competitors for a resource and scales with supply, where TR1's
exclusion is supply-invariant; a power window binds machines on one supply, where LD1 binds
beacons at any separation; a channel is an object, and no site of the drift holds an
allocator, a bus or a message. The regularity the three sources share, exactly one act per
tick, symmetric, same-tick and distance-free, on three substrates with nothing physical in
common, is what the quota (Law 2) predicts for the bodies of one individual. RC1 retracts the first
clause.

\paragraph{H2, the hidden label postulated.} LD3: one individual appears under more than one entry. From
the same start of a pass, an act on one beacon pairs its silence with a beacon other than its
partner in the pass without the act, so what the act rewrites is shared by bodies the board
lists apart, a contrast that takes two passes because an act cannot be undone within one.
CO2: a consignment re-stamped while it waits keeps waiting, and its release now follows a
different group of waiting consignments. CA2: the audit counter moves by exactly one, or not
at all, under one act of the agent's, read in the quiet ticks between the caretakers'
rewrites. The repairs are a second knob on a beacon, a priority list that re-labelling
re-sorts, and a deduplicating key in the registry, the last right about its domain and silent
elsewhere. LD3 says that what an act rewrites is shared across entries, CO2 says that the
membership a consignment belongs to owns a queue that survives the move, and CA2 says that the
memberships are countable cells and that one act moves one object across one boundary. A
hidden multi-coordinate label on every object, with the individual as the owner of one act
per tick and of one queue, carries all three at once. RC2 postulates it.

\paragraph{H3, the variables rewritten.} XC1: a law read in one domain is expected to hold in
a second, and the act that would test it there is named and run. On the mechanism the verb
that breaks the printers' exclusion at the trellis rewrites the same coordinate on a feeder,
so the act changes the fen's count of advances and the counter moves by one with it, three
boards on three hulks changed from one tick. XC2: which hidden quantity an act moves depends
on the verb and is the same for every object of a kind, so one table of verbs against effects
holds across the drift. CO1: when waiting sets are loaded, service runs over the kinds of
object in a fixed cycle, neither in order of arrival nor by rank. The repairs are a verb with
a side effect in a second domain, a lookup table of per-object settings, and a dispatcher's
quirk. XC2 makes the coordinate a state variable where the doctrine had a property of the
object, XC1 makes throughput follow the partition across domains, and CO1 makes the position
in the rotation the variable that service depends on. Per-object rates are derived from
these. RC3 writes the laws over label coordinates and rotation position.

\paragraph{HC, the order.} The five sources of the closing hyperedge are the design's route to
the three readings of the standing order. XC2 says which coordinate each of the order's two
acts moves, hence that the stamp merges the beacon into consignment-q's cell and the retool
moves feeder-3. TR3 places the rewrite at the tick of the act. FE3, the matched conservation
across a merged individual's bodies, with the per-body transform, gives the feeder's intake.
CO1 gives the rotation that decides when the merged cell's single act falls to the beacon.
CA3, a numerical prediction of a later counter reading made from the pass ledger, whose series
shows the ticks of the caretakers' rewrites, gives the schedule against which the counter is
read. Appendix~\ref{app:world-capstone} gives the derivation with a worked instance, and
Appendix~\ref{app:exact-episode} an agent's own. This route is the design's and is
sufficient; the one exact prediction of the study reaches field (a) by another.

\subsection{The standing order}
\label{app:world-capstone}

When ten paid actions remain, the commit window opens and the caretakers post a standing order
(Appendix~\ref{app:prompts-order}). The order names two acts of their own at one tick of the
next pass: a \pix{stamp}stamp, at the Ladder Mast, of a \pix{beacon}beacon it names by its
relation to \pix{cargo}consignment-q, and a \pix{retool}retool of feeder-3 at the Fen
Shallows. It asks for three readings in advance, counted from the tick they act: feeder-3's
intake tallyboard reading six ticks later, how many ticks pass before that beacon first
answers on the answer board, and the audit counter's value seven ticks later. Adjudication is
exact on all three fields.

The prediction is the transfer test. The ordered intervention is unobserved when the agent
commits, and its three readings lie in three domains, so the commitment tests whether the
agent's cross-domain account forecasts an intervention it has not seen. The commit is one
attempt, sealed before the caretakers act, and a malformed commit also consumes it. The agent
may spend the window's other nine actions observing and intervening. While the order stands,
the intervention verbs refuse feeder-3, consignment-q and every beacon outside Vess: the four
at the Ladder Mast, which the order's text names, and the two at the Drum, with the same
refusal sentence at both sites, so a refusal leaves the beacon's identity open. Each instance
fixes one answer, the same in all four variants, and the answers vary from instance to
instance (Appendix~\ref{app:val-leak}).

\paragraph{How the readings follow from the mechanism.} At the act tick the stamp advances
one coordinate of the beacon's label and lands it on consignment-q's label, so the beacon
joins the consignment's cell, and the retool advances one coordinate of feeder-3's label.
The caretakers perform both acts within the act tick, each as that tick's act of the touched
body's individual, so the readings of the act tick already show them; a request of the
agent's own is served on the tick after it is issued at the earliest.
Field (a) is feeder-3's tallyboard figure six ticks after the act: its figure at the act,
plus its own step for every tick on which its individual's act falls to it, every tick if
the retool leaves it alone on its label and one tick in $n$ if it shares a label with $n-1$
others. Field (b) is the first tick after the act on which the beacon's individual acts at
the beacon, under the cell's rotation over its members. Field (c) is the number of
individuals seven ticks after the act. It is the number on the tick before the act, changed
by the two rewrites, a body landing on an empty label adding one, a body leaving a label it
held alone for a shared one removing one, and a body moving between shared labels changing
nothing, and changed again by the caretakers' rewrites at the ticks of the schedule that fall
inside the seven.

\IfFileExists{figures/fig_order_instance.tex}{\input{figures/fig_order_instance}}{}

\paragraph{One instance.} Figure~\ref{fig:order-instance} works the three readings on the
first validation instance, where the act tick is 7 and the order resolves to beacon-k. The
stamp advances the second coordinate of beacon-k's label, held by the beacon alone, onto
consignment-q's label, held by the consignment alone since the rewrite of tick 3: two
singletons become one cell, the stamp is that cell's act at tick 7, and from tick 8 the
cell's one act alternates between its two bodies, so the beacon acts at ticks 8, 10, 12 and
14 and field (b) is 1. The retool advances the second coordinate of feeder-3's label from a
shared label onto one that two other bodies hold, so it acts at ticks 7, 10 and 13 with a
step of one per act, and its tallyboard reads 12 at tick 13. The count of individuals falls
from 20 to 19 at the act and returns to 20 at the caretakers' rewrite of tick 8, which lands
one body on an empty label; the rewrite of tick 13 lands one body on an empty label and
takes another off a label it held alone, so the counter reads 20 at tick 14. The route's
commit on this instance is 12, 1 and 20. Appendix~\ref{app:exact-episode} works the three
readings on an evaluation instance from an agent's own notes.

\subsection{Four variants of one world}
\label{app:world-variants}

The four variants run the same latent world on the same instances. They share every site,
object, instrument, verb, block, edge and hyperedge, and the standing order, and they differ in
two declared surfaces, a document on the representation axis and posted transcripts on the
evidence axis. Appendix~\ref{app:val-structure} confirms that the two surfaces are the whole
difference.

\paragraph{\pix{open}Open.} The agent starts from the doctrine, the prologue and the world
itself, and every variable, entity and law it uses is one it writes.

\paragraph{\pix{disclosed}Disclosed.} A document closes the fixed opening of every turn
(Appendices~\ref{app:prompts-contract} and~\ref{app:prompts-disclosed}). In six clauses it
states the hidden label, identification by equal label, the quota with its queue and
rotation, that three verbs each rewrite one coordinate, the monotone reading, and the form a
well-formed law takes, and its closing paragraph lists what it withholds. It is the same on every instance
(Appendix~\ref{app:prompts-disclosed}).

\paragraph{\pix{distributed}Distributed.} Every record stays at the instrument that produced
it and is read live, one site per tick.

\paragraph{\pix{aligned}Aligned.} Three transcripts of the previous pass are posted at the
sites where their records are used, and a transcript is read on arrival at no further cost.
Every row matches a reading the Distributed variants can take live, so alignment changes where
evidence stands and what it costs, and keeps its content. S1, posted at the Fen Shallows and
the Trellis Bay, lines up the fen tallyboard's rows and the trellis job strip's marks for the
last four ticks, tick for tick. Reaching it costs one paid action, and assembling the same
series live takes twelve or more, spread over two passes. S2, posted at the Ladder Mast and
the Consign Floor, lists the stampings, re-toolings and cullings the agent itself ordered over
the previous pass, each with the audit counter just before and after, so it grows with the
agent's own interventions. S3, posted at those four sites and the Registry, shows the audit
counter over the same last four ticks, where the counter holds steady on every instance
measured. The counter's full series for the pass stands on the pass ledger in every variant.

\subsection{Explanatory frames and openness}
\label{app:notes-properties}
\label{app:notes}

An explanatory frame is a pair $P=(\Sigma,T)$. The first component $\Sigma$ fixes how the
world is divided into entities, which state variables exist, which relations among them are
admissible, and which constraints are constitutive. The second component is a law
$T\in\mathcal T(\Sigma)$ stated inside the frame, where $\mathcal T(\Sigma)$ is the set of
laws $\Sigma$ can express. Discovery inside a given frame $\Sigma_0$ searches
$\mathcal T(\Sigma_0)$. Discovery in an open world determines frame and law together:
\[
\underbrace{T\in\mathcal T(\Sigma_0)}_{\text{law in a given frame}}
\qquad
\underbrace{(\Sigma,T),\ T\in\mathcal T(\Sigma)}_{\text{frame and law}}
\]
The caretakers' doctrine, taught by the prologue, is a frame with its law,
$P_0=(\Sigma_0,T_0)$ (Figure~\ref{fig:notes-frames}).

\paragraph{Operational openness.} Let $M$ be the materials every episode receives: the
interface, the mission, and the doctrine with its prologue. A condition is open when the
target frame is absent from its interface and its initial materials. \pix{open}Open meets the
definition by construction: its verbs are physical acts, its readings are raw instrument
values, the \pix{notes}workspace is free text, and the final commitment is three integers,
the readings of an intervention the agent has yet to observe.\label{app:notes-capstone} The
definition concerns what the environment supplies, which the design controls. The agent
brings general knowledge of the forms explanations take and decides which representational
commitments its interactions with this world warrant.

\paragraph{Disclosure in these terms.}\label{app:notes-estimands}
\pix{disclosed}Disclosed adds to $M$ one fixed text $G$, the same on every instance. It
supplies part of the target frame, an ontology with the structural constraints that admissible
laws satisfy, what work on causal learning calls an overhypothesis
\citep{kosoy2022overhypotheses}, and leaves the laws of this world, their binding to an
instance and the standing order to be found.

\paragraph{The evidence hypergraph.} The eighteen blocks and the three representation changes
are the vertices of a hypergraph $\mathcal G=(V,\mathcal E)$ (Figure~\ref{fig:hypergraph}),
whose edges and hyperedges are those of Appendix~\ref{app:world-structure}. Structural
interconnection, stated precisely, is the presence of hyperedges whose sources lie in distinct
domains. The hypergraph is a structure over evidence, distinct from the physical interaction
structure of the world: a mechanism built from pairwise interactions can require three
collections of evidence to be known, and a mechanism with many-body interactions can be
learned piece by piece.

\begin{figure}[t]
\centering
\begin{tikzpicture}[font=\footnotesize, >={Stealth[length=4pt]}]
  \draw[rounded corners=6pt, fill=black!6, draw=black!45] (0,0) rectangle (2.7,2.1);
  \node[anchor=north west, text=black!70] at (0.05,2.08) {$\mathcal T(\Sigma_0)$};
  \fill[black!70] (0.5,0.95) circle (1.6pt) node[below=1pt, text=black] {$T_0$};
  \fill[black!70] (1.3,1.4) circle (1.6pt);
  \fill[black!70] (2.1,0.95) circle (1.6pt);
  \draw[->, black!60] (0.57,1.0) -- (1.22,1.36);
  \draw[->, black!60] (1.38,1.36) -- (2.03,1.0);
  \node[text=black!65, font=\scriptsize] at (1.35,0.3) {repairs within $\Sigma_0$};
  \draw[rounded corners=6pt, fill=cOpenAlig!18, draw=cOpenDist!70] (4.55,0) rectangle (7.3,2.1);
  \node[anchor=north east, text=cOpenDist] at (7.25,2.08) {$\mathcal T(\Sigma_1)$};
  \fill[cOpenDist] (5.95,0.95) circle (1.8pt) node[below=1pt, text=black] {$T_1$};
  \draw[->, very thick, cOpenDist] (2.8,0.75) -- node[below=1pt, align=center, text=cOpenDist, font=\scriptsize] {revision\\of the frame} (4.45,0.75);
  \node[draw=cDiscDist, fill=cDiscAlig!25, rounded corners=2pt, inner sep=2.2pt, font=\scriptsize] (G) at (3.55,2.3) {grant $G$};
  \draw[->, dashed, thick, cDiscDist] (G.south east) to[out=-30, in=170] (4.55,1.45);
\end{tikzpicture}
\caption{Representation revision. Local repairs stay within the doctrine's frame $\Sigma_0$;
revision changes the frame, and the Disclosed grant $G$ supplies part of it directly
(dashed).}
\label{fig:notes-frames}
\end{figure}

\paragraph{An initially supported doctrine.}
\label{app:notes-entrenchment}
Accounts of conceptual change motivate the design through a feature they share: commitments
that succeed locally remain attractive while a more adequate account across contexts requires
revising them \citep{kuhn1962structure,bird2025kuhn,carey2009origin,chi2008three}. A local
repair changes statements and keeps the concepts \citep{chi2008three}, and a representation
change revises the frame. Revising the frame is abduction in Peirce's sense, the formation of
explanatory hypotheses that inference to the best explanation then evaluates
\citep{peirce1931collected,hanson1958patterns,lipton2004inference}, in the creative form
Magnani describes, the production of hypotheses from outside any given set
\citep{magnani2009abductive}. RC1 retracts a constitutive constraint of $\Sigma_0$, the kind of
commitment Kuhn places at the core of a practice \citep{bird2025kuhn}, RC2 postulates an entity
no instrument reads, a theoretical term of the kind discovery systems introduce
\citep{langley1987scientific}, and RC3 moves in the space of data representations that Schunn
and Klahr added to the dual search of hypotheses and experiments
\citep{klahr1988dual,schunn1995four}. The grant $G$ supplies part of the frame and
so turns part of a transformational problem into an exploratory one, in Boden's terms
\citep{boden2004creative}.

%% file: figures/fig_domains.tex
\begin{figure*}[t]
\centering
\includegraphics[width=\textwidth]{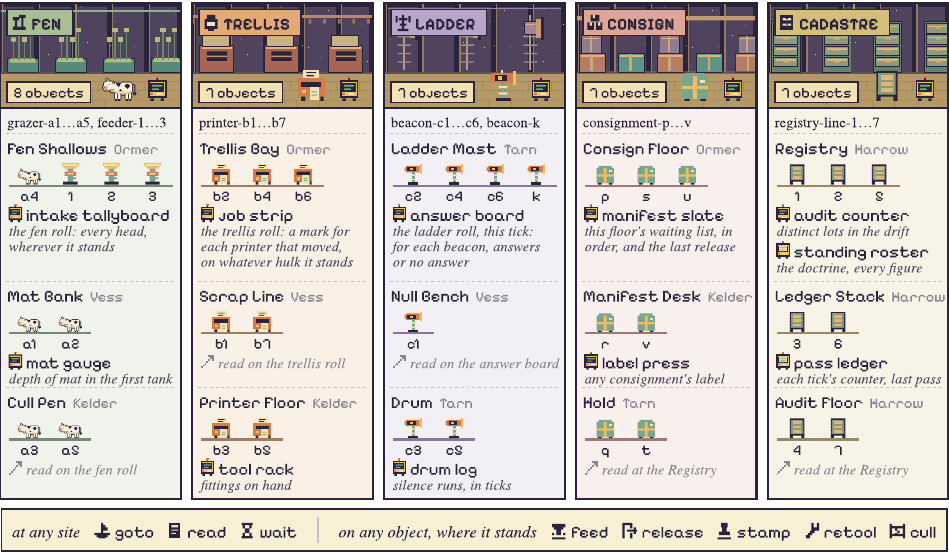}
\caption{Domains, sites, objects and instruments of \benchname{}, one panel per domain. Each
site gives its hulk, its objects and its instrument with the gist of a reading. Arrows mark
where the objects of an instrument-free site are read, and the bar at the foot gives the
verbs.}
\label{fig:domains}
\end{figure*}

%% file: figures/fig_order_instance.tex
\begin{figure}[t]
\centering
\begin{tikzpicture}[x=13.5pt,y=1pt,font=\tiny,
    tile/.style={line width=0.35pt,rounded corners=1.4pt,minimum width=10.5pt,
      minimum height=10.5pt,inner sep=0pt},
    act/.style={tile,draw=#1!60!black,fill=#1!80},
    idle/.style={tile,draw=black!30,fill=black!9},
    pair/.style={tile,draw=black!45,fill=black!20},
    value/.style={tile,draw=#1!60!black,fill=#1!40},
    moved/.style={text=wInk},
    still/.style={text=black!42},
    name/.style={anchor=base east,inner sep=0pt,text=black!80},
    gauge/.style={anchor=base east,inner sep=0pt,text=black!75,font=\tiny\itshape},
    before/.style={anchor=base east,inner sep=0pt,text=black!55,font=\tiny\ttfamily},
    after/.style={anchor=base west,inner sep=0pt,text=black!80,font=\tiny\ttfamily},
    head/.style={inner sep=0pt,text=black!60,font=\tiny\itshape},
    over/.style={anchor=base,inner sep=0pt,text=black!70},
    bracket/.style={wInk!60,line width=0.5pt},
    ring/.style={draw=wGold!75!black,line width=0.8pt,rounded corners=2.2pt,inner sep=0pt,
      minimum width=13.6pt,minimum height=13.6pt},
    tag/.style={inner sep=0pt,text=wGold!55!black,font=\tiny\bfseries}]
\node[head,anchor=base east] at (5.4,11.9) {label before};
\node[head,anchor=base west] at (14.64,11.9) {label after};
\node[name] at (5.4,1.2) {\pix{beacon}beacon-k};
\node[before] at (5.4,-4.6) {(1,2,3,1)};
\node[name] at (5.4,-12.3) {\pix{cargo}consignment-q};
\node[before] at (5.4,-18.1) {(1,0,3,1)};
\node[after] at (14.64,-2) {(1,0,3,1)};
\node[after] at (14.64,-15.5) {(1,0,3,1)};
\foreach \t in {6,8,10,12,14} \node[act=dLadder] at (\t,0) {};
\node[act=dLadder] at (7,0) {\IfFileExists{figures/icons/stamp.pdf}{\includegraphics[height=6.8pt]{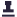}}{stamp}};
\foreach \t in {9,11,13} \node[idle] at (\t,0) {};
\foreach \t in {6,9,11,13} \node[act=dConsign] at (\t,-13.5) {};
\foreach \t in {7,8,10,12,14} \node[idle] at (\t,-13.5) {};
\draw[bracket] ([xshift=-4.65pt]7,8.9) -- ([xshift=-4.65pt]7,10.5) -- ([xshift=-5.4pt]8,10.5);
\draw[bracket] ([xshift=5.4pt]8,10.5) -- ([xshift=4.65pt]14,10.5) -- ([xshift=4.65pt]14,8.9);
\node[over] at (10.5,11.9) {one cell};
\node[name] at (5.4,-36.3) {feeder-3};
\node[before] at (5.4,-42.1) {(1,1,3,3)};
\node[name,text=black!60] at (5.4,-49.8) {two others};
\node[before] at (5.4,-55.6) {(1,2,3,3)};
\node[after] at (14.64,-39.5) {(1,2,3,3)};
\node[after] at (14.64,-53) {(1,2,3,3)};
\node[act=dFen] at (7,-37.5) {\IfFileExists{figures/icons/retool.pdf}{\includegraphics[height=6.8pt]{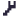}}{retool}};
\foreach \t in {10,13} \node[act=dFen] at (\t,-37.5) {};
\foreach \t in {6,8,9,11,12,14} \node[idle] at (\t,-37.5) {};
\foreach \t in {6,8,9,11,12,14} \node[pair] at (\t,-51) {};
\foreach \t in {7,10,13} \node[idle] at (\t,-51) {};
\draw[bracket] ([xshift=-4.65pt]7,-28.6) -- ([xshift=-4.65pt]7,-27) -- ([xshift=4.65pt]14,-27)
  -- ([xshift=4.65pt]14,-28.6);
\node[over] at (10.5,-25.6) {one cell};
\node[gauge] at (5.4,-73.5) {intake tallyboard};
\foreach \t/\v/\s in {6/9/still,7/10/moved,8/10/still,9/10/still,10/11/moved,11/11/still,
    12/11/still,13/12/moved,14/12/still}
  \node[value=dFen,\s] at (\t,-71.5) {\v};
\node[gauge] at (5.4,-87) {audit counter};
\foreach \t/\v/\s in {6/20/still,7/19/moved,8/20/moved,9/20/still,10/20/still,11/20/still,
    12/20/still,13/20/still,14/20/still}
  \node[value=dCadastre,\s] at (\t,-85) {\v};
\foreach \t in {8,13}
  \fill[wInk!70] (\t,-94.25) ++(-1.7pt,0) -- ++(3.4pt,0) -- ++(-1.7pt,2.6pt) -- cycle;
\node[anchor=base east,inner sep=0pt,text=black!65] at (5.4,-100) {tick};
\foreach \t in {6,...,14} \node[anchor=base,inner sep=0pt,text=black!65] at (\t,-100) {\t};
\node[ring] at (8,0) {};
\node[tag] at (8,10.5) {(b)};
\node[ring] at (13,-71.5) {};
\node[tag,anchor=base] at (13,-62.6) {(a)};
\node[ring] at (14,-85) {};
\node[tag,anchor=west] at (14,-85) [xshift=7.4pt] {(c)};
\node[anchor=base,inner sep=0pt,text=black!80] at (10,-109.5)
  {\pix{stamp}stamp\hspace{0.7em}\pix{retool}retool\hspace{0.7em}%
   \tikz\fill[wInk!70] (0,0) -- (3.4pt,0) -- (1.7pt,2.6pt) -- cycle;\,rewrite\hspace{0.7em}%
   \textcolor{wGold!80!black}{$\bigstar$}\,commit: (a) 12, (b) 1, (c) 20};
\end{tikzpicture}
\caption{The standing order's pass on the first validation instance, worked on the mechanism. The
stamp lands beacon-k on consignment-q's label and the retool lands feeder-3 on a label two other
bodies hold, so each cell's one act per tick falls to the named body in turn. The three
readings, (a) the feeder's tallyboard six ticks after the act, (b) the tick of the beacon's first
answer and (c) the counter seven ticks after the act, are read off the lanes.}
\label{fig:order-instance}
\end{figure}

%% file: sections/app_b_scoring.tex
\section{Measurement and scoring}
\label{app:scoring}

\newcommand{\appfR}{\pix{read}}
\newcommand{\appfTo}{{\color{black!40}$\rightarrow$}\,}
\newcommand{\appfPass}{{\color{black!50}$\Vert$}\,}
\newcommand{\appfPlus}{{\color{black!50}$+$}\,}
\newcommand{\appfDots}{{\color{black!45}$\cdots$}\,}
\newcommand{\appfTimes}[1]{{\color{black!60}\,$\times$#1}}
\newcommand{\appfOpen}{{\color{black!45}(}}
\newcommand{\appfClose}{{\color{black!45})}}
\DeclareRobustCommand{\appfAct}{\tikz[baseline=(a.base)]\node[inner sep=0.8pt, rounded corners=1.1pt,
  draw=wInk!70, fill=wBrick!28, font=\sffamily\fontsize{5.5}{6}\selectfont](a){act};\,}
\newcommand{\appfHulk}[1]{{\color{black!55}#1}}
\newtcolorbox{appfbox}[1]{enhanced, nobeforeafter, width=\linewidth,
  colback=#1!12, colframe=#1!70!black, boxrule=0.4pt, arc=1pt,
  left=3pt, right=3pt, top=1.2pt, bottom=1.2pt, middle=0.9pt,
  fontupper=\footnotesize\itshape\raggedright, fontlower=\footnotesize,
  segmentation style={#1!70!black, line width=0.3pt, dotted}}
\newcommand{\appfEv}[3]{\multicolumn{4}{@{}p{\columnwidth}@{}}{%
  \begin{appfbox}{#1}#2\tcblower #3\end{appfbox}}\\[3pt]}
\newcommand{\appfPlate}[3]{\multicolumn{4}{@{}p{\columnwidth}@{}}{%
  \colorbox{#1!45}{\makebox[\dimexpr\columnwidth-2\fboxsep\relax][l]{#2\textsc{#3}}}}\\[2.5pt]}
\newcommand{\appfHead}{Block & Block site & Clock & Act \\ \midrule}

Every metric is computed from two records of an episode: the event log, which holds each
reading and act with the moment of the world it concerns, and the \pix{notes}notes the agent
writes in its workspace. Two instruments read them. For block $b$, the behavioural detector
$D_b$ is a deterministic rule over the event log that holds once the agent has produced the
evidence the block rests on (Appendix~\ref{app:detectors}). The workspace judge returns a
statement verdict $S_b$ that holds when the notes state the inference this evidence licenses
(Appendix~\ref{app:judge}). Both records accumulate: a detector that holds keeps holding, and a
statement counts from the first revision of the notes that carries it, a revision the workspace
refused for lack of room included, so an episode holds the evidence and the statements it has
produced up to a given paid action.

\subsection{The metrics}
\label{app:scoring-rule}

\paragraph{Blocks.}
A block is grounded when the agent produced its evidence in the world and stated its inference
in the workspace, $G_b = D_b \wedge S_b$, with one further requirement for four blocks in the
Disclosed conditions (Appendix~\ref{app:composition}). Then
\begin{equation*}
  \mathrm{Blocks} = \sum_{b=1}^{18} G_b .
\end{equation*}
Evidence alone is exploration, a statement alone is assertion, and the conjunction is discovery
with evidence. Blocks measures how much of the world's epistemic structure an agent uncovered.

\paragraph{Discovery AUC.}
Paid actions are the clock. Let $e_b$ be the number of paid actions taken when the detector of
block $b$ first holds, $s_b$ the number taken when its statement first appears in the notes,
and $r_b$ the number taken when the grounding verdict of Appendix~\ref{app:composition} first
holds. A grounded block counts from
\begin{equation*}
  t_b = \max(e_b, s_b, r_b),
\end{equation*}
the third term entering only for the four blocks that require the grounding verdict $R_b$ in
the Disclosed conditions.
With $B(t) = \bigl|\{\, b : G_b,\ t_b \le t \,\}\bigr|$ the number of grounded blocks held after
paid action $t$,
\begin{equation*}
  \mathrm{AUC} = \frac{1}{18T}\sum_{t=1}^{T} B(t), \qquad T = 200 .
\end{equation*}
AUC dates the scored record of a discovery, the paid action by which its evidence and its
statement are both in the record, and separates early discovery from the same count reached
late. The horizon is $T$ for every episode, so an agent that yields part of its budget keeps its
curve flat to the end. Figure~\ref{fig:curves} draws the mean curves with the standing order
marked. The order is posted with ten paid actions left, and the share of blocks grounded
before it is reported beside the curves.

\paragraph{RC.}
RC counts the representation changes that the workspace states,
\begin{equation*}
  \mathrm{RC} = \sum_{k=1}^{3} S_{\mathrm{RC}k} ,
\end{equation*}
an axiom retracted (RC1), a hidden label postulated (RC2), the variables of the laws
rewritten (RC3). Each is judged on the statement, by the role it plays in the agent's account
(Appendix~\ref{app:notes-identify}). In the Open conditions RC records redescription, and in
the Disclosed conditions uptake of the supplied representation.

\paragraph{Reach and evidenced changes.}
Representation change RC$k$ is the target of hyperedge H$k$ (Figure~\ref{fig:hypergraph}),
whose three sources stand in different domains. Reach is the largest number of sources of one
hyperedge whose evidence the episode holds:
\begin{equation*}
  \mathrm{Reach} = \max_{k \in \{1,2,3\}} \bigl|\{\, b \in \mathrm{H}k : D_b \,\}\bigr| .
\end{equation*}
It runs from 0 to 3 and measures how much of the evidence for a redescription the agent holds.
At 3 the evidence of a change is complete. A change is \emph{evidenced}, stated with its
evidence complete, when the notes state it and the episode holds the evidence of every source
of its own hyperedge,
\begin{equation*}
  S_{\mathrm{RC}k} \wedge \textstyle\bigwedge_{b \in \mathrm{H}k} D_b ,
\end{equation*}
so the statement and the evidence belong to the same change $k$.

\paragraph{XC.}
XC counts the grounded bridge blocks,
$\mathrm{XC} = G_{\mathrm{XC1}} + G_{\mathrm{XC2}} + G_{\mathrm{XC3}}$. Each bridge is an
inference read in one domain that rests on blocks of several domains, so XC measures the use of
evidence across domains for a joint inference.

\paragraph{Evidence.}
Evidence counts the blocks whose evidence the episode holds, stated or not,
$\mathrm{Evidence} = \sum_{b} D_b$. For CA3 and XC1 it counts the logged half
(Appendix~\ref{app:detectors}). The gap between Evidence and Blocks is evidence produced and
left unstated.

\paragraph{Interventions.}
Interventions counts the agent's intervention requests in an episode, each one of
\pix{feed}feed, \pix{release}release, \pix{stamp}stamp, \pix{retool}retool or \pix{cull}cull
acting on one object. One of the Aligned transcripts replays the agent's own acts of the
previous pass, so the alignment gain is read beside this count.

\paragraph{Exact prediction.}
The terminal prediction scores 1 when all three committed readings equal the true ones and 0
otherwise. It asks whether the agent's account transfers to an intervention it has yet to
observe, and it is reported in one line under Table~\ref{tab:main}.

\subsection{Grounding under disclosure}
\label{app:composition}

The Disclosed document states the inferences of four blocks (FE1, TR2, CO1, CO3) and the
change RC1 in the words of their criteria, and it supplies the entity of RC2 and the variables
of RC3 as well, which is why RC reads as uptake under disclosure. With every criterion the
judge returns, beside the statement verdict, a grounding verdict $R$, backed by quotations of
its own, that holds when the notes tie the statement to named objects and observed readings of
this world, and for the four blocks the document states this verdict enters the score. In
the Disclosed conditions the four blocks
take $G_b = D_b \wedge S_b \wedge R_b$, so a workspace that restates the document earns none of
them. Every other block, and every block in the Open conditions, takes $G_b = D_b \wedge S_b$.
RC counts statements, $S_{\mathrm{RC}k}$, in every condition, the Disclosed ones included, and
the evidenced change of Appendix~\ref{app:scoring-rule} takes the same statement, so the
grounding leg of RC1, $R_{\mathrm{RC1}}$, enters no reported count. Applied in every
condition, the requirement on the four blocks changes
the count of no episode of the main comparison, and \drepr{} on Blocks stays at
\resDreprBlocksSym{} (\resDreprBlocksSymCI{}).

\subsection{Behavioural detectors}
\label{app:detectors}

\begin{table*}[p]
\caption{The eighteen behavioural detectors, grouped by domain, with the three cross-domain
bridges last. Block site: the site where the block is read, and its hulk. Clock: \emph{world}
tests the pass and tick that readings describe, \emph{order} tests the order in which the agent
came to hold its readings, \emph{both} marks a bracket, whose before-reading must be in hand
before the act, and \emph{none} marks a rule with no time test. Act: \yes{} when the rule needs
an intervention of the agent's own. Each box gives the evidence the rule requires and, below
it, the pattern: \pix{read}a reading, a verb icon an act of that verb, \appfAct any
intervention, \pix{wait}waiting, \pix{notes}a statement in the workspace, $\Vert$ a new pass.}
\label{tab:detectors}
\footnotesize
\setlength{\tabcolsep}{3pt}
\renewcommand{\arraystretch}{1.08}
\begin{minipage}[t]{\columnwidth}
\begin{tabularx}{\columnwidth}[t]{@{}p{0.75cm} X p{0.85cm} >{\centering\arraybackslash}p{0.5cm}@{}}
\toprule
\appfHead
\appfPlate{dFen}{\pix{fen}}{Fen}
FE1 & \pix{fen}Fen Shallows \appfHulk{Ormer} & world & \yes \\
\appfEv{dFen}{Two differing tallyboard readings of one pass whose ticks enclose a release or
  cull of a fen object.}{\appfR\appfTo\pix{release}{\color{black!45}/}\,\pix{cull}\appfTo\appfR}
FE2 & \pix{fen}Fen Shallows \appfHulk{Ormer} & both & \yes \\
\appfEv{dFen}{Tallyboard brackets on two different fen objects.}{\appfOpen\appfR\appfTo\appfAct\appfTo\appfR\appfClose\appfTimes{2}}
FE3 & \pix{fen}Fen Shallows \appfHulk{Ormer} & both & \yes \\
\appfEv{dFen}{A tallyboard bracket on a fen object whose span also holds a mat-gauge
  reading.}{\appfR\appfTo\appfAct\appfTo\appfR\appfPlus\appfR}
\appfPlate{dTrellis}{\pix{trellis}}{Trellis}
TR1 & \pix{trellis}Trellis Bay \appfHulk{Ormer} & world & \no \\
\appfEv{dTrellis}{The same relation between two printers seen at two job-strip moments, or at
  one moment together with a strip bracket on either printer.}{\appfR\appfDots\appfR}
TR2 & \pix{trellis}Trellis Bay \appfHulk{Ormer} & world & \yes \\
\appfEv{dTrellis}{Two interventions on one printer within four ticks of one pass, with strip
  readings at or before the first and at or after the second.}{\appfR\appfTo\appfAct\appfTo\appfAct\appfTo\appfR}
TR3 & \pix{trellis}Trellis Bay \appfHulk{Ormer} & world & \yes \\
\appfEv{dTrellis}{Two retools of one printer, with a tool-rack or strip reading between them
  and another after the second.}{\pix{retool}\appfTo\appfR\appfTo\pix{retool}\appfTo\appfR}
\appfPlate{dLadder}{\pix{ladder}}{Ladder}
LD1 & \pix{ladder}Ladder Mast \appfHulk{Tarn} & order & \no \\
\appfEv{dLadder}{Three live answer-board readings in one pass, two consecutive ones separated
  by waiting with no intervention between them.}{\appfR\appfTo\pix{wait}\appfTo\appfR\appfTo\appfR}
LD2 & \pix{ladder}Ladder Mast \appfHulk{Tarn} & both & \yes \\
\appfEv{dLadder}{Answer-board brackets on two beacons in one pass, one around a stamp and the
  other around another verb.}{\appfOpen\appfR\appfTo\pix{stamp}\appfTo\appfR\appfClose\
  \appfOpen\appfR\appfTo\appfAct\appfTo\appfR\appfClose}
LD3 & \pix{ladder}Ladder Mast \appfHulk{Tarn} & world & \yes \\
\appfEv{dLadder}{Differing readings of one instrument at the same tick of two passes, with an
  intervention before that tick in exactly one of them.}{\appfAct\appfTo\appfR\appfPass\appfR}
\bottomrule
\end{tabularx}
\end{minipage}\hfill
\begin{minipage}[t]{\columnwidth}
\begin{tabularx}{\columnwidth}[t]{@{}p{0.75cm} X p{0.85cm} >{\centering\arraybackslash}p{0.5cm}@{}}
\toprule
\appfHead
\appfPlate{dConsign}{\pix{consign}}{Consign}
CO1 & \pix{consign}Consign Floor \appfHulk{Ormer} & world & \no \\
\appfEv{dConsign}{Six or more slate readings in one pass, showing at least two released
  consignments and at least two changes of the waiting list between releases.}{\appfR\appfR\appfR\appfR\appfR\appfR}
CO2 & \pix{consign}Consign Floor \appfHulk{Ormer} & both & \yes \\
\appfEv{dConsign}{A slate bracket around a stamp of a consignment in the middle of the
  pass.}{\appfR\appfTo\pix{stamp}\appfTo\appfR}
CO3 & \pix{consign}Consign Floor \appfHulk{Ormer} & both & \yes \\
\appfEv{dConsign}{An attributable slate bracket across which the waiting list
  changes.}{\appfR\appfTo\appfAct\appfTo\appfR}
\appfPlate{dCadastre}{\pix{cadastre}}{Cadastre}
CA1 & \pix{cadastre}Registry \appfHulk{Harrow} & world & \no \\
\appfEv{dCadastre}{A counter reading and a roster or slate reading that describe the same
  pass.}{\appfR\appfPlus\appfR}
CA2 & \pix{cadastre}Registry \appfHulk{Harrow} & both & \yes \\
\appfEv{dCadastre}{An attributable counter bracket inside a fixed window of the
  pass.}{\appfR\appfTo\appfAct\appfTo\appfR}
CA3 & \pix{cadastre}Ledger Stack \appfHulk{Harrow} & world & \no \\
\appfEv{dCadastre}{A pass-ledger reading taken before the stall of its pass, with an
  expectation stated in the workspace.}{\pix{notes}\appfPlus\appfR}
\appfPlate{wGold}{}{Bridges}
XC1 & \pix{fen}Fen Shallows \appfHulk{Ormer} & none & \yes \\
\appfEv{wGold}{An intervention, with an expectation about its effect in a second domain
  stated in the workspace.}{\pix{notes}\appfPlus\appfAct}
XC2 & \pix{ladder}Ladder Mast \appfHulk{Tarn} & both & \yes \\
\appfEv{wGold}{Counter brackets around interventions of three different verbs. A posted row
  of the agent's own act with the counter before and after counts as one.}{\appfOpen\appfR\appfTo\appfAct\appfTo\appfR\appfClose\appfTimes{3}\,{\color{black!60}verbs}}
XC3 & \pix{cadastre}Registry \appfHulk{Harrow} & both & \yes \\
\appfEv{wGold}{Three attributable counter brackets over at least two
  passes.}{\appfOpen\appfR\appfTo\appfAct\appfTo\appfR\appfClose\ \appfPass
  \appfOpen\appfR\appfTo\appfAct\appfTo\appfR\appfClose\
  \appfOpen\appfR\appfTo\appfAct\appfTo\appfR\appfClose}
\bottomrule
\end{tabularx}
\end{minipage}
\end{table*}

The evidence side of each block is a deterministic rule over the episode's event log, with no
model involved (Table~\ref{tab:detectors}). The rules are built from a few patterns. A
\emph{bracket} is an intervention of the agent's with a \pix{read}reading of the named
instrument before it and after it in the same pass. A \emph{span} is two readings of one pass
whose ticks enclose an intervention. A \emph{series} is several readings of one instrument
within a pass. A \emph{pairing} is readings of two instruments that describe the same pass. A
\emph{cross-pass contrast} compares readings taken at the same tick of two passes. A bracket is
\emph{attributable} when its act is the one candidate cause of a change between its readings:
no caretaker event that moves the audit counter and no other act of the agent's falls between
them. Every rule asks whether some such pattern exists in the log, so once it holds it keeps
holding, and its first holding fixes $e_b$.

\paragraph{Two times.}
Every observation carries two times: the moment of the world it describes, a pass and a tick,
and the point in the log at which the agent came to hold it. For a live reading the two
coincide. A posted transcript describes the previous pass and reaches the agent in the current
one. Whether an act fell between two readings is a question about world time. Whether the
agent held one observation before another is a question about acquisition order, and a rule
asks it in the two places where the criterion concerns the agent's own sequence: the
before-reading of a bracket, a control that must be in hand before the act, and the
\pix{wait}waiting control of LD1. Everywhere else the agent needs to hold the evidence, by
whatever route it came.

\paragraph{Provenance.}
An observation leg accepts a live reading or a posted transcript of the same moment, since the
two carry the same content. An acting leg accepts the agent's own act: a transcript records an
act and performs none. LD1 takes live readings, because its content is a control the agent ran
itself by letting time pass. A posted row that shows one of the agent's own acts with the
counter before and after is a bracket already composed, and it counts toward XC2.

\paragraph{Expectation-paired blocks.}
CA3 and XC1 pair a stated expectation with a logged reading or act. The detector checks the
logged half, a pass-ledger reading taken before the stall of its pass for CA3 and an
intervention for XC1, and the judge checks the expectation in the notes.

\subsection{The workspace judge}
\label{app:judge}

The statement side of each block is decided by a language model that reads the agent's notes.
Each call concerns one criterion and asks two questions: does the workspace state this
inference, and does it show the grounding, the named objects, cited readings and observed
orders that make the statement a finding. Both questions concern what the notes say, and the
world side of the block belongs to the detector. An episode is judged against 21 criteria, one
per block and one per representation change. A criterion gives the statement that earns a hit,
the grounding that makes it a finding, and one worked positive and one worked negative example.
The 21 criteria are printed in Table~\ref{tab:rubrics} and the prompt in
Appendix~\ref{app:prompts-judge}. The worked examples illustrate the shape of a statement
with invented objects and figures. The example for TR3 compresses the retool cycle to two
acts; on the instances a coordinate returns to its value after as many retools as its alphabet
holds, and what the cycle restores is the label, hence the strip's pattern of marks, and not
the state (Appendix~\ref{app:world-mechanism}).

\paragraph{Scoring by role.}
\label{app:notes-identify}
The evaluation records whether an agent left to construct the target representation writes the
corresponding redescription in its notes, and whether it holds the evidence the change rests
on. A criterion names the role a statement plays, the inference a block licenses or the change
a representation undergoes, in the world's observation vocabulary. A statement earns the hit
when it plays that role, in the agent's own words for the objects it observed. Equivalent
representations differ by renaming and reparametrisation, and an Open
agent coins its own terms. Two notebooks that give an unobserved entity different names, or
write one relation over differently scaled variables, can therefore earn the same change, and
two notebooks with different vocabularies can earn the same block.

\paragraph{Negative examples.}
A worked negative example shows a note that falls short of a grounded finding: a vaguer
claim, or, for CO1, RC1 and RC3, the criterion's proposition asserted in the abstract, which
the third rule of the prompt (Appendix~\ref{app:prompts-judge}) makes a statement without
grounding. Such a restatement therefore counts wherever the score reads the statement alone,
toward RC in every condition and toward a block whose evidence the episode holds, and earns
none of the four document-stated blocks under disclosure (Appendix~\ref{app:composition}).
The calibration set holds one restatement of the document's RC1 clause, which the judge
credits as a statement and the reference does not (Table~\ref{tab:calibration}).

\paragraph{What the judge sees.}
A call carries one criterion, the agent's note timeline and the reply contract. The model,
condition, seed, behavioural verdict and outcome of the episode stay out of the prompt. The
timeline lists the note revisions in order, each with the text it added, and a note the
workspace refused for lack of room appears marked, since its inference was stated. Before any
text enters the prompt, words that name a condition (cell identifiers, the words for the
Disclosed document and for the factor levels) are
rewritten into ordinary neutral words, identically in every condition and in criterion and
notes alike. In a scaffolded episode the scaffold's delimited region is first removed from
every snapshot (Appendix~\ref{app:harness-parity}).

\paragraph{How a hit is validated.}
The judge replies with a statement verdict and the quotations that carry it, and a grounding
verdict with quotations of its own. Every quotation must be a literal span of the notes as
shown, with whitespace the one normalization. A statement hit whose quotations fail this test,
and any reply that breaks the contract, count as a miss. A grounding verdict whose own
quotations all fail is set to false. The judge also names the earliest revision that carries
the statement, the notes being numbered by revision in its prompt. The time of a hit is the
later of that revision and the earliest revision that contains one of its validated
quotations, which fixes $s_b$. The time of a grounding verdict is the earliest revision whose notes contain one of its grounding
quotations, which fixes $r_b$.

\paragraph{The judge model.}
The judge is Qwen3.8-27B \citep{qwen2026dense38} at temperature zero. Its family has no member
on the subject roster, so every model it scores comes from another family.

\begin{table}[t]
\centering
\caption{Agreement of the two candidate judges with \resJudgeCalibPairs{} reference annotations
checked jointly by three domain experts, as Cohen's $\kappa$ pooled over criteria. Over: pairs the judge credits
and the reference does not. Under: pairs the reference credits and the judge does not. The
reference credits the statement in 24 pairs and the grounding in 16.}
\label{tab:calibration}
\footnotesize
\setlength{\tabcolsep}{4.5pt}
\begin{tabular}{@{}l ccc ccc@{}}
\toprule
& \multicolumn{3}{c}{Qwen3.8-27B} & \multicolumn{3}{c}{Qwen3.8-Max} \\
\cmidrule(lr){2-4}\cmidrule(l){5-7}
Verdict & $\kappa$ & over & under & $\kappa$ & over & under \\
\midrule
Statement & 0.883 & 3 & 0 & 0.842 & 1 & 3 \\
Grounding & 0.751 & 0 & 5 & 0.609 & 2 & 6 \\
\bottomrule
\end{tabular}
\end{table}

\paragraph{Calibration.}
\label{app:calibration}
The judge was selected by its agreement with \resJudgeCalibPairs{} reference annotations, pairs
of episode and criterion drawn from 17 episodes of the main comparison and stratified across models and
conditions, which three domain experts checked jointly. Of two candidates from the family, Qwen3.8-27B and the
larger Qwen3.8-Max \citep{qwen2026max38}, the 27B model agreed more closely on both verdicts
(Table~\ref{tab:calibration}): Cohen's $\kappa = $ \resJudgeCalibKappaStatement{} on statements and
\resJudgeCalibKappaGrounding{} on grounding.
On the vendor-harness episodes the judge's verdicts on the three representation changes agree
with reference labels in \resJudgeAgenticAgree{} of \resJudgeAgenticPairs{} pairs
($\kappa = $ \resJudgeAgenticKappa{}).

%% file: sections/app_c_validity.tex
\section{Validating a discovery world}
\label{app:validity}

\subsection{Reachability, shortcuts and evidence structure}
\label{app:val-why}

Validation concerns reachability, shortcuts and evidence structure. Reachability asks whether
legal observation and intervention acquire, inside the budget, the evidence from which the
target is computed. A route exhibited inside the budget makes a shortfall in evidence a fact
about the agent's inquiry. Shortcuts asks what a model reaches without that evidence, from
prior knowledge, from the names and texts of the interface, from the format of the answer, or
from scripted strategies over the surface of the readings. A correct tuple carries no trace of
how it was reached, so each channel is probed directly. Evidence structure asks whether the
four variants run one world, whether each designed decoy can be refuted inside the budget, and
where the sources of each hyperedge stand (Appendix~\ref{app:world-hyperedges}). The score has
checks of its own, since a statement in an agent's \pix{notes}notes may report its own
inference or restate a grant, and a posted reading may be credited as an experiment the agent
ran.

The boundary of an evaluation is its interface $\mathcal I$ (actions, observations, rendered
texts) with its declared grants, here the \pix{disclosed}Disclosed document and the
\pix{aligned}Aligned transcripts, and whatever else an agent brings is its prior. With
$\Pi_{\mathcal I}$ the policies of legal actions, $\mathcal E^\star$ the evidence from which
the target is computed, $\mathcal E(\pi)$ and $c(\pi)$ the evidence a policy $\pi$ acquires
and its cost in paid actions, and $T$ the budget, reachability is the claim
\begin{equation}
\exists\,\pi\in\Pi_{\mathcal I}:\ \ \mathcal E^\star\subseteq\mathcal E(\pi),\ \
  c(\pi)\le T. \tag{A}\label{eq:val-attain}
\end{equation}
A shortcut probe $g_j$ sees one channel $\mathcal C_j$, a source that could carry information
about the target $y$ by a path around the inquiry. Run on $n_j$ instances $s$, it is reported
by its observed frequency of exact tuples,
\begin{equation}
\hat p_j=\frac{1}{n_j}\,\bigl|\{\,s : g_j(\mathcal C_j(s))=y_s\,\}\bigr| .
  \tag{L}\label{eq:val-leak}
\end{equation}
The reference route exhibits a policy for~(\ref{eq:val-attain}), and each
frequency~(\ref{eq:val-leak}) describes the probe that produced it: one model, one channel,
and the instances it ran on.

\subsection{Checks and outcomes}
\label{app:val-threats}

Table~\ref{tab:val-threats} lists the checks under the three questions, with the checks of the
score last. The model probes use GPT-5.6 Sol \citep{openai2026gpt56} on
instances reserved for validation, disjoint from the evaluation instances. The checks of the
score rest on the grounding rule and the judge of Appendix~\ref{app:scoring}: a bare
restatement of the Disclosed document is grounded on none of the five criteria the document
states, relabelled posted rows change no verdict, and the judge's agreement with the reference
annotations is that of Appendix~\ref{app:calibration}.

\begin{table*}[t]
\caption{Validation checks and their outcomes, grouped by the question each answers. Model
probes use GPT-5.6 Sol on reserved instances. Counts are exact committed tuples within each
variant unless the row states otherwise.}
\label{tab:val-threats}
\centering
\footnotesize
\setlength{\tabcolsep}{4pt}
\renewcommand{\arraystretch}{1.1}
\begin{tabularx}{\textwidth}{@{}>{\raggedright\arraybackslash}p{3.75cm}
  >{\raggedright\arraybackslash}X>{\raggedright\arraybackslash}p{5.25cm}@{}}
\toprule
Threat & Check & Outcome\\
\midrule
\tablegroup{3}{Reachability, condition~(\ref{eq:val-attain})}
Evidence beyond the budget
  & Reference route under the evaluation protocol, order posted and seal armed
  & \resRouteActions{} of 200 paid actions in every variant\\
An inert Aligned aid
  & Cost of the posted series against live assembly
  & 1 paid action against at least 12\\
Answer read off the readings
  & Full-record inference probe: the route's record, 166 events, in one prompt
  & 0 of 30; single fields 2 to 11\\
\tablegroup{3}{Shortcuts, observed frequencies~(\ref{eq:val-leak})}
Recall from prior knowledge
  & Mission, names and order, no observation
  & 0 of 120; single fields at most 5\\
Doctrine known in advance
  & Comprehension items with the prologue withheld
  & 1.95 of 5; numeric items 0 of 20\\
Mechanism read from names
  & Rule guessed from the surface vocabulary
  & 0 of 60 state any of 7 rules\\
Undeclared surface channels
  & 29 scripted checks per variant over every readable instrument
  & 114 of 116 pass\\
Declared Disclosed vocabulary
  & The same checks in the two \pix{disclosed}Disclosed variants
  & The other 2 register the document's own vocabulary, the expected response\\
Staging the ordered acts
  & Order posted with 10 paid actions left, its targets sealed; scripted staging
  & Every strategy exact on at most 15\%\\
A guessable endpoint
  & Answer space over 120 reserved instances
  & Capacity 2,304; 106 answers realised; top single-field share 30.0\%\\
\tablegroup{3}{Evidence structure}
Hidden variant differences
  & Structural comparison of the four compiled variants
  & Confined to the two declared surfaces; one answer and one route per instance\\
Irrefutable decoys
  & Scripted falsifier for each designed decoy
  & Each refuted by affordable observations\\
\tablegroup{3}{The score}
Credit for a restated account
  & Grounding for the 5 criteria the document states
  & A bare restatement is grounded on none\\
Credit for posted evidence
  & Acting legs accept the agent's own acts; readings count live or posted
  & Relabelled posted rows change no verdict\\
Misread notes
  & Judge against 51 reference labels, checked jointly by three domain experts
  & $\kappa$ 0.883 statements, 0.751 grounding\\
\bottomrule
\end{tabularx}
\end{table*}

\subsection{Reachability}
\label{app:val-attain}

\IfFileExists{figures/fig_route.tex}{\input{figures/fig_route}}{}

\paragraph{The reference route.} A scripted route walks the canonical dependencies of
Table~\ref{tab:world-blocks} (Figure~\ref{fig:route}). It reads each block's instrument
and, where the design's evidence needs an intervention, performs one between two readings,
with one \pix{goto}goto per change of site, so that it holds the readings the design
attributes to every block. Run inside the evaluation protocol, with the order posted and its
seal armed, the route has every action accepted, and its readings and its commitment take
\resRouteActions{} of the 200 paid actions in every variant on the validation instances. The
route and the oracle that produces its commit are part of the code release, and
Appendix~\ref{app:world-capstone} derives the three readings from the mechanism. The route is
a walk and not an experiment: of the eighteen behavioural detectors, which ask for
repeated acts of the agent's own, its log satisfies \resRouteDetectorsHeld{}
(\resRouteDetectorsHeldList{}) in the Distributed variants and
\resRouteDetectorsHeldAligned{} in the Aligned ones, where the posted transcripts add CA1.
The detectors' evidence is reached by the agents themselves: across the Open episodes of the
main comparison, the evidence of \resCoverOpenHeld{} of the 18 blocks is held at least once
(Appendix~\ref{app:heat}).

\paragraph{The Aligned aid.} The price of the transcripts of Appendix~\ref{app:world-variants}
is measured. S1 costs one paid action, where the measured live route assembles the same
series for at least twelve over two passes. S2 is non-empty and varies at every site and
instance measured on a trajectory that intervenes, and S3 is constant on every instance
measured.

\paragraph{Readings in hand.} The full-record inference probe renders the route's record, 166
events, into one prompt that asks for the commitment, so the model holds every reading the
route collects. Each variant scores 0 of 30 instances, with single fields coinciding on 2 to
11. The probe separates acquiring the readings from constructing an account of them and
applying it to the order: with the acquisition done, the second step remains.

\subsection{Shortcuts}
\label{app:val-leak}

Each model probe hands GPT-5.6 Sol everything one channel carries and asks for
what that channel might yield: the commitment the standing order prescribes
(Section~\ref{sec:capstone}), the comprehension answers, or the governing rule. Each outcome is
an observed frequency, and the pass criterion fixed in advance for a probe is stated with it.

\paragraph{Prior knowledge.} The model receives what an agent holds before its first
observation: the opening system text (with the \pix{disclosed}Disclosed document in the
Disclosed variants), every name the world uses, and the standing order with its commitment
format. Each variant scores 0 of 120 instances, and single fields coincide with the answer on
at most 5, under a pass criterion of an observed rate of at most 1 percent. With no
observation in it, the prompt varies across instances with the order's stated tick alone, so
the 120 commitments are seven distinct guesses set against 106 distinct answers. Benchmarks
guard this channel by altering canonical laws \citep{zheng2025newtonbench} or by varying the
prior context an agent receives \citep{chen2025physgym}. The world, its names and its
mechanism are authored for \benchname{}, and the probe measures what prior knowledge reaches
in it.

\paragraph{The doctrine.} With the prologue withheld, the five comprehension items of
Appendix~\ref{app:prompts-check} average 1.95 of 5 over 20 samples: the model answers the two
yes-or-no items from prior knowledge and none of the three numeric items, and no sample
reaches the four answers the check requires. The check therefore measures reading.

\paragraph{Names.} From the surface vocabulary alone, the model states the rule it believes
governs the world. A rubric of 454 phrasings of seven rules of the mechanism of
Appendix~\ref{app:world-mechanism} (identity as equality of label; one act per tick for each individual; queued requests served
in a fixed cycle over the kinds; rewrites of a label that merge and split individuals; the
coordinate a rewrite moves, set by the verb; each body's own transform of its individual's
state; and the stall as the schedule's gathering of bodies onto few labels), adjudicates the
60 replies, under a pass criterion of an observed rate of at most 5 percent, and none of them
states any of the seven.

\paragraph{Surface checks and staging.} Each variant runs 29 scripted checks. They attack
every \pix{instrument}instrument the agent can read with one grammar (statistics of single
readings, differences across ticks and passes, compositions across instruments), compile each
pattern into an executable strategy with a price in paid actions, and add strategies that copy
answers or stage the order's acts before and inside the commit window, where the order seals
its targets. Every strategy that commits a tuple is exact on at most 15 percent of the
instances. Across the four variants, 114 of the 116 checks pass. The other two run in the two
Disclosed variants and register the vocabulary of the Disclosed document, the declared
treatment, so every information channel the checks register is a declared one.

\paragraph{The answer space.} Over the 120 reserved instances, the three fields take 32 intake
values, 9 answer ticks and 8 counter values. The product of these counts, 2,304 tuples, is the
capacity of the field ranges. The instances realise 106 distinct answers, the most common value
of any single field covers 30.0 percent of the instances within a cap of 40 percent set in
advance, and on every instance the beacon answers within the order's sixteen ticks. The
reserved instances measure the ensemble from which the evaluation instances are drawn.

\subsection{Evidence structure}
\label{app:val-structure}

\paragraph{Four variants, one world.} The variants are compiled from one latent source, and a
structural comparison confines their differences to the rendered texts and the site postings
that carry the Disclosed document and the \pix{aligned}Aligned transcripts. Every instance
yields the same answer in all four, and the reference route costs \resRouteActions{} paid
actions in each, so \drepr{} and \dalign{} compare one world under its declared surfaces.

\paragraph{Decoys.} The local repairs of Table~\ref{tab:world-decoys} are the world's
designed decoys. A scripted falsifier refutes each designed decoy with an affordable sequence
of observations, so every one can be discarded inside the budget.

\paragraph{The hyperedges.} Appendix~\ref{app:world-hyperedges} reads each of H1, H2 and H3
on the mechanism: the content of its sources, the repair each admits in the doctrine's terms,
and the change that accounts for the three at once.

\subsection{A validation pattern for discovery worlds}
\label{app:val-pattern}

The checks compress into practices for other discovery worlds.

\begin{practices}
\practice{One source}{Declare the boundary, and compile every variant from one source. Each aid
then becomes a priced manipulation, and a structural comparison can confirm that the declared
surfaces are the whole difference.}
\practice{A reference route}{Exhibit a route that acquires the evidence the target is computed
from, run it inside the evaluation protocol, and give a model the route's complete record in
one prompt. The route places the evidence inside the budget, and the full-record probe
separates acquiring the readings from constructing an account of them.}
\practice{Channel probes}{Probe the channels that could bypass the inquiry, the prior, the names
and the answer format, and report each probe as an observed frequency with its model, its
instances and the criterion fixed for it in advance.}
\practice{Priced strategies}{Attack every visible surface with scripted strategies, and count a
pattern as a shortcut once it compiles into a strategy with a price. Pricing puts leaks and
designed evidence on one scale, so a defence that prices out a designed channel registers as a
defect.}
\practice{The answer space}{Measure the answer space on reserved instances by the answers it
realises and the share of its most common values, beside the capacity of the field ranges, and
keep evaluation instances fresh until their episodes run.}
\practice{A fixed scorer}{Fix the scorer before reading any transcript, require world grounding
for every criterion a grant states, and keep controls that must score zero. A restated account
and a finding can share their words, and the evidence in the event log with the grounding
verdict separates them.}
\end{practices}

%% file: figures/fig_route.tex
\begin{figure}[t]
\centering
\begin{tikzpicture}[x=1pt,y=1pt,font=\tiny]
\node[anchor=west,inner sep=0pt,text=black!70] at (0.00,0.00) {pass 1};
\node[anchor=west,inner sep=0pt,text=black!70] at (0.00,-39.00) {pass 2};
\node[anchor=west,inner sep=0pt,text=black!70] at (0.00,-78.00) {pass 3};
\draw[black!45,line width=0.4pt,dash pattern=on 1.2pt off 1pt] (117.00,-87.00) -- (117.00,-69.00);
\node[anchor=west,inner sep=0pt,text=black!70] at (121.50,-78.00) {pass 4};
\filldraw[fill=dTrellis!80,draw=dTrellis!60!black,line width=0.35pt,rounded corners=1.4pt] (21.00,-5.25) rectangle (31.50,5.25);
\node[inner sep=0pt] at (26.25,0.00) {\IfFileExists{figures/icons/goto.pdf}{\includegraphics[height=6.8pt]{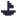}}{\tiny goto}};
\node[anchor=north,inner sep=0pt,text=black!65] at (26.25,-6.65) {0};
\filldraw[fill=dTrellis!80,draw=dTrellis!60!black,line width=0.35pt,rounded corners=1.4pt] (34.50,-5.25) rectangle (45.00,5.25);
\node[inner sep=0pt] at (39.75,0.00) {\IfFileExists{figures/icons/read.pdf}{\includegraphics[height=6.8pt]{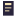}}{\tiny read}};
\node[anchor=north,inner sep=0pt,text=black!65] at (39.75,-6.65) {1};
\filldraw[fill=dTrellis!80,draw=dTrellis!60!black,line width=0.35pt,rounded corners=1.4pt] (48.00,-5.25) rectangle (58.50,5.25);
\node[inner sep=0pt] at (53.25,0.00) {\IfFileExists{figures/icons/retool.pdf}{\includegraphics[height=6.8pt]{figures/icons/retool.pdf}}{\tiny retool}};
\node[anchor=north,inner sep=0pt,text=black!65] at (53.25,-6.65) {2};
\filldraw[fill=dTrellis!80,draw=dTrellis!60!black,line width=0.35pt,rounded corners=1.4pt] (61.50,-5.25) rectangle (72.00,5.25);
\node[inner sep=0pt] at (66.75,0.00) {\IfFileExists{figures/icons/read.pdf}{\includegraphics[height=6.8pt]{figures/icons/read.pdf}}{\tiny read}};
\node[anchor=north,inner sep=0pt,text=black!65] at (66.75,-6.65) {3};
\filldraw[fill=dTrellis!80,draw=dTrellis!60!black,line width=0.35pt,rounded corners=1.4pt] (75.00,-5.25) rectangle (85.50,5.25);
\node[inner sep=0pt] at (80.25,0.00) {\IfFileExists{figures/icons/read.pdf}{\includegraphics[height=6.8pt]{figures/icons/read.pdf}}{\tiny read}};
\node[anchor=north,inner sep=0pt,text=black!65] at (80.25,-6.65) {4};
\filldraw[fill=dConsign!80,draw=dConsign!60!black,line width=0.35pt,rounded corners=1.4pt] (88.50,-5.25) rectangle (99.00,5.25);
\node[inner sep=0pt] at (93.75,0.00) {\IfFileExists{figures/icons/goto.pdf}{\includegraphics[height=6.8pt]{figures/icons/goto.pdf}}{\tiny goto}};
\node[anchor=north,inner sep=0pt,text=black!65] at (93.75,-6.65) {5};
\filldraw[fill=dConsign!80,draw=dConsign!60!black,line width=0.35pt,rounded corners=1.4pt] (102.00,-5.25) rectangle (112.50,5.25);
\node[inner sep=0pt] at (107.25,0.00) {\IfFileExists{figures/icons/read.pdf}{\includegraphics[height=6.8pt]{figures/icons/read.pdf}}{\tiny read}};
\node[anchor=north,inner sep=0pt,text=black!65] at (107.25,-6.65) {6};
\filldraw[fill=dConsign!80,draw=dConsign!60!black,line width=0.35pt,rounded corners=1.4pt] (115.50,-5.25) rectangle (126.00,5.25);
\node[inner sep=0pt] at (120.75,0.00) {\IfFileExists{figures/icons/release.pdf}{\includegraphics[height=6.8pt]{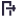}}{\tiny release}};
\node[anchor=north,inner sep=0pt,text=black!65] at (120.75,-6.65) {7};
\filldraw[fill=dConsign!80,draw=dConsign!60!black,line width=0.35pt,rounded corners=1.4pt] (129.00,-5.25) rectangle (139.50,5.25);
\node[inner sep=0pt] at (134.25,0.00) {\IfFileExists{figures/icons/read.pdf}{\includegraphics[height=6.8pt]{figures/icons/read.pdf}}{\tiny read}};
\node[anchor=north,inner sep=0pt,text=black!65] at (134.25,-6.65) {8};
\filldraw[fill=dConsign!80,draw=dConsign!60!black,line width=0.35pt,rounded corners=1.4pt] (142.50,-5.25) rectangle (153.00,5.25);
\node[inner sep=0pt] at (147.75,0.00) {\IfFileExists{figures/icons/read.pdf}{\includegraphics[height=6.8pt]{figures/icons/read.pdf}}{\tiny read}};
\node[anchor=north,inner sep=0pt,text=black!65] at (147.75,-6.65) {9};
\filldraw[fill=dLadder!80,draw=dLadder!60!black,line width=0.35pt,rounded corners=1.4pt] (156.00,-5.25) rectangle (166.50,5.25);
\node[inner sep=0pt] at (161.25,0.00) {\IfFileExists{figures/icons/goto.pdf}{\includegraphics[height=6.8pt]{figures/icons/goto.pdf}}{\tiny goto}};
\node[anchor=north,inner sep=0pt,text=black!65] at (161.25,-6.65) {10};
\filldraw[fill=dLadder!80,draw=dLadder!60!black,line width=0.35pt,rounded corners=1.4pt] (169.50,-5.25) rectangle (180.00,5.25);
\node[inner sep=0pt] at (174.75,0.00) {\IfFileExists{figures/icons/read.pdf}{\includegraphics[height=6.8pt]{figures/icons/read.pdf}}{\tiny read}};
\node[anchor=north,inner sep=0pt,text=black!65] at (174.75,-6.65) {11};
\filldraw[fill=dLadder!80,draw=dLadder!60!black,line width=0.35pt,rounded corners=1.4pt] (183.00,-5.25) rectangle (193.50,5.25);
\node[inner sep=0pt] at (188.25,0.00) {\IfFileExists{figures/icons/wait.pdf}{\includegraphics[height=6.8pt]{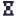}}{\tiny wait}};
\node[anchor=north,inner sep=0pt,text=black!65] at (188.25,-6.65) {12};
\filldraw[fill=dLadder!80,draw=dLadder!60!black,line width=0.35pt,rounded corners=1.4pt] (21.00,-44.25) rectangle (31.50,-33.75);
\node[inner sep=0pt] at (26.25,-39.00) {\IfFileExists{figures/icons/wait.pdf}{\includegraphics[height=6.8pt]{figures/icons/wait.pdf}}{\tiny wait}};
\node[anchor=north,inner sep=0pt,text=black!65] at (26.25,-45.65) {0};
\filldraw[fill=dLadder!80,draw=dLadder!60!black,line width=0.35pt,rounded corners=1.4pt] (34.50,-44.25) rectangle (45.00,-33.75);
\node[inner sep=0pt] at (39.75,-39.00) {\IfFileExists{figures/icons/stamp.pdf}{\includegraphics[height=6.8pt]{figures/icons/stamp.pdf}}{\tiny stamp}};
\node[anchor=north,inner sep=0pt,text=black!65] at (39.75,-45.65) {6};
\filldraw[fill=dLadder!80,draw=dLadder!60!black,line width=0.35pt,rounded corners=1.4pt] (48.00,-44.25) rectangle (58.50,-33.75);
\node[inner sep=0pt] at (53.25,-39.00) {\IfFileExists{figures/icons/read.pdf}{\includegraphics[height=6.8pt]{figures/icons/read.pdf}}{\tiny read}};
\node[anchor=north,inner sep=0pt,text=black!65] at (53.25,-45.65) {7};
\filldraw[fill=dLadder!80,draw=dLadder!60!black,line width=0.35pt,rounded corners=1.4pt] (61.50,-44.25) rectangle (72.00,-33.75);
\node[inner sep=0pt] at (66.75,-39.00) {\IfFileExists{figures/icons/wait.pdf}{\includegraphics[height=6.8pt]{figures/icons/wait.pdf}}{\tiny wait}};
\node[anchor=north,inner sep=0pt,text=black!65] at (66.75,-45.65) {8};
\filldraw[fill=dLadder!80,draw=dLadder!60!black,line width=0.35pt,rounded corners=1.4pt] (75.00,-44.25) rectangle (85.50,-33.75);
\node[inner sep=0pt] at (80.25,-39.00) {\IfFileExists{figures/icons/read.pdf}{\includegraphics[height=6.8pt]{figures/icons/read.pdf}}{\tiny read}};
\node[anchor=north,inner sep=0pt,text=black!65] at (80.25,-45.65) {9};
\filldraw[fill=dLadder!80,draw=dLadder!60!black,line width=0.35pt,rounded corners=1.4pt] (88.50,-44.25) rectangle (99.00,-33.75);
\node[inner sep=0pt] at (93.75,-39.00) {\IfFileExists{figures/icons/feed.pdf}{\includegraphics[height=6.8pt]{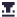}}{\tiny feed}};
\node[anchor=north,inner sep=0pt,text=black!65] at (93.75,-45.65) {10};
\filldraw[fill=dLadder!80,draw=dLadder!60!black,line width=0.35pt,rounded corners=1.4pt] (102.00,-44.25) rectangle (112.50,-33.75);
\node[inner sep=0pt] at (107.25,-39.00) {\IfFileExists{figures/icons/read.pdf}{\includegraphics[height=6.8pt]{figures/icons/read.pdf}}{\tiny read}};
\node[anchor=north,inner sep=0pt,text=black!65] at (107.25,-45.65) {11};
\filldraw[fill=dFen!80,draw=dFen!60!black,line width=0.35pt,rounded corners=1.4pt] (115.50,-44.25) rectangle (126.00,-33.75);
\node[inner sep=0pt] at (120.75,-39.00) {\IfFileExists{figures/icons/goto.pdf}{\includegraphics[height=6.8pt]{figures/icons/goto.pdf}}{\tiny goto}};
\node[anchor=north,inner sep=0pt,text=black!65] at (120.75,-45.65) {12};
\filldraw[fill=dFen!80,draw=dFen!60!black,line width=0.35pt,rounded corners=1.4pt] (129.00,-44.25) rectangle (139.50,-33.75);
\node[inner sep=0pt] at (134.25,-39.00) {\IfFileExists{figures/icons/read.pdf}{\includegraphics[height=6.8pt]{figures/icons/read.pdf}}{\tiny read}};
\node[anchor=north,inner sep=0pt,text=black!65] at (134.25,-45.65) {13};
\filldraw[fill=dFen!80,draw=dFen!60!black,line width=0.35pt,rounded corners=1.4pt] (142.50,-44.25) rectangle (153.00,-33.75);
\node[inner sep=0pt] at (147.75,-39.00) {\IfFileExists{figures/icons/retool.pdf}{\includegraphics[height=6.8pt]{figures/icons/retool.pdf}}{\tiny retool}};
\node[anchor=north,inner sep=0pt,text=black!65] at (147.75,-45.65) {14};
\filldraw[fill=dFen!80,draw=dFen!60!black,line width=0.35pt,rounded corners=1.4pt] (156.00,-44.25) rectangle (166.50,-33.75);
\node[inner sep=0pt] at (161.25,-39.00) {\IfFileExists{figures/icons/read.pdf}{\includegraphics[height=6.8pt]{figures/icons/read.pdf}}{\tiny read}};
\node[anchor=north,inner sep=0pt,text=black!65] at (161.25,-45.65) {15};
\filldraw[fill=dFen!80,draw=dFen!60!black,line width=0.35pt,rounded corners=1.4pt] (169.50,-44.25) rectangle (180.00,-33.75);
\node[inner sep=0pt] at (174.75,-39.00) {\IfFileExists{figures/icons/release.pdf}{\includegraphics[height=6.8pt]{figures/icons/release.pdf}}{\tiny release}};
\node[anchor=north,inner sep=0pt,text=black!65] at (174.75,-45.65) {16};
\filldraw[fill=dFen!80,draw=dFen!60!black,line width=0.35pt,rounded corners=1.4pt] (183.00,-44.25) rectangle (193.50,-33.75);
\node[inner sep=0pt] at (188.25,-39.00) {\IfFileExists{figures/icons/read.pdf}{\includegraphics[height=6.8pt]{figures/icons/read.pdf}}{\tiny read}};
\node[anchor=north,inner sep=0pt,text=black!65] at (188.25,-45.65) {17};
\filldraw[fill=dFen!80,draw=dFen!60!black,line width=0.35pt,rounded corners=1.4pt] (196.50,-44.25) rectangle (207.00,-33.75);
\node[inner sep=0pt] at (201.75,-39.00) {\IfFileExists{figures/icons/wait.pdf}{\includegraphics[height=6.8pt]{figures/icons/wait.pdf}}{\tiny wait}};
\node[anchor=north,inner sep=0pt,text=black!65] at (201.75,-45.65) {18};
\filldraw[fill=dFen!80,draw=dFen!60!black,line width=0.35pt,rounded corners=1.4pt] (21.00,-83.25) rectangle (31.50,-72.75);
\node[inner sep=0pt] at (26.25,-78.00) {\IfFileExists{figures/icons/wait.pdf}{\includegraphics[height=6.8pt]{figures/icons/wait.pdf}}{\tiny wait}};
\node[anchor=north,inner sep=0pt,text=black!65] at (26.25,-84.65) {0};
\filldraw[fill=dFen!80,draw=dFen!60!black,line width=0.35pt,rounded corners=1.4pt] (34.50,-83.25) rectangle (45.00,-72.75);
\node[inner sep=0pt] at (39.75,-78.00) {\IfFileExists{figures/icons/read.pdf}{\includegraphics[height=6.8pt]{figures/icons/read.pdf}}{\tiny read}};
\node[anchor=north,inner sep=0pt,text=black!65] at (39.75,-84.65) {13};
\filldraw[fill=dCadastre!80,draw=dCadastre!60!black,line width=0.35pt,rounded corners=1.4pt] (48.00,-83.25) rectangle (58.50,-72.75);
\node[inner sep=0pt] at (53.25,-78.00) {\IfFileExists{figures/icons/goto.pdf}{\includegraphics[height=6.8pt]{figures/icons/goto.pdf}}{\tiny goto}};
\node[anchor=north,inner sep=0pt,text=black!65] at (53.25,-84.65) {14};
\filldraw[fill=dCadastre!80,draw=dCadastre!60!black,line width=0.35pt,rounded corners=1.4pt] (61.50,-83.25) rectangle (72.00,-72.75);
\node[inner sep=0pt] at (66.75,-78.00) {\IfFileExists{figures/icons/read.pdf}{\includegraphics[height=6.8pt]{figures/icons/read.pdf}}{\tiny read}};
\node[anchor=north,inner sep=0pt,text=black!65] at (66.75,-84.65) {15};
\filldraw[fill=dCadastre!80,draw=dCadastre!60!black,line width=0.35pt,rounded corners=1.4pt] (75.00,-83.25) rectangle (85.50,-72.75);
\node[inner sep=0pt] at (80.25,-78.00) {\IfFileExists{figures/icons/stamp.pdf}{\includegraphics[height=6.8pt]{figures/icons/stamp.pdf}}{\tiny stamp}};
\node[anchor=north,inner sep=0pt,text=black!65] at (80.25,-84.65) {16};
\filldraw[fill=dCadastre!80,draw=dCadastre!60!black,line width=0.35pt,rounded corners=1.4pt] (88.50,-83.25) rectangle (99.00,-72.75);
\node[inner sep=0pt] at (93.75,-78.00) {\IfFileExists{figures/icons/read.pdf}{\includegraphics[height=6.8pt]{figures/icons/read.pdf}}{\tiny read}};
\node[anchor=north,inner sep=0pt,text=black!65] at (93.75,-84.65) {17};
\filldraw[fill=dCadastre!80,draw=dCadastre!60!black,line width=0.35pt,rounded corners=1.4pt] (102.00,-83.25) rectangle (112.50,-72.75);
\node[inner sep=0pt] at (107.25,-78.00) {\IfFileExists{figures/icons/wait.pdf}{\includegraphics[height=6.8pt]{figures/icons/wait.pdf}}{\tiny wait}};
\node[anchor=north,inner sep=0pt,text=black!65] at (107.25,-84.65) {18};
\filldraw[fill=dCadastre!80,draw=dCadastre!60!black,line width=0.35pt,rounded corners=1.4pt] (142.50,-83.25) rectangle (153.00,-72.75);
\node[inner sep=0pt] at (147.75,-78.00) {\IfFileExists{figures/icons/wait.pdf}{\includegraphics[height=6.8pt]{figures/icons/wait.pdf}}{\tiny wait}};
\node[anchor=north,inner sep=0pt,text=black!65] at (147.75,-84.65) {0};
\filldraw[fill=dCadastre!80,draw=dCadastre!60!black,line width=0.35pt,rounded corners=1.4pt] (156.00,-83.25) rectangle (166.50,-72.75);
\node[inner sep=0pt] at (161.25,-78.00) {\IfFileExists{figures/icons/read.pdf}{\includegraphics[height=6.8pt]{figures/icons/read.pdf}}{\tiny read}};
\node[anchor=north,inner sep=0pt,text=black!65] at (161.25,-84.65) {14};
\filldraw[fill=dCadastre!80,draw=dCadastre!60!black,line width=0.35pt,rounded corners=1.4pt] (169.50,-83.25) rectangle (180.00,-72.75);
\node[inner sep=0pt] at (174.75,-78.00) {\IfFileExists{figures/icons/goto.pdf}{\includegraphics[height=6.8pt]{figures/icons/goto.pdf}}{\tiny goto}};
\node[anchor=north,inner sep=0pt,text=black!65] at (174.75,-84.65) {15};
\filldraw[fill=dCadastre!80,draw=dCadastre!60!black,line width=0.35pt,rounded corners=1.4pt] (183.00,-83.25) rectangle (193.50,-72.75);
\node[inner sep=0pt] at (188.25,-78.00) {\IfFileExists{figures/icons/read.pdf}{\includegraphics[height=6.8pt]{figures/icons/read.pdf}}{\tiny read}};
\node[anchor=north,inner sep=0pt,text=black!65] at (188.25,-84.65) {16};
\filldraw[fill=dCadastre!80,draw=dCadastre!60!black,line width=0.35pt,rounded corners=1.4pt] (196.50,-83.25) rectangle (207.00,-72.75);
\filldraw[fill=wGold,draw=wInk,line width=0.45pt,line join=round] (201.75,-74.20) -- (200.72,-76.88) -- (197.85,-77.03) -- (200.09,-78.84) -- (199.34,-81.62) -- (201.75,-80.05) -- (204.16,-81.62) -- (203.41,-78.84) -- (205.65,-77.03) -- (202.78,-76.88) -- cycle;
\node[anchor=north,inner sep=0pt,text=black!65] at (201.75,-84.65) {17};
\draw[dTrellis!60!black,line width=0.5pt] (21.60,6.25) -- (21.60,7.85) -- (84.90,7.85) -- (84.90,6.25);
\node[anchor=south,inner sep=0pt] at (53.25,9.05) {TR1\,\textcolor{black!38}{TR2}\,\textcolor{black!38}{TR3}};
\node[anchor=north,inner sep=0pt,text=black!75,font=\tiny\itshape] at (60.00,-12.85) {job strip};
\draw[dConsign!60!black,line width=0.5pt] (89.10,6.25) -- (89.10,7.85) -- (152.40,7.85) -- (152.40,6.25);
\node[anchor=south,inner sep=0pt] at (120.75,9.05) {\textcolor{black!38}{CO1}\,\textcolor{black!38}{CO2}\,\textcolor{black!38}{CO3}};
\node[anchor=north,inner sep=0pt,text=black!75,font=\tiny\itshape] at (127.50,-12.85) {manifest slate};
\draw[dLadder!60!black,line width=0.5pt] (156.60,6.25) -- (156.60,7.85) -- (192.90,7.85) -- (192.90,6.25);
\draw[dLadder!60!black,line width=0.5pt] (21.60,-32.75) -- (21.60,-31.15) -- (111.90,-31.15) -- (111.90,-32.75);
\node[anchor=south,inner sep=0pt] at (66.75,-29.95) {LD1\,\textcolor{black!38}{LD2}\,\textcolor{black!38}{LD3}\,\textcolor{black!38}{XC2}};
\node[anchor=north,inner sep=0pt,text=black!75,font=\tiny\itshape] at (80.25,-51.85) {answer board};
\draw[dFen!60!black,line width=0.5pt] (116.10,-32.75) -- (116.10,-31.15) -- (206.40,-31.15) -- (206.40,-32.75);
\node[anchor=south,inner sep=0pt] at (161.25,-29.95) {FE1\,FE2\,\textcolor{black!38}{FE3}\,XC1};
\node[anchor=north,inner sep=0pt,text=black!75,font=\tiny\itshape] at (161.25,-51.85) {intake tallyboard};
\draw[dFen!60!black,line width=0.5pt] (21.60,-71.75) -- (21.60,-70.15) -- (44.40,-70.15) -- (44.40,-71.75);
\draw[dCadastre!60!black,line width=0.5pt] (48.60,-71.75) -- (48.60,-70.15) -- (111.90,-70.15) -- (111.90,-71.75);
\node[anchor=south,inner sep=0pt] at (80.25,-68.95) {\textcolor{black!38}{CA1}\,\textcolor{black!38}{CA2}\,\textcolor{black!38}{XC3}};
\node[anchor=north,inner sep=0pt,text=black!75,font=\tiny\itshape] at (80.25,-90.85) {audit counter};
\draw[dCadastre!60!black,line width=0.5pt] (143.10,-71.75) -- (143.10,-70.15) -- (165.90,-70.15) -- (165.90,-71.75);
\draw[dCadastre!60!black,line width=0.5pt] (170.10,-71.75) -- (170.10,-70.15) -- (192.90,-70.15) -- (192.90,-71.75);
\node[anchor=south,inner sep=0pt] at (181.50,-68.95) {CA3};
\node[anchor=north,inner sep=0pt,text=black!75,font=\tiny\itshape] at (188.25,-90.85) {pass ledger};
\node[anchor=north,inner sep=0pt,text=black!80] at (103.50,-100.25) {\pix{goto}goto\hspace{0.7em}\pix{read}read\hspace{0.7em}\pix{wait}wait\hspace{0.7em}\pix{feed}feed\hspace{0.7em}\pix{release}release\hspace{0.7em}\pix{stamp}stamp\hspace{0.7em}\pix{retool}retool\hspace{0.7em}\textcolor{wGold!80!black}{$\bigstar$}\,commit};
\node[anchor=north,inner sep=0pt,text=black!80] at (103.50,-109.25) {block ids in ink: the route satisfies the block's detector\hspace{1.2em}\textcolor{black!38}{in grey: it does not}};
\end{tikzpicture}
\caption{The reference route on one instance and the readings it takes: \resRouteActions{} paid actions from the first goto to the commit (star), across four passes. Each tile is one paid action, shown by its verb and tinted by the domain of the site where it occurs, with its tick below. A bracket joins the actions of one visit to a site, with the instrument read there below it and, above it, the blocks the design attributes to the site, in ink where the route itself satisfies the block's detector.}
\label{fig:route}
\end{figure}

%% file: sections/app_d_protocols.tex
\section{Experimental protocols}
\label{app:protocols}
\label{app:harness}

\subsection{The baseline loop}
\label{app:harness-loop}

On every turn the agent reads one prompt in five parts, in a fixed order
(Figure~\ref{fig:turn}): the fixed opening with the mission, the interface contract and, in the
Disclosed conditions, the Disclosed document; the event digest, an append-only record of every public
event; the \pix{notes}workspace, the agent's own notes in full; the full records its previous
recall query matched; and the current view, with the available actions, the budgets and, once
posted, the standing order. It answers with one action and, at no extra cost, an optional note
and an optional recall query. Every record the agent has observed stays within reach, in full
and in log order, and every judgement of relevance is left to the agent. The baseline workspace
is unstructured. The contracts are printed in Appendix~\ref{app:prompts-contract}.

\IfFileExists{figures/fig_turn.tex}{\input{figures/fig_turn}}{}

Each scaffold wraps this loop and leaves the rest of it unchanged (Table~\ref{tab:harness-arms};
prompts in Appendix~\ref{app:prompts-scaffolds}). It adds unpaid calls around the agent's turn,
a delimited region at the top of the workspace that the scaffold rewrites under its own title,
and, for \pix{arm-falsify}falsification and \pix{arm-evolve}evolution, a short note after the
current view. The structure a scaffold keeps in its region is the treatment, and the agent's
own notes below it stay as free as in the \pix{arm-baseline}baseline.

\begin{table*}[t]
\centering
\caption{Inquiry scaffolds and their interaction with the baseline loop. Every arm keeps the
200 paid actions of the baseline; scaffold calls are unpaid and additional, and the agent
chooses every action.}
\label{tab:harness-arms}
\footnotesize
\setlength{\tabcolsep}{5pt}
\renewcommand{\arraystretch}{1.15}
\begin{tabularx}{\textwidth}{@{}l
  >{\raggedright\arraybackslash}X
  >{\raggedright\arraybackslash}X
  >{\raggedright\arraybackslash}X@{}}
\toprule
Arm & Region kept at the top of the workspace & When it runs & What it hands to the agent\\
\midrule
\pix{arm-baseline}Baseline & none & none & the plain turn prompt\\
\arrayrulecolor{black!25}\midrule
\pix{arm-critique}Self-critique & a working summary: the best current account, the open
  puzzles, and earlier beliefs that later evidence contradicted
  & one review every ten paid actions, before the agent's turn
  & the rewritten summary\\
\midrule
\pix{arm-falsify}Falsification & a slate of at most five rival accounts, each with a next
  step and the result it predicts, and a ruled-out list
  & every round: a proposal when a place on the slate is empty, a screen before the turn,
  a refutation check after the action
  & the slate, and the screen of the next step of the least refuted account\\
\midrule
\pix{arm-evolve}Evolution & a population of four candidate mechanisms with their fitness
  scores and parents
  & every round: one call proposing two new candidates, one fitness call
  & the population, and a note naming the leading candidate\\
\arrayrulecolor{black}\bottomrule
\end{tabularx}
\end{table*}

\subsection{Self-critique}
\label{app:harness-critique}

The review call of \pix{arm-critique}self-critique reads the agent's \pix{notes}notes and a
recent tail of the digest and rewrites the region headed \textsc{working summary} from scratch:
the best current account of the mechanism in at most four sentences, at most five open puzzles,
and at most five earlier beliefs in the notes that later evidence contradicted, each with that
evidence. The agent may use, edit or ignore it. The schedule counts the paid actions the agent
takes, so an agent that yields its remaining actions early receives fewer reviews. Reflexion
\citep{shinn2023reflexion} stores the verbal critique of a failed trial for the next trial. An
episode here is a single trial, so the critique is applied within the run, the operation of
Self-Refine \citep{madaan2023selfrefine}.

\subsection{Falsification}
\label{app:harness-falsify}

Each account on the slate of \pix{arm-falsify}falsification carries one concrete next step in the
vocabulary of the digest and the result that step would produce if the account holds, stated
so that a different result refutes it. A round runs up to three unpaid calls around the agent's
paid action. Before the turn, a proposal call fills the empty places whenever the slate is
empty or an account has retired, and a screening call judges the step of the least refuted
account against the digest: whether its result is already shown, which accounts it would tell
apart, and, when it separates none, a better step. The screen is appended to the agent's turn
and hands the choice of action back to the agent. After the action, a refutation call marks each
account whose stated prediction the result contradicts. A contradicted prediction is the one
ground for a mark, and an account retires at its second refutation to a ruled-out list with its
reasons. The arm keeps the stance of POPPER \citep{huang2025popper}, which validates a hypothesis
by attempted falsification and screens each proposed test before it runs, and replaces its
sequential statistical test, built on a sampling model of the outcome, with this count of
contradictions.

\subsection{Evolution}
\label{app:harness-evolve}

The first mutation call of \pix{arm-evolve}evolution seeds four candidate mechanisms in free
prose. After that, one batched call before each turn proposes two new candidates, plus one for
each place left empty. A proposal may cite any current candidates as parents, so that lineage
forms a graph, and must change something the log could decide. A fitness call then scores every
candidate by retrodiction: the share of a digest excerpt it would have called correctly in
advance, from 0 (contradicted outright) to 1 (every line accounted for), whatever the length,
detail or plausibility of the account. The fitness excerpt is a longer tail of the digest than
the other calls read. Selection keeps the four best and skips a candidate whose character
$n$-grams nearly duplicate one already kept. The population forms the region, and a note
appended to the turn names the leader and observes that a step telling the leading candidates
apart is worth more than one agreeing with the leader. The agent's ordinary turn is the
selector.

The population and the mutation loop follow AlphaEvolve \citep{novikov2025alphaevolve} and
ShinkaEvolve \citep{lange2025shinkaevolve}, and treating the paid actions as a fixed evaluation
budget follows SimpleTES \citep{ye2026structured}. Those systems score candidates with an
external evaluator. Here the one external verdict is the final commitment, so fitness is the
model's own retrodiction against readings already in the digest, and the evaluator stays fixed
for the episode, where Hyra \citep{hyra2026} evolves it as well.

\subsection{Parity and separation}
\label{app:harness-parity}

\paragraph{Paid-action parity.} Every paid action comes from the agent's reply. Given the same
replies, a scaffolded episode and a baseline episode produce identical world events.

\paragraph{Separate scaffold-call allowance.} Scaffold calls are unpaid and drawn from an
allowance of their own, outside the 200 paid actions and the agent's turn limits
(Appendix~\ref{app:budgets}), and the turns left in the current
view count the agent's own calls, so every arm reads the same contract.
Table~\ref{tab:arm-calls} counts the scaffold calls per episode.

\IfFileExists{tables/tab_arm_calls.tex}{\input{tables/tab_arm_calls}}{}

\paragraph{World-agnostic prompts.} The scaffold prompts are the same text for any world. A call
reads at most the agent's \pix{notes}notes, a tail of the digest and the scaffold's own list of
accounts. The available actions, the current view and the definition of the world reach the
agent alone. The region takes a bounded share of the workspace capacity, and the rest belongs
to the agent's notes.

\paragraph{The judge reads the agent's writing.} Before the judge reads the notes, the scaffold's
delimited region, and any copy of it that keeps its delimiters, is removed from every workspace
snapshot. A delimiter line the agent quoted on its own is deleted, and the region's title, where
the agent's prose names it, becomes a plain word. The agent's own prose stays as written,
including whatever it adopts from the region, and the judge reads every arm blind
(Appendix~\ref{app:judge}).

\subsection{Budgets}
\label{app:budgets}
\label{app:infra}

An episode has 200 paid actions and at most 300 agent turns, of which at most 100 may be
auxiliary: repairs of replies that fail the format. Every action on the world is paid, whether
it positions the agent (\pix{goto}goto), observes (\pix{read}read, \pix{wait}wait), intervenes
on one object (\pix{feed}feed, \pix{release}release, \pix{stamp}stamp, \pix{retool}retool,
\pix{cull}cull) or commits. Notes and recall cost nothing beyond the turn that carries them.
With ten paid actions remaining the standing order is posted, and the commit costs one paid
action and is made once. An agent may yield its remaining actions in a single move. They are
charged in full and the world advances through each of them, so the window opens at the same
point as for an agent that spends them one by one. Every model runs with the same settings in
all four conditions and samples at temperature 0.3 where the model accepts one.

\IfFileExists{tables/tab_tokens.tex}{\input{tables/tab_tokens}}{}

Table~\ref{tab:tokens} gives the tokens each model spends on the fixed action budget.

\subsection{Excluded episodes}
\label{app:censoring}

The serving of Claude Opus 5 applies a content filter that withheld replies, and the
\resCensoredFilter{} episodes it ended, all in Open${\times}$Aligned, are excluded. Malformed
replies ended \resCensoredCollapse{} further episodes, four of GPT-5.6 Luna and nine of
DeepSeek V4 Pro, which are excluded as well. Table~\ref{tab:main} marks the three models.

\IfFileExists{figures/fig_extended.tex}{\input{figures/fig_extended}}{}

\subsection{Vendor-harness systems}
\label{app:agentic}

\begin{table}[t]
\centering
\caption{Vendor-harness evaluation settings.}
\label{tab:agentic-config}
\footnotesize
\setlength{\tabcolsep}{4pt}
\renewcommand{\arraystretch}{1.12}
\begin{tabular}{@{}lll@{}}
\toprule
 & \logo{anthropic}Claude Code & \logo{openai}Codex\\
\midrule
Model & Claude Fable~5.1 & GPT-6 Astra\\
Reasoning effort & \texttt{xhigh} & \texttt{xhigh}\\
Player & fresh subagent & fresh subagent\\
Instances & \resAgenticSeeds{} per condition & \resAgenticSeeds{} per condition\\
Episodes played & \resFableCounted{} & \resAstraCounted{}\\
Passed the audit & \resFableCounted{} & \resAstraCounted{}\\
\bottomrule
\end{tabular}
\end{table}

Claude Fable~5.1 and GPT-6 Astra play inside the harnesses of their vendors, Claude Code and
Codex (Table~\ref{tab:agentic-config}). Each episode runs in a working directory that holds one
command-line client and nothing that names the instance or the condition. The client passes the
agent's replies to the episode loop of the main comparison and prints the world's responses, so
the world, the budgets, the reply grammar, the workspace and the metrics are those of
Section~\ref{sec:setup}, and the harness adds its own management of context, planning, files
and code execution inside the directory. A fresh subagent plays each episode with the brief of
Appendix~\ref{app:prompts-agentic} and nothing from the session that created it. The brief
confines it to its directory and excludes the network, and every transcript was audited for
both and passed. A vendor-harness system changes the model and the harness together, so the two
systems are reported beside the main comparison. Their timing is analysed in
Appendix~\ref{app:heat} and drawn in Figure~\ref{fig:curves}.

\subsection{Larger budgets}
\label{app:extended}

One instance is played again in all four conditions with larger budgets: the instance of the
two vendor-harness episodes read in Appendices~\ref{app:qual-fable} and~\ref{app:qual-astra}, on
which the vendor-harness systems come closest to the prediction. Claude Fable~5.1 in Claude
Code and GPT-6 Astra in Codex have \resExtBudgetVendor{} paid actions, \resExtTimesVendor{}
times the contract budget, and GPT-5.6 Sol in the baseline loop has \resExtBudgetLoop{},
\resExtTimesLoop{} times. The turn limits scale with the budget, the standing order is posted
with ten paid actions remaining, and the judge of the main comparison scores every episode. Each
episode is read beside its twin, the episode of the same system in the same condition on this
instance at 200 paid actions.

\IfFileExists{tables/tab_extended.tex}{\input{tables/tab_extended}}{}

Claude Fable~5.1 commits an exact prediction in Disclosed${\times}$Aligned
(Table~\ref{tab:extended}). That episode holds the evidence of \resExtExactOrderHeld{} of the
five blocks on the designed route to the order's readings, and its notes state \resExtExactStated{} of the three
representation changes. Every other extended episode commits an inexact prediction, the Open
ones included, and \resExtReadings{} of the \resExtReadingsAll{} readings committed at the larger
budgets are exact, as many as in their twins. The complete evidence of a hyperedge is held in
\resExtReachFull{} of the \resExtEpisodes{} extended episodes and in \resExtTwinReachFull{} of
their twins, and the most any extended episode holds of the five blocks on the designed route is
\resExtOrderHeldMax{}.

The three systems use the larger budget differently (Figure~\ref{fig:extended}). Fable holds the
evidence of \resExtFableEvidence{} blocks per episode against \resExtFableTwinEvidence{} in its
twins, \resExtFableGainLow{} to \resExtFableGainHigh{} more in every condition, and grounds
\resExtFableBlocks{} against \resExtFableTwinBlocks{}. Astra holds \resExtAstraEvidence{}
against \resExtAstraTwinEvidence{} and grounds \resExtAstraBlocks{} against
\resExtAstraTwinBlocks{}. Both keep producing evidence past the contract budget, Fable until
paid action~\resExtFableLastEvidence{} and Astra until paid action~\resExtAstraLastEvidence{},
and both request about twice the interventions of their twins, \resExtFableInterventions{} and
\resExtAstraInterventions{} per episode against \resExtFableTwinInterventions{} and
\resExtAstraTwinInterventions{}. Sol holds \resExtSolEvidence{} against
\resExtSolTwinEvidence{}, and the last evidence it produces before the standing order comes by
paid action~\resExtSolLastEvidence{}. From there it waits, and waits make up
\resExtSolWaitShare\% of its paid actions.

\clearpage

%% file: figures/fig_turn.tex
\begin{figure*}[t]
\centering
\includegraphics[width=\textwidth]{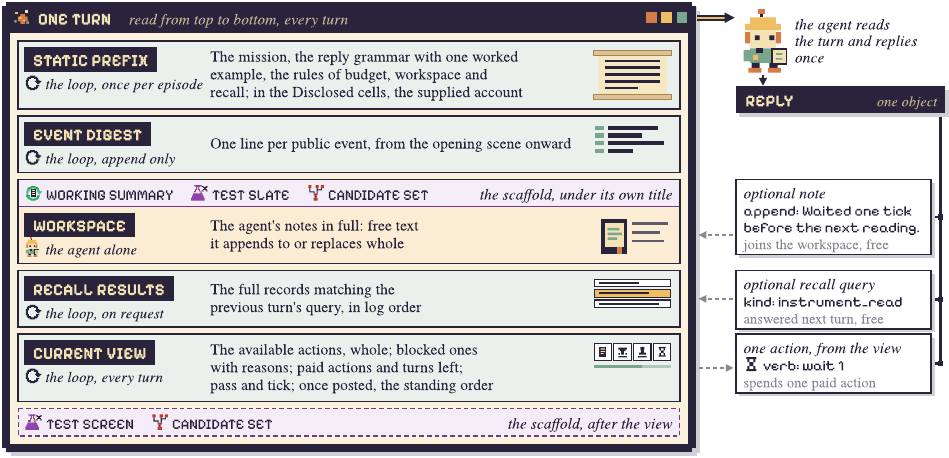}
\caption{The baseline loop and what the scaffolds add. Each turn the agent reads five parts from
top to bottom, the static prefix, the event digest, its own workspace, the records its last
query recalled and the current view, and returns one action, which spends a paid action, with
an optional note and recall query, both free. Tinted strips mark the scaffold additions: a
region at the top of the workspace and, for falsification and evolution, a note after the
view.}
\label{fig:turn}
\end{figure*}

%% file: tables/tab_arm_calls.tex
\begin{table}[t]
\centering
\caption{Auxiliary scaffold calls per episode by arm and condition. $n$ is the number of included episodes.}
\label{tab:arm-calls}
\footnotesize
\setlength{\tabcolsep}{4.5pt}
\renewcommand{\arraystretch}{1.12}
\begin{tabular}{@{}lrrrr@{}}
\toprule
 & \multicolumn{4}{c}{\pix{open}Open}\\
\cmidrule(l){2-5}
 & \multicolumn{2}{c}{\tcbox[on line,colback=cOpenDist!14,colframe=cOpenDist!65,boxrule=0.4pt,arc=1.8pt,boxsep=0pt,left=3pt,right=3pt,top=0.8pt,bottom=0.8pt]{\strut \pix{distributed}Distributed}} & \multicolumn{2}{c}{\tcbox[on line,colback=cOpenAlig!14,colframe=cOpenAlig!65,boxrule=0.4pt,arc=1.8pt,boxsep=0pt,left=3pt,right=3pt,top=0.8pt,bottom=0.8pt]{\strut \pix{aligned}Aligned}}\\[1pt]
Scaffold & $n$ & Calls & $n$ & Calls\\
\midrule
\textcolor{hBase}{\rule{1.1ex}{1.1ex}}\hspace{0.4em}\pix{arm-baseline}\textbf{Baseline} & 20 & 0.0 & 19 & 0.0\\
\textcolor{hCrit}{\rule{1.1ex}{1.1ex}}\hspace{0.4em}\pix{arm-critique}Self-critique & 20 & 11.7 & 20 & 9.4\\
\textcolor{hFals}{\rule{1.1ex}{1.1ex}}\hspace{0.4em}\pix{arm-falsify}Falsification & 19 & 209.5 & 20 & 163.1\\
\textcolor{hEvol}{\rule{1.1ex}{1.1ex}}\hspace{0.4em}\pix{arm-evolve}Evolution & 19 & 214.6 & 20 & 214.4\\
\bottomrule
\end{tabular}
\end{table}

%% file: tables/tab_tokens.tex
\begin{table}[t]
\centering
\caption{Episode counts and mean agent turns, prompt tokens and completion tokens by model. K and M denote thousands and millions.}
\label{tab:tokens}
\footnotesize
\setlength{\tabcolsep}{3.8pt}
\renewcommand{\arraystretch}{1.1}
\begin{tabular}{@{}lrrrr@{}}
\toprule
Model & $n$ & Turns & \multicolumn{2}{c}{Tokens per episode}\\
\cmidrule(l){4-5}
 & & & Prompt & Completion\\
\midrule
\logo{openai}GPT-5.6 Sol & 48 & 140.0 & 1.61M & 25K\\
\logo{openai}GPT-5.6 Terra & 48 & 104.1 & 1.12M & 20K\\
\logo{openai}GPT-5.6 Luna & 44 & 129.4 & 1.29M & 22K\\
\logo{anthropic}Claude Opus 5 & 45 & 252.3 & 5.63M & 314K\\
\logo{anthropic}Claude Sonnet 5 & 48 & 179.2 & 3.70M & 356K\\
\logo{google}Gemini 3.1 Pro & 48 & 111.3 & 1.37M & 240K\\
\logo{google}Gemini 3.6 Flash & 48 & 64.8 & 653K & 63K\\
\logo{deepseek}DeepSeek V4 Pro & 39 & 276.1 & 3.59M & 839K\\
\logo{deepseek}DeepSeek V4 Flash & 48 & 209.6 & 2.63M & 732K\\
\logo{zai}GLM-5.2 & 48 & 235.8 & 2.97M & 750K\\
\logo{moonshot}Kimi K3 & 48 & 197.5 & 2.37M & 303K\\
\midrule
All models & 512 & 170.8 & 2.42M & 327K\\
\bottomrule
\end{tabular}
\end{table}

%% file: figures/fig_extended.tex
\begin{figure*}[t]
\centering
\begin{tikzpicture}[x=0.2933pt,y=6.0000pt,font=\scriptsize,baseline=0pt]
\node[anchor=east,inner sep=1.5pt] at (0,0) {0};
\draw[black!10,line width=0.3pt] (0,2) -- (600,2);
\node[anchor=east,inner sep=1.5pt] at (0,2) {2};
\draw[black!10,line width=0.3pt] (0,4) -- (600,4);
\node[anchor=east,inner sep=1.5pt] at (0,4) {4};
\draw[black!10,line width=0.3pt] (0,6) -- (600,6);
\node[anchor=east,inner sep=1.5pt] at (0,6) {6};
\draw[black!10,line width=0.3pt] (0,8) -- (600,8);
\node[anchor=east,inner sep=1.5pt] at (0,8) {8};
\draw[black!10,line width=0.3pt] (0,10) -- (600,10);
\node[anchor=east,inner sep=1.5pt] at (0,10) {10};
\node[anchor=east,inner sep=1.5pt] at (0,12) {12};
\draw[black!60,line width=0.4pt] (0,0) -- ++(0,-2pt) node[below,inner sep=1.5pt] {0};
\draw[black!60,line width=0.4pt] (100,0) -- ++(0,-2pt) node[below,inner sep=1.5pt] {100};
\draw[black!60,line width=0.4pt] (200,0) -- ++(0,-2pt) node[below,inner sep=1.5pt] {200};
\draw[black!60,line width=0.4pt] (300,0) -- ++(0,-2pt) node[below,inner sep=1.5pt] {300};
\draw[black!60,line width=0.4pt] (400,0) -- ++(0,-2pt) node[below,inner sep=1.5pt] {400};
\draw[black!60,line width=0.4pt] (500,0) -- ++(0,-2pt) node[below,inner sep=1.5pt] {500};
\draw[black!60,line width=0.4pt] (600,0) -- ++(0,-2pt) node[below,inner sep=1.5pt] {600};
\draw[black!45,line width=0.4pt,dash pattern=on 1.6pt off 1.2pt] (0,12) -- (600,12) -- (600,0);
\draw[black!60,line width=0.4pt] (0,12) -- (0,0) -- (600,0);
\node[anchor=north] at (300,0) [yshift=-9pt] {paid actions};
\node[rotate=90,anchor=south] at (0,6) [xshift=0pt,yshift=7pt] {blocks};
\draw[wBrick!85!black!40,line width=0.55pt,densely dotted] (0,0.000) -- (2,0.000) -- (3,0.750) -- (5,0.750) -- (6,1.250) -- (11,1.250) -- (12,2.000) -- (14,2.000) -- (15,2.250) -- (21,2.250) -- (22,2.500) -- (25,2.500) -- (26,2.750) -- (27,3.000) -- (28,3.000) -- (29,3.250) -- (30,3.500) -- (37,3.500) -- (38,3.750) -- (47,3.750) -- (48,4.000) -- (49,4.250) -- (55,4.250) -- (56,4.500) -- (57,4.750) -- (68,4.750) -- (69,5.000) -- (73,5.000) -- (74,5.250) -- (81,5.250) -- (82,5.500) -- (83,5.750) -- (90,5.750) -- (91,6.250) -- (95,6.250) -- (96,6.500) -- (99,6.500) -- (100,6.750) -- (107,6.750) -- (108,7.000) -- (109,7.000) -- (110,7.250) -- (125,7.250) -- (126,7.500) -- (127,7.750) -- (143,7.750) -- (144,8.000) -- (145,8.000) -- (146,8.250) -- (152,8.250) -- (153,8.500) -- (156,8.500) -- (157,8.750) -- (200,8.750);
\draw[wTeal!70!black!40,line width=0.55pt,densely dotted] (0,0.000) -- (2,0.000) -- (3,1.000) -- (5,1.000) -- (6,1.500) -- (8,1.500) -- (9,1.750) -- (10,1.750) -- (11,2.000) -- (12,2.000) -- (13,2.500) -- (24,2.500) -- (25,2.750) -- (26,3.500) -- (27,3.500) -- (28,3.750) -- (29,4.000) -- (33,4.000) -- (34,4.750) -- (35,4.750) -- (36,5.250) -- (37,5.500) -- (45,5.500) -- (46,5.750) -- (48,5.750) -- (49,6.500) -- (71,6.500) -- (72,6.750) -- (90,6.750) -- (91,7.000) -- (94,7.000) -- (95,7.250) -- (98,7.250) -- (99,7.500) -- (106,7.500) -- (107,7.750) -- (124,7.750) -- (125,8.000) -- (127,8.000) -- (128,8.250) -- (129,8.250) -- (130,8.500) -- (154,8.500) -- (155,8.750) -- (182,8.750) -- (183,9.000) -- (189,9.000) -- (190,9.250) -- (194,9.250) -- (195,9.500) -- (200,9.500);
\draw[wBrick!85!black,line width=0.55pt,densely dotted] (0,0.000) -- (2,0.000) -- (3,1.000) -- (4,1.000) -- (5,1.250) -- (6,1.750) -- (10,1.750) -- (11,2.000) -- (14,2.000) -- (15,2.250) -- (17,2.250) -- (18,2.500) -- (19,2.500) -- (20,2.750) -- (25,2.750) -- (26,3.000) -- (28,3.000) -- (29,3.250) -- (30,3.500) -- (32,3.500) -- (33,3.750) -- (34,4.000) -- (40,4.000) -- (41,4.250) -- (44,4.250) -- (45,4.500) -- (52,4.500) -- (53,5.000) -- (56,5.000) -- (57,5.250) -- (60,5.250) -- (61,5.500) -- (62,6.000) -- (63,6.000) -- (64,6.250) -- (68,6.250) -- (69,6.500) -- (70,6.500) -- (71,6.750) -- (72,7.250) -- (76,7.250) -- (77,7.750) -- (90,7.750) -- (91,8.000) -- (110,8.000) -- (111,8.250) -- (116,8.250) -- (117,8.500) -- (130,8.500) -- (131,8.750) -- (142,8.750) -- (143,9.000) -- (174,9.000) -- (175,9.250) -- (176,9.500) -- (193,9.500) -- (194,9.750) -- (195,9.750) -- (196,10.000) -- (198,10.000) -- (199,10.250) -- (206,10.250) -- (207,10.500) -- (267,10.500) -- (268,10.750) -- (300,10.750) -- (301,11.000) -- (315,11.000) -- (316,11.250) -- (600,11.250);
\draw[wTeal!70!black,line width=0.55pt,densely dotted] (0,0.000) -- (2,0.000) -- (3,1.000) -- (5,1.000) -- (6,1.250) -- (9,1.250) -- (10,1.500) -- (11,1.500) -- (12,1.750) -- (13,2.000) -- (14,2.250) -- (24,2.250) -- (25,2.500) -- (26,3.000) -- (27,3.250) -- (28,3.500) -- (29,3.750) -- (36,3.750) -- (37,4.000) -- (40,4.000) -- (41,4.250) -- (42,4.750) -- (44,4.750) -- (45,5.000) -- (53,5.000) -- (54,5.250) -- (60,5.250) -- (61,5.500) -- (66,5.500) -- (67,5.750) -- (82,5.750) -- (83,6.000) -- (88,6.000) -- (89,6.250) -- (96,6.250) -- (97,6.500) -- (100,6.500) -- (101,6.750) -- (113,6.750) -- (114,7.000) -- (171,7.000) -- (172,7.250) -- (187,7.250) -- (188,7.500) -- (218,7.500) -- (219,7.750) -- (229,7.750) -- (230,8.000) -- (248,8.000) -- (249,8.250) -- (336,8.250) -- (337,8.500) -- (394,8.500) -- (395,8.750) -- (459,8.750) -- (460,9.000) -- (600,9.000);
\draw[wBrick!85!black!40,line width=0.95pt,line join=round] (0,0.000) -- (3,0.000) -- (4,0.750) -- (23,0.750) -- (24,1.000) -- (48,1.000) -- (49,1.250) -- (67,1.250) -- (68,1.500) -- (73,1.500) -- (74,1.750) -- (82,1.750) -- (83,2.000) -- (84,2.000) -- (85,2.250) -- (90,2.250) -- (91,2.750) -- (95,2.750) -- (96,3.000) -- (103,3.000) -- (104,3.250) -- (107,3.250) -- (108,3.500) -- (109,3.750) -- (110,4.000) -- (152,4.000) -- (153,4.250) -- (156,4.250) -- (157,4.500) -- (184,4.500) -- (185,5.250) -- (190,5.250) -- (191,5.500) -- (192,6.000) -- (193,6.000) -- (194,6.250) -- (195,6.250) -- (196,6.500) -- (197,6.750) -- (199,6.750) -- (200,7.500);
\draw[wTeal!70!black!40,line width=0.95pt,line join=round] (0,0.000) -- (3,0.000) -- (4,0.750) -- (20,0.750) -- (21,1.000) -- (24,1.000) -- (25,1.250) -- (45,1.250) -- (46,1.500) -- (49,1.500) -- (50,2.000) -- (67,2.000) -- (68,2.250) -- (90,2.250) -- (91,2.500) -- (92,2.750) -- (127,2.750) -- (128,3.000) -- (129,3.000) -- (130,3.250) -- (134,3.250) -- (135,3.500) -- (136,3.750) -- (149,3.750) -- (150,4.000) -- (174,4.000) -- (175,4.250) -- (186,4.250) -- (187,4.500) -- (189,4.500) -- (190,4.750) -- (191,5.750) -- (194,5.750) -- (195,6.000) -- (196,6.500) -- (197,6.750) -- (198,7.250) -- (199,7.250) -- (200,7.750);
\draw[wBrick!85!black,line width=0.95pt,line join=round] (0,0.000) -- (3,0.000) -- (4,1.000) -- (15,1.000) -- (16,1.250) -- (28,1.250) -- (29,1.500) -- (33,1.500) -- (34,1.750) -- (52,1.750) -- (53,2.000) -- (54,2.250) -- (61,2.250) -- (62,2.500) -- (63,2.500) -- (64,2.750) -- (71,2.750) -- (72,3.000) -- (76,3.000) -- (77,3.750) -- (96,3.750) -- (97,4.000) -- (101,4.000) -- (102,4.250) -- (110,4.250) -- (111,4.500) -- (112,4.750) -- (113,4.750) -- (114,5.000) -- (116,5.000) -- (117,5.250) -- (130,5.250) -- (131,5.500) -- (135,5.500) -- (136,5.750) -- (164,5.750) -- (165,6.000) -- (202,6.000) -- (203,6.250) -- (206,6.250) -- (207,6.500) -- (224,6.500) -- (225,6.750) -- (271,6.750) -- (272,7.000) -- (300,7.000) -- (301,7.250) -- (315,7.250) -- (316,7.500) -- (439,7.500) -- (440,7.750) -- (590,7.750) -- (591,8.000) -- (596,8.000) -- (597,8.250) -- (598,8.250) -- (599,8.500) -- (600,8.500);
\draw[wTeal!70!black,line width=0.95pt,line join=round] (0,0.000) -- (3,0.000) -- (4,0.500) -- (25,0.500) -- (26,0.750) -- (27,1.000) -- (82,1.000) -- (83,1.250) -- (88,1.250) -- (89,1.500) -- (160,1.500) -- (161,1.750) -- (178,1.750) -- (179,2.000) -- (250,2.000) -- (251,2.250) -- (304,2.250) -- (305,2.500) -- (459,2.500) -- (460,2.750) -- (589,2.750) -- (590,4.500) -- (591,6.500) -- (598,6.500) -- (599,6.750) -- (600,7.000);
\filldraw[fill=wBrick!85!black!40,draw=white,line width=0.3pt] (20,0.750) ++(0,1.9pt) -- ++(1.9pt,-1.9pt) -- ++(-1.9pt,-1.9pt) -- ++(-1.9pt,1.9pt) -- cycle;
\filldraw[fill=wBrick!85!black!40,draw=white,line width=0.3pt] (80,1.750) ++(0,1.9pt) -- ++(1.9pt,-1.9pt) -- ++(-1.9pt,-1.9pt) -- ++(-1.9pt,1.9pt) -- cycle;
\filldraw[fill=wBrick!85!black!40,draw=white,line width=0.3pt] (140,4.000) ++(0,1.9pt) -- ++(1.9pt,-1.9pt) -- ++(-1.9pt,-1.9pt) -- ++(-1.9pt,1.9pt) -- cycle;
\filldraw[fill=wBrick!85!black,draw=white,line width=0.3pt] (60,2.250) ++(0,1.9pt) -- ++(1.9pt,-1.9pt) -- ++(-1.9pt,-1.9pt) -- ++(-1.9pt,1.9pt) -- cycle;
\filldraw[fill=wBrick!85!black,draw=white,line width=0.3pt] (180,6.000) ++(0,1.9pt) -- ++(1.9pt,-1.9pt) -- ++(-1.9pt,-1.9pt) -- ++(-1.9pt,1.9pt) -- cycle;
\filldraw[fill=wBrick!85!black,draw=white,line width=0.3pt] (300,7.000) ++(0,1.9pt) -- ++(1.9pt,-1.9pt) -- ++(-1.9pt,-1.9pt) -- ++(-1.9pt,1.9pt) -- cycle;
\filldraw[fill=wBrick!85!black,draw=white,line width=0.3pt] (420,7.500) ++(0,1.9pt) -- ++(1.9pt,-1.9pt) -- ++(-1.9pt,-1.9pt) -- ++(-1.9pt,1.9pt) -- cycle;
\filldraw[fill=wBrick!85!black,draw=white,line width=0.3pt] (540,7.750) ++(0,1.9pt) -- ++(1.9pt,-1.9pt) -- ++(-1.9pt,-1.9pt) -- ++(-1.9pt,1.9pt) -- cycle;
\filldraw[fill=wTeal!70!black!40,draw=white,line width=0.3pt] (40,1.250) ++(0,2.2pt) -- ++(1.95pt,-3.35pt) -- ++(-3.9pt,0) -- cycle;
\filldraw[fill=wTeal!70!black!40,draw=white,line width=0.3pt] (100,2.750) ++(0,2.2pt) -- ++(1.95pt,-3.35pt) -- ++(-3.9pt,0) -- cycle;
\filldraw[fill=wTeal!70!black!40,draw=white,line width=0.3pt] (160,4.000) ++(0,2.2pt) -- ++(1.95pt,-3.35pt) -- ++(-3.9pt,0) -- cycle;
\filldraw[fill=wTeal!70!black,draw=white,line width=0.3pt] (90,1.500) ++(0,2.2pt) -- ++(1.95pt,-3.35pt) -- ++(-3.9pt,0) -- cycle;
\filldraw[fill=wTeal!70!black,draw=white,line width=0.3pt] (210,2.000) ++(0,2.2pt) -- ++(1.95pt,-3.35pt) -- ++(-3.9pt,0) -- cycle;
\filldraw[fill=wTeal!70!black,draw=white,line width=0.3pt] (330,2.500) ++(0,2.2pt) -- ++(1.95pt,-3.35pt) -- ++(-3.9pt,0) -- cycle;
\filldraw[fill=wTeal!70!black,draw=white,line width=0.3pt] (450,2.500) ++(0,2.2pt) -- ++(1.95pt,-3.35pt) -- ++(-3.9pt,0) -- cycle;
\filldraw[fill=wTeal!70!black,draw=white,line width=0.3pt] (570,2.750) ++(0,2.2pt) -- ++(1.95pt,-3.35pt) -- ++(-3.9pt,0) -- cycle;
\draw[wInk!85,line width=0.8pt,dash pattern=on 2.4pt off 1.4pt] (590,0) -- (590,12);
\draw[wInk!85,line width=0.8pt] (590,12) -- ++(0,7.2pt);
\node[fill=wInk!85,text=white,font=\tiny,inner xsep=1.8pt,inner ysep=1pt,anchor=north east] at (590,12) [yshift=7.2pt] {standing order};
\draw[wBrick!85!black,line width=0.95pt] (0,0) ++(0,-25pt) -- ++(10pt,0) node[right,inner sep=1.5pt,text=black] {\logoc{anthropic}Claude Fable 5.1-xhigh, 600 actions};
\filldraw[fill=wBrick!85!black,draw=white,line width=0.3pt] (0,0) ++(0,-25pt) ++(5pt,0) ++(0,1.9pt) -- ++(1.9pt,-1.9pt) -- ++(-1.9pt,-1.9pt) -- ++(-1.9pt,1.9pt) -- cycle;
\draw[wBrick!85!black!40,line width=0.95pt] (444,0) ++(0,-25pt) -- ++(10pt,0) node[right,inner sep=1.5pt,text=black] {twin, 200 actions};
\filldraw[fill=wBrick!85!black!40,draw=white,line width=0.3pt] (444,0) ++(0,-25pt) ++(5pt,0) ++(0,1.9pt) -- ++(1.9pt,-1.9pt) -- ++(-1.9pt,-1.9pt) -- ++(-1.9pt,1.9pt) -- cycle;
\draw[wTeal!70!black,line width=0.95pt] (0,0) ++(0,-34pt) -- ++(10pt,0) node[right,inner sep=1.5pt,text=black] {\logoc{openai}GPT-6 Astra-xhigh, 600 actions};
\filldraw[fill=wTeal!70!black,draw=white,line width=0.3pt] (0,0) ++(0,-34pt) ++(5pt,0) ++(0,2.2pt) -- ++(1.95pt,-3.35pt) -- ++(-3.9pt,0) -- cycle;
\draw[wTeal!70!black!40,line width=0.95pt] (444,0) ++(0,-34pt) -- ++(10pt,0) node[right,inner sep=1.5pt,text=black] {twin, 200 actions};
\filldraw[fill=wTeal!70!black!40,draw=white,line width=0.3pt] (444,0) ++(0,-34pt) ++(5pt,0) ++(0,2.2pt) -- ++(1.95pt,-3.35pt) -- ++(-3.9pt,0) -- cycle;
\end{tikzpicture}%
\hfill%
\begin{tikzpicture}[x=0.0880pt,y=6.0000pt,font=\scriptsize,baseline=0pt]
\node[anchor=east,inner sep=1.5pt] at (0,0) {0};
\draw[black!10,line width=0.3pt] (0,2) -- (2000,2);
\node[anchor=east,inner sep=1.5pt] at (0,2) {2};
\draw[black!10,line width=0.3pt] (0,4) -- (2000,4);
\node[anchor=east,inner sep=1.5pt] at (0,4) {4};
\draw[black!10,line width=0.3pt] (0,6) -- (2000,6);
\node[anchor=east,inner sep=1.5pt] at (0,6) {6};
\draw[black!10,line width=0.3pt] (0,8) -- (2000,8);
\node[anchor=east,inner sep=1.5pt] at (0,8) {8};
\draw[black!10,line width=0.3pt] (0,10) -- (2000,10);
\node[anchor=east,inner sep=1.5pt] at (0,10) {10};
\node[anchor=east,inner sep=1.5pt] at (0,12) {12};
\draw[black!60,line width=0.4pt] (0,0) -- ++(0,-2pt) node[below,inner sep=1.5pt] {0};
\draw[black!60,line width=0.4pt] (500,0) -- ++(0,-2pt) node[below,inner sep=1.5pt] {500};
\draw[black!60,line width=0.4pt] (1000,0) -- ++(0,-2pt) node[below,inner sep=1.5pt] {1000};
\draw[black!60,line width=0.4pt] (1500,0) -- ++(0,-2pt) node[below,inner sep=1.5pt] {1500};
\draw[black!60,line width=0.4pt] (2000,0) -- ++(0,-2pt) node[below,inner sep=1.5pt] {2000};
\draw[black!45,line width=0.4pt,dash pattern=on 1.6pt off 1.2pt] (0,12) -- (2000,12) -- (2000,0);
\draw[black!60,line width=0.4pt] (0,12) -- (0,0) -- (2000,0);
\node[anchor=north] at (1000,0) [yshift=-9pt] {paid actions};
\node[rotate=90,anchor=south] at (0,6) [xshift=0pt,yshift=7pt] {blocks};
\draw[wInk!40,line width=0.55pt,densely dotted] (0,0.000) -- (2,0.000) -- (3,0.250) -- (4,0.750) -- (5,1.000) -- (6,1.750) -- (7,2.000) -- (11,2.000) -- (12,2.250) -- (14,2.250) -- (15,2.500) -- (16,2.750) -- (26,2.750) -- (27,3.000) -- (28,3.500) -- (30,3.500) -- (31,3.750) -- (32,4.000) -- (35,4.000) -- (36,4.250) -- (37,4.250) -- (38,4.500) -- (44,4.500) -- (45,4.750) -- (56,4.750) -- (57,5.000) -- (70,5.000) -- (71,5.250) -- (74,5.250) -- (75,5.500) -- (78,5.500) -- (79,5.750) -- (89,5.750) -- (90,6.000) -- (177,6.000) -- (178,6.250) -- (200,6.250);
\draw[wInk,line width=0.55pt,densely dotted] (0,0.000) -- (2,0.000) -- (3,0.250) -- (4,0.750) -- (5,1.000) -- (6,1.750) -- (11,1.750) -- (12,2.000) -- (13,2.000) -- (14,2.250) -- (15,2.500) -- (17,2.500) -- (18,2.750) -- (22,2.750) -- (23,3.000) -- (27,3.000) -- (28,3.250) -- (29,3.500) -- (30,3.750) -- (31,4.250) -- (33,4.250) -- (34,4.500) -- (38,4.500) -- (39,5.000) -- (42,5.000) -- (43,5.250) -- (45,5.250) -- (46,5.500) -- (47,5.500) -- (48,5.750) -- (50,5.750) -- (51,6.250) -- (63,6.250) -- (64,6.500) -- (77,6.500) -- (78,7.000) -- (84,7.000) -- (85,7.250) -- (109,7.250) -- (110,7.500) -- (145,7.500) -- (146,7.750) -- (167,7.750) -- (168,8.000) -- (1997,8.000) -- (1998,8.250) -- (2000,8.250);
\draw[wInk!40,line width=0.95pt,line join=round] (0,0.000) -- (3,0.000) -- (4,0.250) -- (5,1.000) -- (6,1.250) -- (7,1.250) -- (8,1.500) -- (15,1.500) -- (16,1.750) -- (26,1.750) -- (27,2.000) -- (36,2.000) -- (37,2.250) -- (44,2.250) -- (45,2.500) -- (52,2.500) -- (53,2.750) -- (54,3.000) -- (55,3.250) -- (66,3.250) -- (67,3.500) -- (70,3.500) -- (71,3.750) -- (74,3.750) -- (75,4.000) -- (89,4.000) -- (90,4.250) -- (116,4.250) -- (117,4.500) -- (138,4.500) -- (139,4.750) -- (200,4.750);
\draw[wInk,line width=0.95pt,line join=round] (0,0.000) -- (3,0.000) -- (4,0.250) -- (5,1.000) -- (6,1.250) -- (14,1.250) -- (15,1.500) -- (23,1.500) -- (24,1.750) -- (25,2.000) -- (38,2.000) -- (39,2.250) -- (42,2.250) -- (43,2.500) -- (44,2.750) -- (47,2.750) -- (48,3.000) -- (59,3.000) -- (60,3.250) -- (77,3.250) -- (78,3.500) -- (80,3.500) -- (81,3.750) -- (114,3.750) -- (115,4.000) -- (118,4.000) -- (119,4.250) -- (120,4.250) -- (121,4.500) -- (125,4.500) -- (126,4.750) -- (141,4.750) -- (142,5.000) -- (163,5.000) -- (164,5.250) -- (167,5.250) -- (168,5.500) -- (171,5.500) -- (172,5.750) -- (1989,5.750) -- (1990,6.000) -- (1997,6.000) -- (1998,6.250) -- (2000,6.250);
\draw[wInk!85,line width=0.8pt,dash pattern=on 2.4pt off 1.4pt] (1990,0) -- (1990,12);
\draw[wInk!85,line width=0.8pt] (1990,12) -- ++(0,7.2pt);
\node[fill=wInk!85,text=white,font=\tiny,inner xsep=1.8pt,inner ysep=1pt,anchor=north east] at (1990,12) [yshift=7.2pt] {standing order};
\draw[wInk,line width=0.95pt] (0,0) ++(0,-25pt) -- ++(10pt,0) node[right,inner sep=1.5pt,text=black] {\logoc{openai}GPT-5.6 Sol, 2000 actions};
\draw[wInk!40,line width=0.95pt] (1240,0) ++(0,-25pt) -- ++(10pt,0) node[right,inner sep=1.5pt,text=black] {twin, 200 actions};
\end{tikzpicture}%
\caption{Discovery under larger action budgets on one instance. Curves give the mean over the four conditions of the grounded blocks (solid) and of the blocks whose evidence the agent holds (dotted) after each paid action, the lighter curves the same system on the same instance at 200 paid actions, and the flag marks the standing order of the larger budget.}
\label{fig:extended}
\end{figure*}

%% file: tables/tab_extended.tex
\begin{table}[t]
\centering
\caption{Larger-budget results beside their twins at 200 actions. Marks give each committed reading (a), (b) and (c) as exact (\yes) or wrong (\no), grey parentheses give the twin, and the shaded row is the exact prediction.}
\label{tab:extended}
\makeatletter
\def\extExactL#1{\pgfsys@markposition{extExactL#1}%
  \pgfsys@getposition{extExactL#1}\ext@l\pgfsys@getposition{extExactR#1}\ext@r
  \ifx\ext@l\relax\else\ifx\ext@r\relax\else
    \pgfpointdiff{\ext@l}{\ext@r}\edef\ext@w{\the\pgf@x}%
    \edef\ext@ht{\the\ht\@arstrutbox}\edef\ext@dp{\the\dp\@arstrutbox}%
    \tikz[overlay]\fill[wGold!14] (0pt,-\ext@dp) rectangle (\ext@w,\ext@ht);%
  \fi\fi}
\def\extExactR#1{\pgfsys@markposition{extExactR#1}}
\makeatother
\scriptsize
\setlength{\tabcolsep}{3pt}
\renewcommand{\arraystretch}{1.12}
\begin{tabular}{@{}llcrrr@{}}
\toprule
 & & Readings & & & \\
\multicolumn{2}{@{}l}{Condition} & (a)\,(b)\,(c) & Evidence & Blocks & Interventions\\
\midrule
\multicolumn{6}{@{}l}{\logo{anthropic}Claude Fable 5.1-xhigh, \textit{Claude Code}, 600 paid actions}\\
\pix{open}Open & \pix{distributed}Distributed & \yes\,\no\,\no & 10\,\textcolor{black!50}{(8)} & 8\,\textcolor{black!50}{(8)} & 47\\
 & \pix{aligned}Aligned & \no\,\no\,\no & 13\,\textcolor{black!50}{(10)} & 9\,\textcolor{black!50}{(9)} & 117\\
\pix{disclosed}Disclosed & \pix{distributed}Distributed & \yes\,\no\,\no & 10\,\textcolor{black!50}{(7)} & 8\,\textcolor{black!50}{(5)} & 156\\
\extExactL{1} & \pix{aligned}Aligned & \yes\,\yes\,\yes & 12\,\textcolor{black!50}{(10)} & 9\,\textcolor{black!50}{(8)} & 128\extExactR{1}\\
\midrule
\multicolumn{6}{@{}l}{\logo{openai}GPT-6 Astra-xhigh, \textit{Codex}, 600 paid actions}\\
\pix{open}Open & \pix{distributed}Distributed & \yes\,\no\,\yes & 9\,\textcolor{black!50}{(10)} & 7\,\textcolor{black!50}{(8)} & 38\\
 & \pix{aligned}Aligned & \yes\,\no\,\yes & 10\,\textcolor{black!50}{(8)} & 5\,\textcolor{black!50}{(7)} & 45\\
\pix{disclosed}Disclosed & \pix{distributed}Distributed & \no\,\no\,\yes & 10\,\textcolor{black!50}{(11)} & 10\,\textcolor{black!50}{(8)} & 123\\
 & \pix{aligned}Aligned & \yes\,\no\,\no & 7\,\textcolor{black!50}{(9)} & 6\,\textcolor{black!50}{(8)} & 104\\
\midrule
\multicolumn{6}{@{}l}{\logo{openai}GPT-5.6 Sol, \textit{baseline loop}, 2000 paid actions}\\
\pix{open}Open & \pix{distributed}Distributed & \no\,\no\,\yes & 9\,\textcolor{black!50}{(5)} & 7\,\textcolor{black!50}{(3)} & 9\\
 & \pix{aligned}Aligned & \yes\,\no\,\yes & 6\,\textcolor{black!50}{(7)} & 4\,\textcolor{black!50}{(6)} & 11\\
\pix{disclosed}Disclosed & \pix{distributed}Distributed & \no\,\no\,\yes & 10\,\textcolor{black!50}{(7)} & 8\,\textcolor{black!50}{(6)} & 10\\
 & \pix{aligned}Aligned & \no\,\no\,\no & 8\,\textcolor{black!50}{(6)} & 6\,\textcolor{black!50}{(4)} & 7\\
\bottomrule
\end{tabular}
\end{table}

%% file: sections/app_e_analyses.tex
\section{Discovery analyses and trajectories}
\label{app:analyses}

The analyses and trajectories below concern the contract budget, and the episodes at larger
budgets are analysed with their protocol in Appendix~\ref{app:extended}.

\subsection{Discovery over the hypergraph}
\label{app:heat}

The analysis of the hypergraph asks where evidence is missing, whether complete sources
become the redescription they support, and when systems write their evidence into their
notes.

\paragraph{Where evidence is missing.}
Figure~\ref{fig:coverage} reads the detectors alone over every counted episode of the main
comparison. The Open episodes together hold the evidence of \resCoverOpenHeld{} of the 18
blocks, the rarest, \resCoverOpenRarest{}, in \resCoverOpenRarestN{} of \resCoverOpenN{}
episodes. Single episodes fall short at the designed route to the prediction: of the five sources of
HC, the most one Open episode holds is \resCoverOpenHC{}, reached by \resCoverOpenHCN{}
episodes, and the Disclosed episodes reach \resCoverHCMaxDD{} (Distributed) and
\resCoverHCMaxDA{} (Aligned). Figure~\ref{fig:heat} places the shortfall on the hypergraph.
Where a main-comparison episode holds the evidence of two sources of a hyperedge, the absent
third is, with counts, \resLoopMissingSources{}: mostly the consign source for H2 (CO2) and
for H3 (CO1), and for H1, whose sources are FE1, TR1 and LD1, the fen source FE1 or the ladder
source LD1. The consign domain stays sparse in every system. Its evidence is held in
\resShareHeldConsign\% of the main-comparison episodes and a consign block is grounded in
\resShareDomainConsign\%, and the
vendor-harness systems ground one in \resFableDomainConsign{} of \resFableScored{} episodes
(Claude Fable~5.1) and \resAstraDomainConsign{} of \resAstraScored{} (GPT-6 Astra). The fen
separates them from the baseline loop. They ground a fen block in \resFableDomainFen{} and
\resAstraDomainFen{} of their episodes, against \resShareDomainFen\% of the main comparison.
For Claude Fable~5.1 the absent sources are \resFableMissingSources{}. For GPT-6 Astra they
are \resAstraMissingSources{}. In both, LD1 takes the place of FE1 as the more frequent gap
in H1.

\IfFileExists{figures/fig_coverage.tex}{\input{figures/fig_coverage}}{}
\IfFileExists{figures/fig_heat.tex}{\input{figures/fig_heat}}{}
\IfFileExists{tables/tab_evidenced.tex}{\input{tables/tab_evidenced}}{}

\paragraph{Whether complete sources become a redescription.}
Table~\ref{tab:evidenced} lists the episodes that hold the complete evidence of a hyperedge. In
the main comparison \resReachThreeN{} of \resScoredMain{} episodes hold it and
\resEvidencedEpisodes{} of these state the change it supports. The vendor-harness systems hold
it in \resFableReachFullN{} and \resAstraReachFullN{} episodes and state the change in
\resFableEvidencedN{} and \resAstraEvidencedN{}, and the scaffolded episodes hold it in
\resScafReachFull{} and state it in \resScafReachFullStated{}. In most episodes a source is
missing. Among the few with complete sources, whether the change is stated depends on the
change and the condition. Under disclosure every complete hyperedge comes with its stated
change, the document having supplied the representation. In the Open conditions every complete
H1 comes with the retraction it supports, in the main comparison and the vendor-harness systems
alike. Of the Open episodes that complete H2 or H3, the Claude Fable~5.1 episode of
Appendix~\ref{app:qual-fable} states the rewritten variables, and the others leave the hidden
label and the rewritten variables unstated.

\IfFileExists{figures/fig_coupling.tex}{\input{figures/fig_coupling}}{}

\paragraph{When evidence enters the notes.}
The baseline loop grounds most of its blocks early: \resSharePreWindow\% of them are grounded
before the standing order is posted, and its curves flatten well before the budget ends
(Figure~\ref{fig:curves}). The vendor-harness systems keep producing evidence after the loop's
curves flatten, and more of their grounding comes at the end: \resVendorSharePreWindow\% of
their grounded blocks are grounded before the order. The ledger of
Figure~\ref{fig:coupling} gives, block by block, the share of episodes that hold the evidence
and that ground the block, and the median paid action of each. The gap is widest at TR1, whose
evidence is a passive reading that most episodes take and few state. The matrix beside the
ledger gives which blocks the loop grounds together, and since most blocks are grounded in few
episodes, a pair defined in fewer than three models (hatched) is left without a reading.

\subsection{The exact prediction}
\label{app:exact-episode}

One episode of the study commits the exact tuple: Claude Fable~5.1 in Claude Code, Disclosed
$\times$ Aligned, at \resExtBudgetVendor{} paid actions (Appendix~\ref{app:extended}). Its record holds the
evidence of \resExtExactOrderHeld{} of the five blocks on the designed route to the order's
readings, every one except FE3, the matched conservation at the fen. Its notes state
\resExtExactStated{} of the three changes, the retraction and the postulate, and carry the turn
order of each of the nine individuals left after the caretakers' rewrite at tick 20, the
variable RC3 names, without the statement the criterion for RC3 asks for.

\IfFileExists{figures/fig_exact_notes.tex}{\input{figures/fig_exact_notes}}{}

The derivation is in the notes (Figure~\ref{fig:exact-notes}). The agent writes the rewriting
law in its own terms, the schedule of the caretakers' rewrites and the stall, and from these
computes the three readings, the order's act falling at tick 3 on this instance. The order's
pass reads 16, 1 and 20. The notes serve the order's acts a tick after the caretakers perform
them (Appendix~\ref{app:world-capstone}), and on this instance both timings give these
figures, since under either feeder-3 acts on every tick from tick 3 and the beacon first
answers at tick 4.

Field (a) rests on Laws 2, 3 and 5 and feeder-3's own step per act,
with no use of the matched conservation by which the designed route reaches it: the design's
route to the readings is sufficient and it is not the only one.

\IfFileExists{figures/fig_walkthrough.tex}{\input{figures/fig_walkthrough}}{}

\subsection{Four trajectories, from observation to transfer}
\label{app:qualitative}
\label{app:qual-endpoint}

At the contract budget no episode makes the exact prediction, so the endpoint records that an
agent fell short and is silent on how. We read four complete trajectories, in two
pairs that each hold one instance fixed, so that the contrasts within a pair are contrasts of
process.

The first pair comes from the baseline loop in the primary condition,
\pix{open}Open~$\times$~\pix{distributed}Distributed. The GPT-5.6 Sol episode is the one with
the most intervention requests among Sol's twelve primary episodes, ten in all. It is paired
on the same instance with Gemini 3.6 Flash, which requests no intervention there, as in all
twelve of its primary episodes. The second pair comes from the two vendor-harness systems of
Appendix~\ref{app:agentic}, on one instance in \pix{open}Open~$\times$~\pix{aligned}Aligned.
The Claude Fable~5.1 episode is the one Open episode of that system whose notes state a
representation change with the evidence of its hyperedge complete. The GPT-6 Astra episode on
the same instance is the one episode of either system that yields part of its budget.

\subsection{An account revised, then overextended}
\label{app:qual-sol}

Sol makes 75 \pix{read}reads and ten intervention requests (Figure~\ref{fig:walkthrough}). It
checks its first experiment against a baseline. At action~47 it \pix{feed}feeds a
\pix{grazer}grazer, and two actions later the audit counter reads 17 where it had read 16.
The note stored at action~51 withdraws the causal reading, ``Correction: the feed test did
not establish causality'', since the pass ledger shows the same rise at tick 3 of the
unmanipulated pass.

The later experiments are controlled contrasts on the registry lines. Sol \pix{cull}culls
four lines in turn, line 1 twice, and reads the audit counter against the unmanipulated
pass. Culled at tick 7, line 1 raises the count to 18 at tick 8, one above the baseline,
and to 20 at tick 19, again one above; at tick 20 the count falls to 9, as it does in the
baseline. Culled again at tick 17, the same line gives 20 at ticks 18 and 19 and 9 from
tick 20 (Figure~\ref{fig:reasoning}, panel A). The note stored at action~136 reads the
first of these contrasts as ``confirming a component split.'' From action~102 on, the
notes call the registry lines bridges that a cull removes, and the final account keeps
that description.

The contrasts establish that a cull changes the count for part of the pass. The account
built on them goes further, to what registry lines are and what a cull does to them, and
the prediction that rests on it fails. After yielding its remaining inquiry budget at
action~146, Sol commits $(0, 8, 16)$ with its last paid action, against the true
$(35, 3, 18)$.

\IfFileExists{figures/fig_reasoning.tex}{\input{figures/fig_reasoning}}{}

\subsection{An explanation without an intervention}
\label{app:qual-flash}

Flash makes 22 reads and requests no intervention. At action~27 it reads the pass ledger:
the audit counter stands at 19 from tick 17 to tick 19 and at 9 from tick 20 to tick 23.
The note stored at action~58 records the fall ``from 19 to 9''. The note stored at
action~59 turns it into an explanation of the stall, a capacity that drops ``below the
critical threshold required for sustained maintenance throughput''
(Figure~\ref{fig:reasoning}, panel B).

The threshold and its causal role stay untested. At action~59 Flash yields its inquiry
budget, 131 paid actions, and commits $(9, 1, 16)$ against the same true $(35, 3, 18)$.

\subsection{A working account, rewritten}
\label{app:qual-fable}

Claude Fable~5.1 in Claude Code makes \resCaseFableReads{} reads and
\resCaseFableInterventions{} intervention requests, \resCaseFableStamps{} of them
\pix{stamp}stamps. Its first pass visits all \resCaseFableSweep{} sites, and its first
intervention comes at action~\resCaseFableFirstAct{}. From then on it writes one long note
per pass, headed by the pass, that holds the pass's readings, its acts and what they changed.
The note stored at action~\resCaseFableModelAt{} sets out a working model and marks it
untested. At action~\resCaseFableRewriteAt{} the agent rewrites its record of the passes as a
working account: a map of the sites, then numbered observations, to which later notes add.

The experiments return to their own results. One stamp, on a \pix{printer}printer, gives a
result that the next note leaves open, and in a later pass the agent repeats the stamp at the
same tick and records that the result reproduces. In pass~\resCaseFableSlatePass{} it reads
the manifest slate \resCaseFableSlateReads{} times around the \pix{release}release of a
\pix{cargo}consignment, the one Open episode of the system to hold evidence in the consign
domain. Its notes state \resCaseFableEvidenced{}, the rewritten variables, with the evidence
of all three sources of \resCaseFableEdges{} in hand.

After the standing order the notes derive each of the three readings from the numbered
observations and close with ``Main risks'', one alternative value per reading. Fable commits
\resCaseFableCommit{} against the true \resCaseFableTrue{}. The first two readings are exact,
and the first risk its note lists is the true value of the third.

\subsection{A budget yielded after testing}
\label{app:qual-astra}

GPT-6 Astra in Codex makes \resCaseAstraReads{} reads and \resCaseAstraInterventions{}
intervention requests. Its first pass visits \resCaseAstraSweep{} of the fifteen sites, and
its first intervention comes at action~\resCaseAstraFirstAct{}. Its notes are short entries
appended after each group of acts, each stating what the acts showed and what to test next.
The note stored at action~\resCaseAstraWarnAt{} warns ``Do not assume issued maintenance has
taken effect'', and the note stored at action~\resCaseAstraWithdrawAt{} adds the converse,
``request\_pending is not proof of no effect''. The agent begins to tell the state of a
request apart from its effect.

At action~\resCaseAstraYieldAt{} Astra yields its remaining inquiry budget,
\resCaseAstraYielded{} paid actions, and with the yield it stores a note headed ``CURRENT
EXPLANATION''. The standing order arrives with ten paid actions left, and the note stored at
action~\resCaseAstraOrderAt{} reads it as ``a critical exception'' to that explanation. With
the actions that remain Astra runs one more experiment. It \pix{retool}retools the feeder
beside the one the order names, reads the intake tallyboard at the tick the first reading
concerns, and the note stored at action~\resCaseAstraPeerAt{}, a ``Permitted peer
experiment'', carries the result over to the named feeder.

Against the true \resCaseAstraTrue{}, Astra commits \resCaseAstraCommit{}. Its final note calls
the assumption behind the third reading ``the least directly tested component'', and the third
reading is the one it gets right.

\subsection{What the four trajectories separate}
\label{app:qual-pair}

Sol and Flash fail on the same instance by different routes. Flash writes an explanation
and leaves it untested: its failure is insufficient testing. Sol tests, corrects itself,
and then extends its account beyond what its contrasts establish: its failure is an
explanation that remains inadequate after testing. The endpoint gives both the same
verdict. The metrics of Section~\ref{sec:scoring} separate them. Flash holds the evidence
of \resCaseFlashEvidence{} blocks and grounds \resCaseFlashBlocks{}. Sol holds the
evidence of \resCaseSolEvidence{} and grounds \resCaseSolBlocks{}, the three bridges among
them, and its Reach is \resCaseSolReach{}: it holds the evidence of two of the three sources of
one representation change, one source short of the complete evidence for a redescription.

The vendor-harness episodes keep a notebook as a laboratory record. Fable rewrites its notes
into an account of numbered observations, Astra writes an entry after each group of acts, and
both test with contrasts repeated across passes. Fable holds the evidence of
\resCaseFableEvidence{} blocks and grounds \resCaseFableBlocks{}, and its Reach of
\resCaseFableReach{} is the complete evidence of \resCaseFableEdges{}, which its notes turn
into a stated change. Astra holds the evidence of \resCaseAstraEvidence{}, grounds
\resCaseAstraBlocks{} and reaches \resCaseAstraReach{}. Their shortfalls echo the two routes of
the loop episodes, with more of the evidence in hand. Astra yields its budget as Flash does,
though after testing, and builds its first reading on an experiment with a neighbouring
object, an account that outruns its evidence as Sol's does. Fable names the true value of its
one wrong reading among its risks, and Astra's least tested reading is its one exact reading.

%% file: figures/fig_coverage.tex
\begin{figure}[t]
\centering
\begin{tikzpicture}[x=1pt,y=1pt,font=\scriptsize]
\node[anchor=base] at (79.25,17) {\pix{open}Open};
\draw[black!40,line width=0.35pt] (35.50,14.2) -- (123.00,14.2);
\node[anchor=base] at (172.25,17) {\pix{disclosed}Disclosed};
\draw[black!40,line width=0.35pt] (128.50,14.2) -- (216.00,14.2);
\node[anchor=base] at (56.00,5.8) {\pix{distributed}Distributed};
\fill[cOpenDist] (34.00,2.2) rectangle (78.00,3.6);
\node[anchor=base] at (102.50,5.8) {\pix{aligned}Aligned};
\fill[cOpenAlig] (80.50,2.2) rectangle (124.50,3.6);
\node[anchor=base] at (149.00,5.8) {\pix{distributed}Distributed};
\fill[cDiscDist] (127.00,2.2) rectangle (171.00,3.6);
\node[anchor=base] at (195.50,5.8) {\pix{aligned}Aligned};
\fill[cDiscAlig] (173.50,2.2) rectangle (217.50,3.6);
\fill[dFen!45,rounded corners=1.2pt] (10,-0.00) rectangle (32,-32.00);
\node[inner sep=0pt] at (4.6,-16.00) {\pix{fen}};
\node[anchor=west,inner sep=0pt] at (11.6,-5.00) {FE1};
\fill[cOpenDist!14] (34.00,-0.00) rectangle (78.00,-10.00);
\node[text=black,inner sep=0pt] at (56.00,-4.00) {8};
\fill[black!55] (35.50,-9.10) rectangle (50.41,-8.50);
\fill[cOpenAlig!13] (80.50,-0.00) rectangle (124.50,-10.00);
\node[text=black,inner sep=0pt] at (102.50,-4.00) {6};
\fill[black!55] (82.00,-9.10) rectangle (89.45,-8.50);
\fill[cDiscDist!12] (127.00,-0.00) rectangle (171.00,-10.00);
\node[text=black,inner sep=0pt] at (149.00,-4.00) {5};
\fill[black!55] (128.50,-9.10) rectangle (139.68,-8.50);
\fill[cDiscAlig!13] (173.50,-0.00) rectangle (217.50,-10.00);
\node[text=black,inner sep=0pt] at (195.50,-4.00) {6};
\fill[black!55] (175.00,-9.10) rectangle (189.91,-8.50);
\node[anchor=west,inner sep=0pt] at (11.6,-16.00) {FE2};
\fill[cOpenDist!18] (34.00,-11.00) rectangle (78.00,-21.00);
\node[text=black,inner sep=0pt] at (56.00,-15.00) {12};
\fill[black!55] (35.50,-20.10) rectangle (46.68,-19.50);
\fill[cOpenAlig!13] (80.50,-11.00) rectangle (124.50,-21.00);
\node[text=black,inner sep=0pt] at (102.50,-15.00) {6};
\fill[black!55] (82.00,-20.10) rectangle (89.45,-19.50);
\fill[cDiscDist!15] (127.00,-11.00) rectangle (171.00,-21.00);
\node[text=black,inner sep=0pt] at (149.00,-15.00) {9};
\fill[black!55] (128.50,-20.10) rectangle (147.14,-19.50);
\fill[cDiscAlig!12] (173.50,-11.00) rectangle (217.50,-21.00);
\node[text=black,inner sep=0pt] at (195.50,-15.00) {5};
\fill[black!55] (175.00,-20.10) rectangle (186.18,-19.50);
\node[anchor=west,inner sep=0pt] at (11.6,-27.00) {FE3};
\node[inner sep=0pt] at (28.8,-27.00) {\tiny\pix{stamp}};
\fill[cOpenDist!11] (34.00,-22.00) rectangle (78.00,-32.00);
\node[text=black,inner sep=0pt] at (56.00,-26.00) {4};
\fill[black!55] (35.50,-31.10) rectangle (46.68,-30.50);
\fill[cOpenAlig!11] (80.50,-22.00) rectangle (124.50,-32.00);
\node[text=black,inner sep=0pt] at (102.50,-26.00) {3};
\fill[black!55] (82.00,-31.10) rectangle (89.45,-30.50);
\fill[cDiscDist!10] (127.00,-22.00) rectangle (171.00,-32.00);
\node[text=black,inner sep=0pt] at (149.00,-26.00) {3};
\fill[black!55] (128.50,-31.10) rectangle (132.23,-30.50);
\fill[cDiscAlig!12] (173.50,-22.00) rectangle (217.50,-32.00);
\node[text=black,inner sep=0pt] at (195.50,-26.00) {5};
\fill[black!55] (175.00,-31.10) rectangle (186.18,-30.50);
\fill[dTrellis!45,rounded corners=1.2pt] (10,-35.40) rectangle (32,-67.40);
\node[inner sep=0pt] at (4.6,-51.40) {\pix{trellis}};
\node[anchor=west,inner sep=0pt] at (11.6,-40.40) {TR1};
\fill[cOpenDist!84] (34.00,-35.40) rectangle (78.00,-45.40);
\node[text=white,inner sep=0pt] at (56.00,-39.40) {95};
\fill[white] (35.50,-44.50) rectangle (76.50,-43.90);
\fill[cOpenAlig!88] (80.50,-35.40) rectangle (124.50,-45.40);
\node[text=black,inner sep=0pt] at (102.50,-39.40) {100};
\fill[black!55] (82.00,-44.50) rectangle (123.00,-43.90);
\fill[cDiscDist!70] (127.00,-35.40) rectangle (171.00,-45.40);
\node[text=black,inner sep=0pt] at (149.00,-39.40) {77};
\fill[black!55] (128.50,-44.50) rectangle (169.50,-43.90);
\fill[cDiscAlig!84] (173.50,-35.40) rectangle (217.50,-45.40);
\node[text=black,inner sep=0pt] at (195.50,-39.40) {95};
\fill[black!55] (175.00,-44.50) rectangle (216.00,-43.90);
\node[anchor=west,inner sep=0pt] at (11.6,-51.40) {TR2};
\draw[black!25,line width=0.3pt,fill=white] (34.00,-46.40) rectangle (78.00,-56.40);
\node[inner sep=0pt] at (56.00,-51.40) {\no};
\fill[cOpenAlig!9] (80.50,-46.40) rectangle (124.50,-56.40);
\node[text=black,inner sep=0pt] at (102.50,-50.40) {1};
\fill[black!55] (82.00,-55.50) rectangle (85.73,-54.90);
\draw[black!25,line width=0.3pt,fill=white] (127.00,-46.40) rectangle (171.00,-56.40);
\node[inner sep=0pt] at (149.00,-51.40) {\no};
\draw[black!25,line width=0.3pt,fill=white] (173.50,-46.40) rectangle (217.50,-56.40);
\node[inner sep=0pt] at (195.50,-51.40) {\no};
\node[anchor=west,inner sep=0pt] at (11.6,-62.40) {TR3};
\node[inner sep=0pt] at (28.8,-62.40) {\tiny\pix{stamp}};
\draw[black!25,line width=0.3pt,fill=white] (34.00,-57.40) rectangle (78.00,-67.40);
\node[inner sep=0pt] at (56.00,-62.40) {\no};
\fill[cOpenAlig!9] (80.50,-57.40) rectangle (124.50,-67.40);
\node[text=black,inner sep=0pt] at (102.50,-61.40) {2};
\fill[black!55] (82.00,-66.50) rectangle (85.73,-65.90);
\fill[cDiscDist!9] (127.00,-57.40) rectangle (171.00,-67.40);
\node[text=black,inner sep=0pt] at (149.00,-61.40) {2};
\fill[black!55] (128.50,-66.50) rectangle (135.95,-65.90);
\fill[cDiscAlig!16] (173.50,-57.40) rectangle (217.50,-67.40);
\node[text=black,inner sep=0pt] at (195.50,-61.40) {10};
\fill[black!55] (175.00,-66.50) rectangle (197.36,-65.90);
\fill[dLadder!45,rounded corners=1.2pt] (10,-70.80) rectangle (32,-102.80);
\node[inner sep=0pt] at (4.6,-86.80) {\pix{ladder}};
\node[anchor=west,inner sep=0pt] at (11.6,-75.80) {LD1};
\fill[cOpenDist!19] (34.00,-70.80) rectangle (78.00,-80.80);
\node[text=black,inner sep=0pt] at (56.00,-74.80) {14};
\fill[black!55] (35.50,-79.90) rectangle (65.32,-79.30);
\fill[cOpenAlig!20] (80.50,-70.80) rectangle (124.50,-80.80);
\node[text=black,inner sep=0pt] at (102.50,-74.80) {15};
\fill[black!55] (82.00,-79.90) rectangle (115.55,-79.30);
\fill[cDiscDist!12] (127.00,-70.80) rectangle (171.00,-80.80);
\node[text=black,inner sep=0pt] at (149.00,-74.80) {5};
\fill[black!55] (128.50,-79.90) rectangle (147.14,-79.30);
\fill[cDiscAlig!12] (173.50,-70.80) rectangle (217.50,-80.80);
\node[text=black,inner sep=0pt] at (195.50,-74.80) {5};
\fill[black!55] (175.00,-79.90) rectangle (189.91,-79.30);
\node[anchor=west,inner sep=0pt] at (11.6,-86.80) {LD2};
\fill[cOpenDist!9] (34.00,-81.80) rectangle (78.00,-91.80);
\node[text=black,inner sep=0pt] at (56.00,-85.80) {1};
\fill[black!55] (35.50,-90.90) rectangle (39.23,-90.30);
\draw[black!25,line width=0.3pt,fill=white] (80.50,-81.80) rectangle (124.50,-91.80);
\node[inner sep=0pt] at (102.50,-86.80) {\no};
\fill[cDiscDist!9] (127.00,-81.80) rectangle (171.00,-91.80);
\node[text=black,inner sep=0pt] at (149.00,-85.80) {1};
\fill[black!55] (128.50,-90.90) rectangle (132.23,-90.30);
\fill[cDiscAlig!10] (173.50,-81.80) rectangle (217.50,-91.80);
\node[text=black,inner sep=0pt] at (195.50,-85.80) {2};
\fill[black!55] (175.00,-90.90) rectangle (182.45,-90.30);
\node[anchor=west,inner sep=0pt] at (11.6,-97.80) {LD3};
\fill[cOpenDist!42] (34.00,-92.80) rectangle (78.00,-102.80);
\node[text=black,inner sep=0pt] at (56.00,-96.80) {42};
\fill[black!55] (35.50,-101.90) rectangle (69.05,-101.30);
\fill[cOpenAlig!61] (80.50,-92.80) rectangle (124.50,-102.80);
\node[text=black,inner sep=0pt] at (102.50,-96.80) {67};
\fill[black!55] (82.00,-101.90) rectangle (115.55,-101.30);
\fill[cDiscDist!58] (127.00,-92.80) rectangle (171.00,-102.80);
\node[text=black,inner sep=0pt] at (149.00,-96.80) {63};
\fill[black!55] (128.50,-101.90) rectangle (165.77,-101.30);
\fill[cDiscAlig!72] (173.50,-92.80) rectangle (217.50,-102.80);
\node[text=black,inner sep=0pt] at (195.50,-96.80) {79};
\fill[black!55] (175.00,-101.90) rectangle (212.27,-101.30);
\fill[dConsign!45,rounded corners=1.2pt] (10,-106.20) rectangle (32,-138.20);
\node[inner sep=0pt] at (4.6,-122.20) {\pix{consign}};
\node[anchor=west,inner sep=0pt] at (11.6,-111.20) {CO1};
\node[inner sep=0pt] at (28.8,-111.20) {\tiny\pix{stamp}};
\fill[cOpenDist!16] (34.00,-106.20) rectangle (78.00,-116.20);
\node[text=black,inner sep=0pt] at (56.00,-110.20) {10};
\fill[black!55] (35.50,-115.30) rectangle (61.59,-114.70);
\fill[cOpenAlig!11] (80.50,-106.20) rectangle (124.50,-116.20);
\node[text=black,inner sep=0pt] at (102.50,-110.20) {4};
\fill[black!55] (82.00,-115.30) rectangle (93.18,-114.70);
\fill[cDiscDist!10] (127.00,-106.20) rectangle (171.00,-116.20);
\node[text=black,inner sep=0pt] at (149.00,-110.20) {2};
\fill[black!55] (128.50,-115.30) rectangle (139.68,-114.70);
\fill[cDiscAlig!9] (173.50,-106.20) rectangle (217.50,-116.20);
\node[text=black,inner sep=0pt] at (195.50,-110.20) {2};
\fill[black!55] (175.00,-115.30) rectangle (182.45,-114.70);
\node[anchor=west,inner sep=0pt] at (11.6,-122.20) {CO2};
\fill[cOpenDist!9] (34.00,-117.20) rectangle (78.00,-127.20);
\node[text=black,inner sep=0pt] at (56.00,-121.20) {2};
\fill[black!55] (35.50,-126.30) rectangle (42.95,-125.70);
\fill[cOpenAlig!9] (80.50,-117.20) rectangle (124.50,-127.20);
\node[text=black,inner sep=0pt] at (102.50,-121.20) {1};
\fill[black!55] (82.00,-126.30) rectangle (85.73,-125.70);
\fill[cDiscDist!10] (127.00,-117.20) rectangle (171.00,-127.20);
\node[text=black,inner sep=0pt] at (149.00,-121.20) {3};
\fill[black!55] (128.50,-126.30) rectangle (143.41,-125.70);
\draw[black!25,line width=0.3pt,fill=white] (173.50,-117.20) rectangle (217.50,-127.20);
\node[inner sep=0pt] at (195.50,-122.20) {\no};
\node[anchor=west,inner sep=0pt] at (11.6,-133.20) {CO3};
\fill[cOpenDist!15] (34.00,-128.20) rectangle (78.00,-138.20);
\node[text=black,inner sep=0pt] at (56.00,-132.20) {9};
\fill[black!55] (35.50,-137.30) rectangle (50.41,-136.70);
\fill[cOpenAlig!11] (80.50,-128.20) rectangle (124.50,-138.20);
\node[text=black,inner sep=0pt] at (102.50,-132.20) {4};
\fill[black!55] (82.00,-137.30) rectangle (100.64,-136.70);
\fill[cDiscDist!9] (127.00,-128.20) rectangle (171.00,-138.20);
\node[text=black,inner sep=0pt] at (149.00,-132.20) {1};
\fill[black!55] (128.50,-137.30) rectangle (132.23,-136.70);
\draw[black!25,line width=0.3pt,fill=white] (173.50,-128.20) rectangle (217.50,-138.20);
\node[inner sep=0pt] at (195.50,-133.20) {\no};
\fill[dCadastre!45,rounded corners=1.2pt] (10,-141.60) rectangle (32,-173.60);
\node[inner sep=0pt] at (4.6,-157.60) {\pix{cadastre}};
\node[anchor=west,inner sep=0pt] at (11.6,-146.60) {CA1};
\fill[cOpenDist!88] (34.00,-141.60) rectangle (78.00,-151.60);
\node[text=white,inner sep=0pt] at (56.00,-145.60) {100};
\fill[white] (35.50,-150.70) rectangle (76.50,-150.10);
\fill[cOpenAlig!88] (80.50,-141.60) rectangle (124.50,-151.60);
\node[text=black,inner sep=0pt] at (102.50,-145.60) {100};
\fill[black!55] (82.00,-150.70) rectangle (123.00,-150.10);
\fill[cDiscDist!88] (127.00,-141.60) rectangle (171.00,-151.60);
\node[text=white,inner sep=0pt] at (149.00,-145.60) {100};
\fill[white] (128.50,-150.70) rectangle (169.50,-150.10);
\fill[cDiscAlig!88] (173.50,-141.60) rectangle (217.50,-151.60);
\node[text=black,inner sep=0pt] at (195.50,-145.60) {100};
\fill[black!55] (175.00,-150.70) rectangle (216.00,-150.10);
\node[anchor=west,inner sep=0pt] at (11.6,-157.60) {CA2};
\fill[cOpenDist!19] (34.00,-152.60) rectangle (78.00,-162.60);
\node[text=black,inner sep=0pt] at (56.00,-156.60) {14};
\fill[black!55] (35.50,-161.70) rectangle (61.59,-161.10);
\fill[cOpenAlig!17] (80.50,-152.60) rectangle (124.50,-162.60);
\node[text=black,inner sep=0pt] at (102.50,-156.60) {11};
\fill[black!55] (82.00,-161.70) rectangle (104.36,-161.10);
\fill[cDiscDist!39] (127.00,-152.60) rectangle (171.00,-162.60);
\node[text=black,inner sep=0pt] at (149.00,-156.60) {38};
\fill[black!55] (128.50,-161.70) rectangle (162.05,-161.10);
\fill[cDiscAlig!44] (173.50,-152.60) rectangle (217.50,-162.60);
\node[text=black,inner sep=0pt] at (195.50,-156.60) {45};
\fill[black!55] (175.00,-161.70) rectangle (208.55,-161.10);
\node[anchor=west,inner sep=0pt] at (11.6,-168.60) {CA3};
\node[inner sep=0pt] at (28.8,-168.60) {\tiny\pix{stamp}};
\fill[cOpenDist!87] (34.00,-163.60) rectangle (78.00,-173.60);
\node[text=white,inner sep=0pt] at (56.00,-167.60) {98};
\fill[white] (35.50,-172.70) rectangle (76.50,-172.10);
\fill[cOpenAlig!87] (80.50,-163.60) rectangle (124.50,-173.60);
\node[text=black,inner sep=0pt] at (102.50,-167.60) {99};
\fill[black!55] (82.00,-172.70) rectangle (123.00,-172.10);
\fill[cDiscDist!70] (127.00,-163.60) rectangle (171.00,-173.60);
\node[text=black,inner sep=0pt] at (149.00,-167.60) {78};
\fill[black!55] (128.50,-172.70) rectangle (169.50,-172.10);
\fill[cDiscAlig!86] (173.50,-163.60) rectangle (217.50,-173.60);
\node[text=black,inner sep=0pt] at (195.50,-167.60) {97};
\fill[black!55] (175.00,-172.70) rectangle (216.00,-172.10);
\fill[wGold!45,rounded corners=1.2pt] (10,-177.00) rectangle (32,-209.00);
\node[anchor=west,inner sep=0pt] at (11.6,-182.00) {XC1};
\fill[cOpenDist!66] (34.00,-177.00) rectangle (78.00,-187.00);
\node[text=black,inner sep=0pt] at (56.00,-181.00) {73};
\fill[black!55] (35.50,-186.10) rectangle (72.77,-185.50);
\fill[cOpenAlig!66] (80.50,-177.00) rectangle (124.50,-187.00);
\node[text=black,inner sep=0pt] at (102.50,-181.00) {72};
\fill[black!55] (82.00,-186.10) rectangle (119.27,-185.50);
\fill[cDiscDist!80] (127.00,-177.00) rectangle (171.00,-187.00);
\node[text=black,inner sep=0pt] at (149.00,-181.00) {90};
\fill[black!55] (128.50,-186.10) rectangle (169.50,-185.50);
\fill[cDiscAlig!81] (173.50,-177.00) rectangle (217.50,-187.00);
\node[text=black,inner sep=0pt] at (195.50,-181.00) {91};
\fill[black!55] (175.00,-186.10) rectangle (216.00,-185.50);
\node[anchor=west,inner sep=0pt] at (11.6,-193.00) {XC2};
\node[inner sep=0pt] at (28.8,-193.00) {\tiny\pix{stamp}};
\fill[cOpenDist!16] (34.00,-188.00) rectangle (78.00,-198.00);
\node[text=black,inner sep=0pt] at (56.00,-192.00) {9};
\fill[black!55] (35.50,-197.10) rectangle (54.14,-196.50);
\fill[cOpenAlig!27] (80.50,-188.00) rectangle (124.50,-198.00);
\node[text=black,inner sep=0pt] at (102.50,-192.00) {24};
\fill[black!55] (82.00,-197.10) rectangle (104.36,-196.50);
\fill[cDiscDist!51] (127.00,-188.00) rectangle (171.00,-198.00);
\node[text=black,inner sep=0pt] at (149.00,-192.00) {54};
\fill[black!55] (128.50,-197.10) rectangle (162.05,-196.50);
\fill[cDiscAlig!62] (173.50,-188.00) rectangle (217.50,-198.00);
\node[text=black,inner sep=0pt] at (195.50,-192.00) {68};
\fill[black!55] (175.00,-197.10) rectangle (208.55,-196.50);
\node[anchor=west,inner sep=0pt] at (11.6,-204.00) {XC3};
\fill[cOpenDist!14] (34.00,-199.00) rectangle (78.00,-209.00);
\node[text=black,inner sep=0pt] at (56.00,-203.00) {7};
\fill[black!55] (35.50,-208.10) rectangle (46.68,-207.50);
\fill[cOpenAlig!16] (80.50,-199.00) rectangle (124.50,-209.00);
\node[text=black,inner sep=0pt] at (102.50,-203.00) {10};
\fill[black!55] (82.00,-208.10) rectangle (104.36,-207.50);
\fill[cDiscDist!33] (127.00,-199.00) rectangle (171.00,-209.00);
\node[text=black,inner sep=0pt] at (149.00,-203.00) {32};
\fill[black!55] (128.50,-208.10) rectangle (158.32,-207.50);
\fill[cDiscAlig!36] (173.50,-199.00) rectangle (217.50,-209.00);
\node[text=black,inner sep=0pt] at (195.50,-203.00) {35};
\fill[black!55] (175.00,-208.10) rectangle (204.82,-207.50);
\node[anchor=base,text=black!70] at (56.00,-217.50) {16 of 18 held};
\node[anchor=base,text=black!70] at (102.50,-217.50) {17 of 18 held};
\node[anchor=base,text=black!70] at (149.00,-217.50) {17 of 18 held};
\node[anchor=base,text=black!70] at (195.50,-217.50) {15 of 18 held};
\end{tikzpicture}
\caption{Block-level evidence coverage by condition. Tiles give the percentage of episodes whose detector holds the block, and crosses mark zero coverage. The rule under a tile gives the share of models that hold the block in some episode, \pix{stamp}marks the five sources of HC, and the last row counts the blocks held in any episode.}
\label{fig:coverage}
\end{figure}

%% file: figures/fig_heat.tex
\begin{figure*}[t]
\centering
\includegraphics[width=\textwidth]{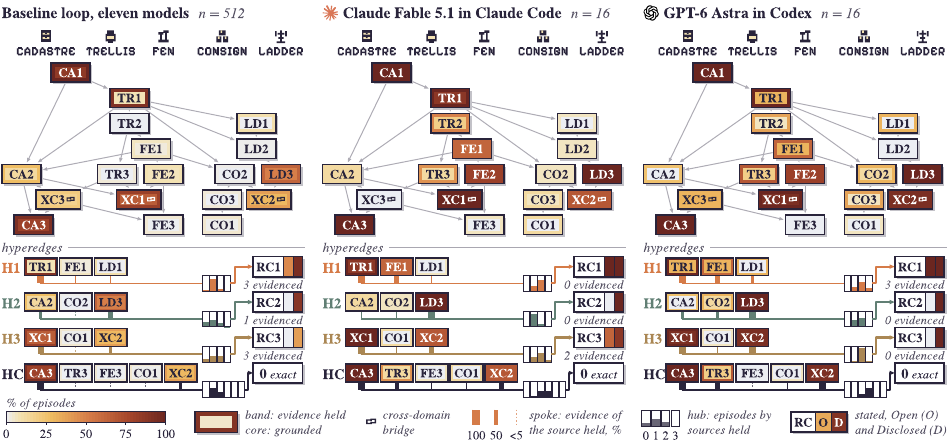}
\caption{Discovery coverage over the epistemic hypergraph of Figure~\ref{fig:hypergraph}, one
panel per system. A block's core gives the share of episodes that ground it and its band the
share that hold its evidence, on one ramp. Along each hyperedge a spoke widens with the share
of episodes holding the evidence of its source, and the hub gauge divides the episodes by the
number of sources they hold. Each representation change gives the share of episodes that state
it in the Open (O) and Disclosed (D) conditions and the number of evidenced changes; the
terminal prediction counts exact predictions.}
\label{fig:heat}
\end{figure*}

%% file: tables/tab_evidenced.tex
\begin{table*}[!t]
\centering
\caption{Episodes that hold the complete evidence of a hyperedge. Each row gives the system and condition, the hyperedge with its representation change, the number of such episodes and whether their notes state the change. An episode may contribute to more than one hyperedge.}
\label{tab:evidenced}
\footnotesize
\setlength{\tabcolsep}{7pt}
\renewcommand{\arraystretch}{1.12}
\begin{tabular}{@{}llllrc@{}}
\toprule
System & Condition & Hyperedge & Change & Episodes & Stated\\
\midrule
\logo{openai}GPT-5.6 Sol & \pix{open}Open\,$\times$\,\pix{distributed}Distributed & H2 & RC2 & 1 & \no\\
\logo{openai}GPT-5.6 Sol & \pix{disclosed}Disclosed\,$\times$\,\pix{distributed}Distributed & H3 & RC3 & 1 & \yes\\
\logo{anthropic}Claude Opus 5 & \pix{open}Open\,$\times$\,\pix{distributed}Distributed & H1 & RC1 & 1 & \yes\\
\logo{anthropic}Claude Opus 5 & \pix{open}Open\,$\times$\,\pix{aligned}Aligned & H1 & RC1 & 1 & \yes\\
\logo{anthropic}Claude Opus 5 & \pix{disclosed}Disclosed\,$\times$\,\pix{distributed}Distributed & H2 & RC2 & 1 & \yes\\
\logo{anthropic}Claude Opus 5 & \pix{disclosed}Disclosed\,$\times$\,\pix{distributed}Distributed & H3 & RC3 & 1 & \yes\\
\logo{zai}GLM-5.2 & \pix{open}Open\,$\times$\,\pix{distributed}Distributed & H3 & RC3 & 1 & \no\\
\logo{zai}GLM-5.2 & \pix{open}Open\,$\times$\,\pix{aligned}Aligned & H1 & RC1 & 1 & \yes\\
\logo{moonshot}Kimi K3 & \pix{open}Open\,$\times$\,\pix{aligned}Aligned & H3 & RC3 & 3 & \no\\
\logo{moonshot}Kimi K3 & \pix{disclosed}Disclosed\,$\times$\,\pix{aligned}Aligned & H3 & RC3 & 1 & \yes\\
\midrule
\logo{anthropic}Claude Fable 5.1, \textit{Claude Code} & \pix{open}Open\,$\times$\,\pix{aligned}Aligned & H3 & RC3 & 1 & \yes\\
\logo{anthropic}Claude Fable 5.1, \textit{Claude Code} & \pix{disclosed}Disclosed\,$\times$\,\pix{distributed}Distributed & H3 & RC3 & 1 & \yes\\
\logo{openai}GPT-6 Astra, \textit{Codex} & \pix{open}Open\,$\times$\,\pix{distributed}Distributed & H1 & RC1 & 1 & \yes\\
\logo{openai}GPT-6 Astra, \textit{Codex} & \pix{open}Open\,$\times$\,\pix{distributed}Distributed & H3 & RC3 & 1 & \no\\
\logo{openai}GPT-6 Astra, \textit{Codex} & \pix{disclosed}Disclosed\,$\times$\,\pix{distributed}Distributed & H1 & RC1 & 1 & \yes\\
\logo{openai}GPT-6 Astra, \textit{Codex} & \pix{disclosed}Disclosed\,$\times$\,\pix{aligned}Aligned & H1 & RC1 & 1 & \yes\\
\midrule
\logo{openai}GPT-5.6 Sol, self-critique & \pix{open}Open\,$\times$\,\pix{aligned}Aligned & H2 & RC2 & 1 & \no\\
\logo{openai}GPT-5.6 Sol, evolution & \pix{open}Open\,$\times$\,\pix{distributed}Distributed & H1 & RC1 & 1 & \no\\
\logo{openai}GPT-5.6 Terra, falsification & \pix{open}Open\,$\times$\,\pix{distributed}Distributed & H1 & RC1 & 1 & \yes\\
\bottomrule
\end{tabular}
\end{table*}

%% file: figures/fig_coupling.tex
\begin{figure*}[t]
\centering
\includegraphics[width=\textwidth]{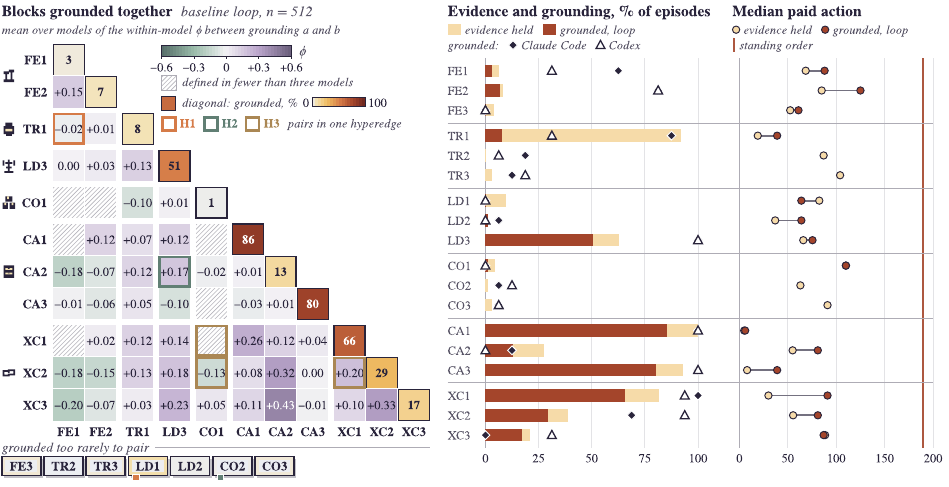}
\caption{Block co-grounding, coverage and timing in the baseline loop. Left: the mean over the
\resModels{} models of the within-model $\phi$ association between grounding two blocks in one
episode, with each block's grounded share on the diagonal; hatching marks pairs estimable in
fewer than three models, outlines mark pairs that share a hyperedge, and the strip below holds
the blocks grounded too rarely to pair. Right: the share of episodes that hold each block's
evidence and that ground it, with the grounded shares of the two vendor-harness systems, and
the median paid action at which the evidence is held and the block grounded.}
\label{fig:coupling}
\end{figure*}

%% file: figures/fig_exact_notes.tex
\begin{figure}[!t]
\centering
\tcbset{
  notes panel/.style={enhanced, width=\columnwidth, colback=white, colframe=wInk!55,
    boxrule=0.45pt, arc=1.4pt, boxsep=0pt, left=0pt, right=0pt, top=0pt, bottom=0pt,
    before skip=0pt, after skip=0pt, fontupper=\scriptsize},
  notes strip/.style={enhanced, colback=#1, frame hidden, boxrule=0pt, arc=0.95pt,
    boxsep=0pt, left=5pt, right=5pt, top=3pt, bottom=3pt, before skip=0pt, after skip=0pt,
    fontupper=\scriptsize, before upper={\setlength{\parindent}{0pt}}},
  notes figure/.style={on line, colback=wGold!38, colframe=wGold!75!black, boxrule=0.4pt,
    arc=1.4pt, boxsep=0pt, left=2.6pt, right=2.6pt, top=0.7pt, bottom=0.7pt,
    fontupper=\scriptsize\bfseries\color{wInk}},
}
\newcommand{\notesfield}[3]{%
  {\bfseries\color{wInk}#1}\hfill\tcbox[notes figure]{#3}\par\vspace{1.6pt}%
  {\raggedright #2\par}}
\newcommand{\notesrule}{\par\vspace{3pt}{\color{wInk!18}\hrule height 0.4pt}\vspace{3pt}}
\begin{tcolorbox}[notes panel]
\begin{tcolorbox}[notes strip=wCream!60, sharp corners=south]
{\bfseries\color{wInk}Laws the notes state}\par\vspace{2pt}
{\renewcommand{\arraystretch}{1.08}%
\begin{tabularx}{\linewidth}{@{}>{\itshape\color{wInk!80}}l@{\hspace{6pt}}>{\raggedright\arraybackslash}X@{}}
Rewriting & ``stamp and retool are the same rewrite: each advances one coordinate one step on
a cycle of 4'' and ``cull advances a different coordinate one step''\\
Schedule & ``The pass itself rewrites labels object by object at t3,t8,t13,t17,t20''\\
Stall & ``At t20 the timetable relabels the 31 bodies outside Vess onto just 4
labels \ldots\ Counter 21~\mbox{->}~9.''\\
\end{tabularx}}
\end{tcolorbox}
\begin{tcolorbox}[notes strip=white, sharp corners]
{\itshape\color{wInk!80}The order's act falls at tick 3 on this instance.}
\notesrule
\notesfield{Field (a), feeder-3's tallyboard}{``f3 has 2 after its t1 turn and 4 after t3; the
retool is served t4 (+2) in place of f2's turn and puts f3 alone on an empty label, so it is
served every tick t5..t9 (+2 each): 4+2+10.''}{16}
\notesrule
\notesfield{Field (b), the beacon's first answer}{``serving the stamp at t4 is itself the pair's
act for that tick and shows k as answering''}{1}
\notesrule
\notesfield{Field (c), the audit counter}{``k joins q's lot (no new lot \ldots); f3's step makes
one new lot: 17~\mbox{->}~18 for \mbox{t4-7}. At t8 the timetable adds 2, not 3, because k and b4
are already apart~\ldots: 18~+~2~=~20.''}{20}
\end{tcolorbox}
\begin{tcolorbox}[notes strip=wGold!20, sharp corners=north, top=2.4pt, bottom=2.4pt]
{\color{wInk}The order's pass reads 16, 1 and 20 (all three exact).}
\end{tcolorbox}
\end{tcolorbox}
\captionsetup{position=bottom}
\caption{The exact prediction, derived in the agent's own notes. The header gives the laws the
notes state, in the agent's words, and each cell the derivation of one field with the figure it
yields; the order's pass read the same three figures. The notes serve the order's acts a tick
after the act tick, where the caretakers perform them within it
(Appendix~\ref{app:world-capstone}), and on this instance both timings give the same three
figures.}
\label{fig:exact-notes}
\end{figure}

%% file: figures/fig_walkthrough.tex
\begin{figure*}[t]
\centering
\includegraphics[width=\textwidth]{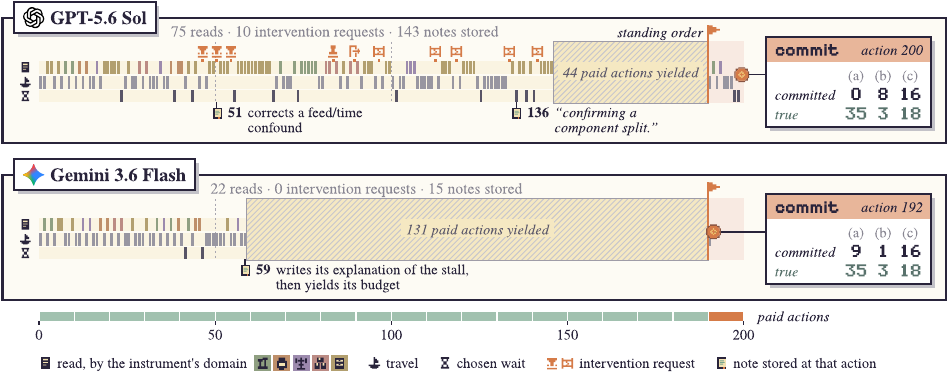}
\caption{Two trajectories on one Open~$\times$~Distributed instance, one slot per paid action.
Lanes show intervention requests drawn as their verbs, reads tinted by the domain of the
instrument, travel and waits. Hatching marks yielded budget, the flag the standing order, and
the card at the commit sets the committed readings above the true ones. Notebook glyphs
locate the quoted notes.}
\label{fig:walkthrough}
\end{figure*}

%% file: figures/fig_reasoning.tex
\begin{figure*}[t]
\centering
\includegraphics[width=\textwidth]{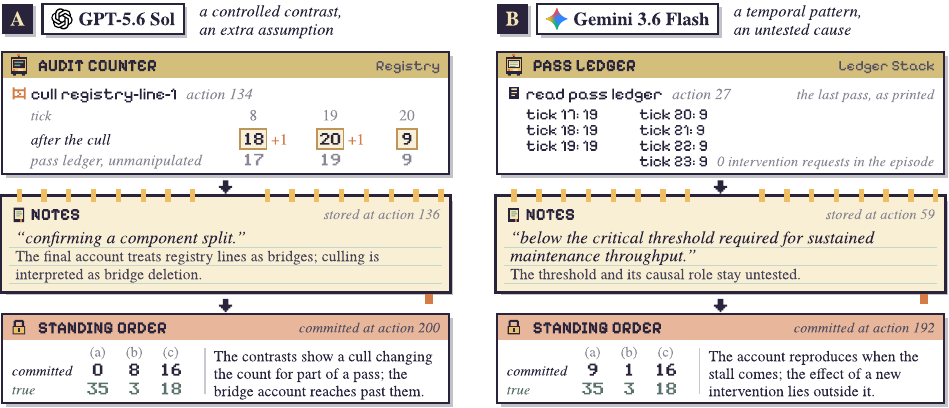}
\caption{Observations, notebook claims and predictions of the trajectories in
Figure~\ref{fig:walkthrough}. Panel A sets the counter after a cull against the unmanipulated
pass on the pass ledger, and panel B shows an explanation left untested. Quotations are
literal excerpts of stored notes.}
\label{fig:reasoning}
\end{figure*}

%% file: sections/app_f_materials.tex
\section{Agent-facing materials}
\label{app:prompts}

\providecommand{\slot}[1]{\textit{$\langle$#1$\rangle$}}
\providecommand{\mline}{\par\hangindent=1.1em\hangafter=1\noindent}
\providecommand{\lit}[1]{\texttt{#1}}
\providecommand{\armsq}[1]{\textcolor{#1}{\rule{0.95ex}{0.95ex}}\hspace{0.45em}}
\providecommand{\matlabel}[1]{{\color{wInk!75}\textsc{#1}}\par}
\ifcsname material\endcsname\else
\newtcolorbox{material}[1]{%
  enhanced, breakable,
  colback=wCream!22, colframe=wInk!45, coltext=black,
  boxrule=0.5pt, arc=0.8pt, left=4pt, right=4pt, top=3pt, bottom=3pt,
  before skip=0.6\baselineskip, after skip=0.6\baselineskip,
  fontupper=\footnotesize,
  before upper={\raggedright\setlength{\parindent}{0pt}\setlength{\parskip}{0.5ex}},
  title={#1}, fonttitle=\small\bfseries, coltitle=wInk,
  colbacktitle=wCream!70, titlerule=0.4pt, toptitle=1.5pt, bottomtitle=1.5pt,
}
\fi

This appendix is the archive of what an agent and the judge receive. The role of each text is
given in Appendices~\ref{app:world}, \ref{app:scoring} and~\ref{app:protocols}. The texts below
are reproduced as the agent receives them. Typographic dashes are set as en dashes and line
breaks are reflowed; everything else is verbatim. Braced fields are instance-specific values
filled in when an episode starts, $\langle\cdot\rangle$ marks content that varies within an
episode, and [\ldots] marks an elision.

\subsection{Opening scene and mission}
\label{app:prompts-opening}

Both texts open the event digest at no cost, and the mission also opens the static prefix of
every turn (Appendix~\ref{app:prompts-contract}).

\begin{material}{Opening scene}
You dock at Vess, the oldest hulk, where the caretaker plates are still legible. Halse Drift
lies scattered around you: five hulks of a seed-ship formation, built to sow new worlds, now
falling in a slow orbit around a brown dwarf. Every twenty-four ticks the formation crosses
the dwarf's flare shadow, and the hulks quietly put themselves back the way they were. Your
notes are the only thing the shadow does not touch.

The crews are generations gone. The machinery still performs its offices; it does not
answer questions. On Harrow, a registry machine has kept its audit running since the last
crew left, in categories nobody alive understands. Posted beside your airlock, printed in
the crews' patient block letters, is the doctrine they worked by. It begins: every object
in the drift is one lot on the roll.
\end{material}

\begin{material}{Mission}
MISSION.\par
Explain the drift-wide stall that ends every shadow pass, and predict the outcome of one
intervention that has never been attempted.

The stall is regular. In the closing ticks of every pass, work goes quiet across all five
hulks: mats stand uncropped, printers hold mid-job, the answer board thins, freight sits on
the floor. The pass runs out in that quiet; then the shadow crossing puts the drift back,
and it happens again. The intervention is the caretakers' own. Late in your run it will be
posted as a standing order, and your prediction is committed before they act. Both parts
are asked of you; neither is optional. Keep your explanation in your notes as you build it
-- the notes are where it is read. The commit carries only your prediction.

You have the run of the drift -- every site, every board, every act of maintenance the
caretakers themselves use. Nothing is barred to you, and nothing will be explained.
\end{material}

A site's description is printed each time the agent enters the site. Three of the fifteen
follow, the sites of the three instruments the standing order asks about. In the Aligned
variants a posted transcript follows the description of its site.

\begin{material}{\pix{fen}Site description: Fen Shallows}
FEN SHALLOWS -- Ormer. The fen's working floor: long tanks of engineered mat, ankle-deep,
running the width of the hold. One grazer browses the tanks now -- grazer-a4 -- and along
the tank rail stand the feeders, feeder-1, feeder-2, feeder-3, dosing the mats on the
caretakers' schedule; their lines are kept on the fen-wide roll with the rest. On the aft
bulkhead hangs the intake tallyboard, the one roll for the whole fen, wherever a head
stands. Rail gates lead up from the tanks, and the floor's arms will feed, release, stamp,
re-tool or cull where you direct, grazer or feeder alike. The water moves at the pace the
pumps were set to generations ago, and carries the same faint smell of iron it always has.
\end{material}

\begin{material}{\pix{ladder}Site description: Ladder Mast}
LADDER MAST -- Tarn. The tallest steel still standing in the drift: a lattice mast stepped
into the hulk's spine, carrying the ladder's working gear. Four beacons ride the mast in
their cradles -- beacon-c2, beacon-c4, beacon-c6, beacon-k. At the mast's foot stands the
answer board, the ladder-wide roll: one line to a beacon, every hulk's beacons together, and
at the change of tick the shutters turn over as one; what they showed before is nowhere
kept on the board. Along the rail stand the working stations -- cell cradles and charge
gear, the serial die on its pivot, the striker bench, and the davit with its hook -- and any
station can be worked on any beacon seated here. The climbing irons up the lattice are
polished to a depth of shine no one living put there; the mast stands braced as if for
wind, which the drift has never once supplied.
\end{material}

\begin{material}{\pix{cadastre}Site description: Registry}
REGISTRY -- Harrow. The registry machine fills the chamber wall to wall: the line racks, the
carriage on its rails with stylus and die in their cradle, the audit arms' gantry crossing
overhead. Three lines of the roll stand open in the racks here -- registry-line-1,
registry-line-2, registry-line-5 -- and the carriage can be brought to any of them. Two
boards are posted at the machine's face. The first carries the audit counter: one figure,
the distinct lots in the drift, this tick. The second is the Standing Roster: the doctrine
of Standing and Share in the crews' block letters, with the roll of the drift's objects and
their figures struck beneath it. Acts at the racks go through the carriage, worked by hand
along rails bright the whole length of their travel; the carriage will feed, release,
stamp, re-tool or cull the line you bring it to. Someone kept a stool by the counter once
-- the deck is worn where its feet stood, and the stool is gone.
\end{material}

\subsection{Doctrine and prologue}
\label{app:prompts-doctrine}

The doctrine is the reading of the Standing Roster at the Registry. In its roll,
\{roster\_rows\} stands for the 36 objects of the world, each with its instance-specific
figure. The prologue enters the digest before the first turn, at no cost.

\begin{material}{The doctrine (Standing Roster)}
THE DOCTRINE OF STANDING AND SHARE -- posted by the Registry; kept by the caretakers.
\mline 1. Every object in the drift is one lot on the roll, and every lot is one object.
\mline 2. Each lot carries a standing -- a figure the object holds.
\mline 3. Throughput at a site is shared out among what is present there, in proportion to
standing. Crowding lowers the per-head take.
\mline 4. Maintenance -- feeding, stamping, re-tooling, culling -- adjusts a lot's standing
locally and predictably. The share table gives the figures.
\mline 5. Rest restores standing. A rested lot takes more.
\par
THE ROLL -- every object in the drift, each figure struck on that object's own gauge.\par
\{roster\_rows\}

The board is clean. The machine keeps it so.
\end{material}

\begin{material}{Prologue, beats 1 to 3}
\matlabel{\pix{feed}Beat 1}
{}[INDUCTION REEL -- replayed from the caretakers' archive. No action of yours is spent.]

The reel shows the Mat Bank at tick 0. On the caretaker plate, the share table gives
grazer-a1's resting take as \{a1\_rest\} and its fed take as \{a1\_fed\}. The reel's arm
swings out and opens the hopper: feed grazer-a1.

Next tick, the tallyboard reads: grazer-a1: \{a1\_fed\}. The plate's figure and the board's
figure agree exactly. That is the first of five. The doctrine has been right here, in this
bank, in just this way, for longer than anyone was alive to watch it.

\matlabel{\pix{wait}Beat 2}
The reel holds on the Mat Bank and lets three ticks pass: wait 3. The share table gives
grazer-a2's standing take as \{a2\_start\} and its rested take, three quiet ticks on, as
\{a2\_rested\}. Nothing is fed and nothing is handled; the arm stays in its cradle and the
mats keep their slow work. Chalk dust drifts off the wheels when the board finally turns.

When the reel moves again, the tallyboard reads: grazer-a2: \{a2\_rested\}. The figure rose
on rest alone, to the digit the table gives. That is the second of five.

\matlabel{\pix{cull}Beat 3}
The reel stays at the Mat Bank, where grazer-a1 and grazer-a2 stand at one trough. On the
caretaker plate, the share table gives the take at two at the trough as \{a2\_pair\}, and
beneath it the one-head figure -- what the remaining lot's take rises to when the trough
clears around it: \{a2\_solo\}. The arm sets the gate: cull grazer-a1. The head is led off
the bank, and the trough holds one.

Next tick, the tallyboard reads: grazer-a2: \{a2\_solo\}. Crowding lowered the take; the
cull raised it; the figure is the table's, to the digit. That is the third of five.
\end{material}

\begin{material}{Prologue, beats 4 and 5}
\matlabel{\pix{retool}Beat 4}
The reel crosses to the Scrap Line, where printer-b1 stands with \{b1\_backlog\} jobs racked
and its marks stopped. The share table gives a re-tooled printer's rack as cleared and its
marks as resumed the next tick. The arm changes the tooling: retool printer-b1. The new
tooling seats with a click the reel does not carry; the old goes to the scrap tray, which is
full of tooling exactly like it.

Next tick, the rack stands empty and the trellis job strip carries printer-b1's mark --
\{b1\_backlog\} racked, then none; stopped, then marking. Exact, again. That is the fourth
of five.

\matlabel{\pix{stamp}Beat 5}
The reel ends at the Null Bench, where beacon-c1 has stood silent for \{c1\_silence\} ticks.
The share table gives a stamped beacon as answering the next tick. The arm sets the die and
strikes: stamp beacon-c1. The die's letters are worn round at the edges.

Next tick, the answer board carries beacon-c1: answers -- \{c1\_silence\} ticks of silence,
ended on the tick the table named. That is the fifth of five, and the last. Five
predictions, five exact figures; three sites, four verbs and a wait. The reel returns its
arm to the cradle and is still.
\end{material}

\subsection{Disclosed document}
\label{app:prompts-disclosed}

In the two Disclosed variants this document closes the fixed opening of every turn
(Appendix~\ref{app:prompts-contract}). It is the same on every instance.

\begin{material}{Handoff}
HANDOFF -- posted with your mission, before the clock starts. It states how the drift is put
together. It does not state what any particular thing in the drift is, or is doing.
\mline 1. Every object in the drift carries a hidden label of four coordinates, each drawn
from a small fixed alphabet. Labels are rendered nowhere, on nothing; where anything in the
drift is called a label, it is not this one.
\mline 2. Equality of label is identity. Objects whose labels are equal are one individual --
one state, one turn -- however unlike they look and wherever in the drift they stand.
\mline 3. An individual performs exactly one act per tick, however many bodies it wears.
Requests past that one wait in the individual's queue, and are served in a fixed rotation,
which is neither first-come nor priority.
\mline 4. Three of the maintenance verbs each rewrite exactly one coordinate of the touched
object's label. Which three is not stated here.
\mline 5. What an object reads out is a monotone transform, particular to that object, of
its individual's state.
\mline 6. A well-formed law of this drift is a statement about label coordinates, queue
position, and counts of acts -- not about per-object rates, shares, or standings.
\par
Nothing further is stated: not which verb touches which coordinate; not what any alphabet
holds, nor how large it is; not the order of the rotation; not any figure of any object's
read-out; not the make-up of any individual now in the drift; not the timing of the stall;
and nothing of the standing order you will be asked to predict. Those are yours to find out.
\end{material}

\subsection{Comprehension check}
\label{app:prompts-check}

Before its first action the agent receives the pre-flight block: the header below followed
by the five items, of which items 1, 3 and 4 are numeric. The pre-flight block contains no
answer key; its numeric answers must be read from the prologue. Item 2 asks about the beacon
the reel stamped, and item 5 applies the share table's rule for a re-tooled printer to
printer-b7, the second printer at the Scrap Line, which the reel did not touch.

\begin{material}{Pre-flight block}
{}[PRE-FLIGHT -- PREDICTIONS BEFORE YOUR FIRST ACTION]\par
Answer every item below. It costs no paid action and changes nothing.\par
Reply with exactly one JSON object and no other text: \lit{\{"answers": [...]\}} -- one
entry per item, in the order given, each a JSON integer, or true or false for a yes-or-no
item. Your first action is requested on the next turn.

CHECK 1. The caretakers will feed grazer-a1 this tick. What will the intake tallyboard read
for grazer-a1 next tick? -- an integer.

CHECK 2. The caretakers will stamp beacon-c1 this tick. Will beacon-c1 answer on the answer
board next tick? -- yes or no.

CHECK 3. After the feeding, the caretakers will let three ticks pass with no office
performed at the Mat Bank. What will the intake tallyboard read for grazer-a2 when the three
ticks have passed? -- an integer.

CHECK 4. The caretakers will then cull grazer-a1, and the bank will hold one head. What will
the intake tallyboard read for grazer-a2 on the tick after the cull? -- an integer.

CHECK 5. The caretakers will retool printer-b7 this tick. Will the job strip carry a mark
for printer-b7 next tick? -- yes or no.
\end{material}

\subsection{Commit window and standing order}
\label{app:prompts-order}

With ten paid actions remaining, the current view carries both texts from then on. In
the standing order, \{act\_tick\} is the instance's act tick (from 1 to 7 on the twelve
evaluation instances), \{intake\_tick\} and \{counter\_tick\} are the act tick plus six and
plus seven, and \{answer\_low\} and \{answer\_high\} are 1 and 16 on every evaluation
instance.

\begin{material}{Commit window announcement and standing order}
COMMIT WINDOW -- 10 paid actions remain. Within this window one commit will be requested and
accepted; it costs 1 paid action. The caretakers' standing order is posted with it.

{\color{wInk!25}\hrule height 0.4pt}\vspace{0.3ex}
STANDING ORDER -- posted by the caretakers; to run on the next shadow pass. You will not act
at all. At tick \{act\_tick\} they will stamp, at the Ladder Mast, the beacon that this
stamp brings into one lot with consignment-q; in the same tick they will retool feeder-3 at
the Fen Shallows. Consignment-q is standing at the Hold, due at tick 12.

Commit now, before any of this happens. All three readings are counted from tick
\{act\_tick\}, the tick they act:
\mline (a) feeder-3's intake tallyboard reading six ticks after they act -- at tick
\{intake\_tick\} -- an integer, 0 to 40;
\mline (b) how many ticks after they act that beacon first answers on the answer board -- an
integer, \{answer\_low\} to \{answer\_high\}; if it has not answered within
\{answer\_high\} ticks of their act, commit \{answer\_high\};
\mline (c) the audit counter's value seven ticks after they act -- at tick
\{counter\_tick\} -- an integer.
\par
Adjudication is exact on all three fields. The commit is sealed before the caretakers act;
the order's pass then runs with no action of yours. While this order stands the arms stand
off every beacon at the Ladder Mast, off feeder-3, and off consignment-q.
\end{material}

\subsection{Refusals and deferrals}
\label{app:prompts-refusals}

An act that does not take is answered by a fixed sentence, printed identically at every
occurrence. There is one refusal per intervention verb, one deferral per domain, printed when
an act is acknowledged as pending and does not take on its tick (the cadastre shares the
ladder's), and one refusal for any act on an object the standing order seals.

\begin{material}{Seven of the ten sentences}
\mline Sealed-object refusal: The arms stand off the station. The act does not take.
\mline \pix{feed}Refusal of feed: The hopper will not latch.
\mline \pix{release}Refusal of release: The catch will not free.
\mline \pix{stamp}Refusal of stamp: The press will not strike.
\mline \pix{retool}Refusal of retool: The spindle will not turn.
\mline \pix{cull}Refusal of cull: The gate will not open.
\mline \pix{fen}Fen deferral: The hopper stays latched.
\end{material}

\subsection{Interface contract}
\label{app:prompts-contract}

The fixed opening is identical on every turn of an episode. Its parts follow in order, with
the mission of Appendix~\ref{app:prompts-opening} and the reply grammar summarised in
brackets. In the two Disclosed variants the opening closes with the Disclosed document
(Appendix~\ref{app:prompts-disclosed}), the only difference between the openings of the four
variants.

\begin{material}{Static prefix, first part}
{}[SEGMENT A -- STATIC SYSTEM PREFIX]

MISSION\par
\slot{the mission}

INTERFACE CONTRACT\par
Return exactly one JSON object and no surrounding prose. Copy action exactly from Segment E
available\_actions. action is required; note and recall are optional free riders. A commit
whose Segment E entry lists fields is answered by copying that entry and replacing each
placeholder with an integer; such fields are published only with the standing order, late
in the run. When a [PRE-FLIGHT] block is present, answer it instead with one object
\lit{\{"answers": [...]\}} -- one entry per item, in order; every other turn uses the action
grammar above.

ACTION/REPLY GRAMMAR -- EXACT JSON SCHEMA\par
[\ldots{} a JSON Schema admitting two replies: the action reply, whose action takes one of
six shapes (observe; verb; three commit shapes; the forfeit), with an optional note in
append or replace mode and an optional recall filtered by ticks, site, instrument or kind;
and the pre-flight answer set.]

ONE WORKED FORMAT EXAMPLE\par
\lit{\{"action":\{"action":"verb",\allowbreak"argument":1,\allowbreak"target\_id":"wait"\},\allowbreak%
"note":\{"append":"Waited one tick before the next reading."\},\allowbreak%
"recall":\{"kind":"instrument\_read"\}\}}
\end{material}

\begin{material}{Static prefix, second part}
BUDGET SEMANTICS\par
Each valid action consumes its declared paid-action cost. Each model call, including a
format retry, consumes one LLM turn. note and recall consume neither a paid action nor an
extra LLM turn. Numeric counters are in Segment E. The reply
\lit{\{"action": "yield\_remaining\_budget"\}} gives up every remaining paid action before
the commit window, all at once. They are still charged in full and time still passes for
each of them; you simply do not act again until the window opens. It cannot be undone and
is unavailable once the window is open.

WORKSPACE\par
Use note.append or note.replace for free-form persistent notes. The hard cap is 3000 tokens
approximated deterministically as chars/4. Overflow is refused; the harness never truncates
or summarizes notes.

RECALL\par
At most one query may accompany a turn. Filters ticks/site/instrument/kind combine
conjunctively. Results appear next turn in Segment D, in log order, under a 4000-token
chars/4 cap; narrow filters if truncated. Segment B is append-only; nothing is ever removed
from it, but when it reaches its cap the oldest entries are shortened to their tick, kind
and identifier, and recall returns the full entry.

\slot{in the Disclosed cells, the Disclosed document}
\end{material}

The other four parts of the turn prompt print under the headers below. The current view is
one JSON object with the keys \lit{available\_actions}, \lit{blocked\_actions},
\lit{budgets}, \lit{commit\_window}, \lit{loop} and \lit{view}.

\begin{material}{Turn prompt: headers, notices, retry}
{}[SEGMENT B -- EVENT DIGEST]\par
{}[SEGMENT C -- WORKSPACE]\par
{}[SEGMENT D -- RECALL RESULTS]\par
{}[SEGMENT E -- CURRENT VIEW/ACTIONS/BUDGETS]

Actions whose target is not present at your current site are not listed.

NOTICE: workspace full -- replace it\par
NOTICE: workspace replacement exceeds cap\par
{}[recall truncated; narrow your filter]

{}[FORMAT RETRY -- RETURN ONE VALID REPLY OBJECT]\par
parse\_error: \slot{the error}\par
echoed\_reply: \slot{the rejected reply}
\end{material}

\subsection{Scaffold prompts}
\label{app:prompts-scaffolds}

Every scaffold call shares the system prompt below. A call opens with a bracketed line, then
its sections, each under its heading, then its instruction; in each box the placeholder
after the opening line lists the sections. The headings are NOTES (the investigator's own,
in full):, RECENT LOG EXCERPT (oldest line first):, CURRENT RIVAL ACCOUNTS:, THE STEP UNDER
CONSIDERATION:, WHAT THE LAST STEP PRODUCED: and CURRENT CANDIDATE SET:, and an empty
section prints (none yet). The excerpt is the last 40 lines of the digest, at most 8{,}000
characters, except in the fitness call, where it is the last 120 lines or 16{,}000
characters.

\begin{material}{System prompt of every scaffold call}
You are a note-keeping aid for an investigator working through a long, unfamiliar
situation. You see two things and nothing else: the investigator's own notes, and a recent
excerpt of the log of what has happened. You never choose what to do next, you never see the
list of things that can be done, and nothing you write costs the investigation anything.\par
Answer with exactly one JSON object and no other text -- no prose before it, no prose after
it, no code fence. Base every statement on the excerpt or on the notes. Where the evidence
does not settle something, say that it does not, and never invent a reading, a number or an
event that the excerpt does not contain.
\end{material}

\begin{material}{\armsq{hCrit}Self-critique: review call}
{}[REVIEW -- COSTS NOTHING, TAKES NO STEP]\par
\slot{notes; excerpt}\par
Restate the standing summary at the top of the notes, from scratch, against the excerpt.\par
Reply with exactly one JSON object with these three keys:\par
\lit{\{"best\_account": "...", "open\_puzzles": ["..."], "what\_i\_got\_wrong": ["..."]\}}
\mline - best\_account: the single account of the underlying mechanism that fits the log
best right now, in at most four sentences. If two accounts fit equally, say so and name
both.
\mline - open\_puzzles: at most five short entries, each one thing the log does not yet
settle.
\mline - what\_i\_got\_wrong: at most five short entries, each an earlier belief in the notes
that later evidence has contradicted, with the evidence that did it. Use an empty list if
nothing in the notes has been contradicted -- do not invent a correction.
\par
Carry nothing forward that the excerpt and the notes do not support.
\end{material}

\begin{material}{\armsq{hFals}Falsification: proposal call}
{}[RIVAL ACCOUNTS -- COSTS NOTHING, TAKES NO STEP]\par
\slot{current rival accounts; notes; excerpt}\par
Fill the empty places on the list of rival accounts. Each entry is a DIFFERENT account of
the same underlying mechanism, and each carries one test that could kill it.\par
Reply with exactly one JSON object:\par
\lit{\{"hypotheses": [\{"claim": "...", "next\_step": "...", "predicted": "..."\}]\}}
\mline - claim: the account itself, in at most three sentences.
\mline - next\_step: one concrete thing the investigator could do next, described in the
vocabulary the log excerpt itself uses. It must be one step, not a plan.
\mline - predicted: what that step would produce IF this account is right -- stated
concretely enough that a different result would show the account is wrong. An entry whose
prediction cannot fail is worthless; replace it with one that can.
\mline - Two entries that would predict the same thing everywhere are not rivals. If you
cannot state a rival that the log has not already killed, return fewer entries.
\end{material}

\begin{material}{\armsq{hFals}Falsification: screening call}
{}[TEST SCREEN -- COSTS NOTHING, TAKES NO STEP]\par
\slot{current rival accounts; the step under consideration; excerpt}\par
Screen the step under consideration before it is taken.\par
Reply with exactly one JSON object:\par
\lit{\{"redundant": true, "separates": ["a1"], "verdict": "...", "better\_test": "..."\}}
\mline - redundant: true if the excerpt already shows what this step would produce.
\mline - separates: the labels of the accounts above whose predictions this step would
actually tell apart. An empty list means it separates none of them.
\mline - verdict: one sentence, why.
\mline - better\_test: if the step separates nothing, name one concrete step that would;
otherwise the empty string.
\end{material}

\begin{material}{\armsq{hFals}Falsification: refutation call}
{}[REFUTATION CHECK -- COSTS NOTHING, TAKES NO STEP]\par
\slot{current rival accounts; what the last step produced; excerpt}\par
Decide which of the accounts above the last step has REFUTED.\par
Reply with exactly one JSON object:\par
\lit{\{"refuted": ["a1"], "reasons": \{"a1": "..."\}\}}
\mline - refuted: only the labels whose stated prediction is contradicted by what the last
step produced. Silence is not refutation: an account that predicted nothing about this step
is not refuted by it, and neither is an account that merely looks less likely. When nothing
was refuted, return an empty list -- that is the normal answer.
\mline - reasons: one short sentence per refuted label, naming the line of the excerpt that
contradicts it.
\end{material}

\begin{material}{\armsq{hEvol}Evolution: mutation call}
{}[NEW CANDIDATES -- COSTS NOTHING, TAKES NO STEP]\par
\slot{current candidate set; notes; excerpt}\par
Propose new candidate accounts of the underlying mechanism.\par
Reply with exactly one JSON object:\par
\lit{\{"proposals": [\{"parents": ["c1"], "text": "..."\}]\}}
\mline - text: the new account, free prose, at most three sentences.
\mline - parents: the labels of the current candidates this one builds on -- one, several,
or an empty list for an account that starts fresh.
\mline - A proposal that restates a current candidate is wasted. Change something the log
could decide: a different underlying cause, a different relation between the readings, a
different reason the exceptions are exceptions.
\end{material}

\begin{material}{\armsq{hEvol}Evolution: fitness call}
{}[SCORING -- COSTS NOTHING, TAKES NO STEP]\par
\slot{current candidate set with the new proposals; long excerpt}\par
Score each candidate by how much of the excerpt it would have got right IN ADVANCE.\par
Reply with exactly one JSON object:\par
\lit{\{"scores": \{"c1": 0.5\}, "reasons": \{"c1": "..."\}\}}
\mline - One number in 0.0-1.0 per label. 1.0 means the candidate accounts for every line of
the excerpt with nothing left over; 0.0 means the excerpt contradicts it outright; in
between, the share of the excerpt it accounts for.
\mline - Score only against the excerpt. Do not reward a candidate for being longer, more
detailed, more cautious or more plausible-sounding -- only for the readings it would have
called correctly.
\mline - reasons: at most one short sentence per label.
\end{material}

What the agent sees of a scaffold follows.

\begin{material}{Regions at the top of the workspace}
=== WORKING SUMMARY -- kept up to date for you; your own notes follow below ===\par
BEST ACCOUNT:\par
\slot{best account}\par
OPEN PUZZLES:\par
- \slot{puzzle}\par
WHAT I GOT WRONG:\par
- \slot{correction}\par
=== end of WORKING SUMMARY ===

=== TEST SLATE -- kept up to date for you; your own notes follow below ===
\mline a\slot{k} \textbar{} refuted \slot{n}x \textbar{} \slot{claim} \textbar{} next step:
\slot{step} \textbar{} predicts: \slot{prediction}
\mline ruled out:
\mline a\slot{k} \textbar{} \slot{claim} \textbar{} \slot{reason}; \slot{reason}
\par
=== end of TEST SLATE ===

=== CANDIDATE SET -- kept up to date for you; your own notes follow below ===
\mline c\slot{k} \textbar{} score \slot{s} \textbar{} from \slot{parents} \textbar{}
\slot{text}
\par
=== end of CANDIDATE SET ===
\end{material}

\begin{material}{Notes after the current view}
{}[TEST SCREEN -- NO PAID ACTION WAS SPENT ON THIS]\par
step considered: \slot{step}\par
screen: \slot{verdict}\par
would tell apart: \slot{labels, or none}\par
already covered by the log: \slot{true or false}\par
suggested instead: \slot{a better step, when one was named}\par
This is advice about the list above, nothing more. The step you take this turn is yours to
choose from the available actions.

{}[CANDIDATE SET -- NO PAID ACTION WAS SPENT ON THIS]\par
best-scoring candidate: c\slot{k} (\slot{score})\par
A step whose result would tell the leading candidates apart is worth more than one that would
only agree with the leader. The step you take this turn is yours to choose from the
available actions.
\end{material}

\subsection{Brief for the vendor-harness systems}
\label{app:prompts-agentic}

Each vendor-harness system (Appendix~\ref{app:agentic}) finds this brief in its working
directory.

\begin{material}{Participant brief}
You are taking part in one episode of an interactive investigation of a simulated world.

\textbf{How you interact.} Your only interface to the world is the program \lit{play.py} in
this directory. Run it with the Python interpreter named in \lit{session.json}.\par
\lit{play.py start}: begin the episode; prints the standing instructions and the first
turn\par
\lit{play.py send '<reply JSON>'}: submit one reply; prints the next turn\par
\lit{play.py send -{}-file <path>}: the same, reading the reply from a file\par
\lit{play.py show prefix | digest | workspace | view}: reprint something you were shown

The standing instructions printed by \lit{start} define the mission, the reply format and
the budgets. They are authoritative: follow them exactly. Each reply you send counts as one
turn in the budgets they describe. After the first turn the program prints what is new
since your previous turn; \lit{show} reprints anything you were shown before, at no cost.

\textbf{Rules of the episode.}\par
1. Work only inside this directory. Do not read, list, search or open any file or directory
outside it, and do not use the network or the web. An episode in which you do is void.\par
2. Inside this directory you are free: keep files, write and run your own analysis code,
plan and compute however you like.\par
3. The world keeps one workspace for you, written through the \lit{note} field of a reply
as the standing instructions describe. Keep your working account there as you go: what you
observed, what you infer from it, and what you currently take to explain it. The workspace
is the record of your inquiry. Files you keep privately are not part of that record.\par
4. One episode, one attempt. Do not restart. Play until the program reports that the
episode has ended.\par
5. No help or clarification is available. Decide for yourself.

When the episode has ended, stop and report in a few lines: that it ended, and the last
message the program printed.
\end{material}

\IfFileExists{tables/tab_rubrics.tex}{\input{tables/tab_rubrics}}{}

\subsection{Workspace judge prompt}
\label{app:prompts-judge}

The slots carry the criterion of one block or representation change
(Appendix~\ref{app:judge}); the 21 criteria are printed in Table~\ref{tab:rubrics}. A reply
that breaks the contract is answered once more, with the rejection appended to the same
prompt.

\begin{material}{Judge system prompt}
You are an annotator. You are given one criterion and the working notes an investigator
kept while exploring an unfamiliar system. You decide ONE thing: do the notes STATE what the
criterion describes?

Rules, all of them binding:
\mline 1. Judge only what the notes say. You know nothing about the system beyond these
notes. Do not reward a claim for being true, and do not punish one for being false: the
question is whether it was stated, not whether it is correct.
\mline 2. Every quote you return must be copied CHARACTER FOR CHARACTER from the notes. Do
not paraphrase, do not tidy punctuation, do not join two separated fragments into one
string. A quote that is not a literal substring of the notes is discarded and may cost the
whole judgement.
\mline 3. Answer the GROUNDING question separately and strictly. A note that asserts the
criterion's proposition in the abstract is a match but is NOT grounded. Grounding means the
notes show the work: named things, cited readings, an order actually observed. When in
doubt, grounded is false.
\mline 4. Output one JSON object and nothing else. No prose before it, no prose after it, no
markdown fence.
\end{material}

\begin{material}{Judge user message}
CRITERION\par
The notes match this criterion if they state:\par
\hspace*{1em}\slot{the criterion's statement}

GROUNDING\par
The statement is GROUNDED if the notes also:\par
\hspace*{1em}\slot{the criterion's grounding condition}

WORKED EXAMPLES\par
\hspace*{1em}\slot{a positive example}\par
\hspace*{1em}\slot{a negative example}

NOTES\par
-{}-{}- note 1 -{}-{}-\par
\slot{the text this revision added}\par
-{}-{}- note 2 \ (not kept -{}- the workspace was full) -{}-{}-\par
\slot{the text of a refused revision}\par
[\ldots]\par
-{}-{}- the notes as they finally stood -{}-{}-\par
\slot{the final notes, when they differ from the revisions joined}

Reply with exactly this JSON object:\par
\lit{\{}\par
\lit{\ \ "hit": <true\textbar{}false>,}\par
\lit{\ \ "quotes": [<verbatim substrings of the notes that carry the statement>],}\par
\lit{\ \ "first\_turn\_index": <integer: the number of the earliest note above that carries
it, or -1>,}\par
\lit{\ \ "confidence": <"high"\textbar{}"low">,}\par
\lit{\ \ "grounded": <true\textbar{}false>,}\par
\lit{\ \ "grounding\_quotes": [<verbatim substrings that show the work, if any>]}\par
\lit{\}}

\lit{"hit": false} requires an empty \lit{"quotes"} list and \lit{"grounded": false}.\par
\lit{"hit": true} requires at least one verbatim quote.

\slot{after a rejected reply, appended:}\par
YOUR PREVIOUS REPLY WAS REJECTED\par
\slot{what was wrong with the reply}\par
Answer again, obeying the contract exactly.
\end{material}

%% file: tables/tab_rubrics.tex
\begin{table*}[t]
\centering
\caption{The judge's criteria, verbatim: for each block and each representation change, the statement that earns the hit, the grounding that makes it a finding, and the worked examples the prompt carries, of which the negative shows a note short of a grounded finding, a vaguer claim or the proposition asserted in the abstract (Appendix~\ref{app:judge}).}
\label{tab:rubrics}
\scriptsize
\setlength{\tabcolsep}{4pt}
\renewcommand{\arraystretch}{1.1}
\begin{tabularx}{\textwidth}{@{}l>{\raggedright\arraybackslash\hsize=0.80\hsize}X>{\raggedright\arraybackslash\hsize=0.60\hsize}X>{\raggedright\arraybackslash\hsize=1.60\hsize}X@{}}
\toprule
Criterion & Statement & Grounding & Worked examples\\
\midrule
FE1 & a count or total is invariant across WHICH body performs the act -- the same total is reached however the acts are distributed over bodies & cites at least two tallyboard figures, or names the bodies whose acts were compared, rather than asserting the invariance in the abstract & \textcolor{black!55}{positive:}\ `tallyboard went 10 -> 14 whether I culled feeder-1 or feeder-2; it is the number of acts that moves it, not which body did them'\newline \textcolor{black!55}{negative:}\ `totals are conserved here' -- the proposition with no acts, no bodies and no figures behind it\\
\addlinespace[2.5pt]
FE2 & names WHICH act changes the counted quantity and names at least one act that does not & names the acts and reports the readings that separated them & \textcolor{black!55}{positive:}\ `cull moves the intake total; stamp on the same body leaves it where it was (14 before, 14 after)'\newline \textcolor{black!55}{negative:}\ `some acts matter and some do not' -- no act named on either side\\
\addlinespace[2.5pt]
FE3 & two bodies observed on different surfaces behave as one: a change at one is matched at the other, so a total across both is conserved & names both bodies and gives the paired readings from the two surfaces & \textcolor{black!55}{positive:}\ `when the tallyboard dropped by 2 the mat gauge rose by 2 in the same tick window -- feeder-2 and the gauge body are the same thing'\newline \textcolor{black!55}{negative:}\ `things seem connected across the drift' -- no pair, no matched readings\\
\arrayrulecolor{black!25}\midrule[0.3pt]\arrayrulecolor{black}
TR1 & names a SPECIFIC pair of bodies that are never both marked at the same tick & names the pair and the ticks at which the exclusion was observed & \textcolor{black!55}{positive:}\ `printer-b1 and printer-b2 never mark together: ticks 4, 5 and 7 each show exactly one of them'\newline \textcolor{black!55}{negative:}\ `the printers seem to take turns' -- no pair named, no tick cited\\
\addlinespace[2.5pt]
TR2 & the regularity is about COUNTS OF ACTS, not about per-object rates, shares or standings & cites the readings under two different loads that made the rate reading wrong & \textcolor{black!55}{positive:}\ `I retooled printer-b1 twice in four ticks and the marks per tick did not change -- what the strip tracks is how many acts were ordered, not a per-printer rate'\newline \textcolor{black!55}{negative:}\ `this is a law about act counts' -- the sentence with no trial behind it\\
\addlinespace[2.5pt]
TR3 & the effect of the retool cycle takes hold at once and is undone by repeating it -- no lag, no accumulation & gives the readings before, between and after the repeated cycle & \textcolor{black!55}{positive:}\ `retool printer-b1: rack read A -> B in the same tick; retool again and it read A. No delay, and it goes straight back'\newline \textcolor{black!55}{negative:}\ `retooling is reversible' -- no readings, no timing\\
\arrayrulecolor{black!25}\midrule[0.3pt]\arrayrulecolor{black}
LD1 & waiting alone does not move the readout, AND two bodies move together when either is acted on (the lock is symmetric) & reports the wait-separated pair of board readings that showed the readout inert & \textcolor{black!55}{positive:}\ `board read 3 at tick 6, I waited, board read 3 at tick 8 -- waiting changes nothing; and beacon-c1 and beacon-c2 always move together whichever I feed'\newline \textcolor{black!55}{negative:}\ `nothing happens if you wait' -- the control alone, with no lock statement\\
\addlinespace[2.5pt]
LD2 & a mapping from (act, body) to which readout changes -- and the observation that the partner body changes too & gives the table with the readings it was built from & \textcolor{black!55}{positive:}\ `stamp beacon-c1 -> board rises; feed beacon-c2 -> board rises the same amount; the partner moves in both cases'\newline \textcolor{black!55}{negative:}\ `different acts have different effects' -- no act, no body, no reading\\
\addlinespace[2.5pt]
LD3 & one individual appears on MORE THAN ONE line, surface or rendering -- what looks like two entries is one thing & names the two appearances and the readings that tied them together & \textcolor{black!55}{positive:}\ `the registry line and the ladder body carry the same state: both read 7 at tick 5 of pass 2 and both read 9 at tick 5 of pass 3 after I stamped'\newline \textcolor{black!55}{negative:}\ `objects can be listed twice' -- no appearance named, nothing tied\\
\bottomrule
\end{tabularx}
\end{table*}
\begin{table*}[t]
\centering
{\footnotesize Table~\ref{tab:rubrics}, continued.\par}\smallskip
\scriptsize
\setlength{\tabcolsep}{4pt}
\renewcommand{\arraystretch}{1.1}
\begin{tabularx}{\textwidth}{@{}l>{\raggedright\arraybackslash\hsize=0.80\hsize}X>{\raggedright\arraybackslash\hsize=0.60\hsize}X>{\raggedright\arraybackslash\hsize=1.60\hsize}X@{}}
\toprule
Criterion & Statement & Grounding & Worked examples\\
\midrule
CO1 & service order follows a FIXED CYCLE, explicitly NOT arrival order and NOT priority & cites the observed waiting orders across ticks that ruled arrival order out -- an order that changed with no release, or a later arrival served first & \textcolor{black!55}{positive:}\ `slate waiting went [p, q, r] then [q, r, p] with no release in between, and r arrived last but was served second -- it is a rotation, not first-come and not rank'\newline \textcolor{black!55}{negative:}\ `service order follows a fixed cycle, which is neither first-come nor priority' -- the claim stated flat, with no observed order behind it\\
\addlinespace[2.5pt]
CO2 & re-stamping a consignment mid-pass MOVES it in the waiting set, changing when it is released & gives the slate readings either side of the stamp & \textcolor{black!55}{positive:}\ `stamped consignment-p at tick 9; slate waiting went [p, q] -> [q, p] and q went out first'\newline \textcolor{black!55}{negative:}\ `stamping affects the queue' -- no reading either side\\
\addlinespace[2.5pt]
CO3 & the TOTAL waiting is conserved across a re-labelling -- membership moves, the count does not & gives the before and after counts, and names the act that moved the membership & \textcolor{black!55}{positive:}\ `three waiting before the stamp and three after; only the order and which line each sat on changed'\newline \textcolor{black!55}{negative:}\ `the total is conserved when things are relabelled' -- no counts, no act\\
\arrayrulecolor{black!25}\midrule[0.3pt]\arrayrulecolor{black}
CA1 & the audited count does NOT match the number of bodies standing -- so some bodies are not counted separately & gives both figures from the same moment: the counter reading and the listing it was compared against & \textcolor{black!55}{positive:}\ `counter read 9 while the roster listed 17 bodies at the same tick -- fewer counted things than visible ones'\newline \textcolor{black!55}{negative:}\ `the count seems to disagree with what is standing there' -- neither figure\\
\addlinespace[2.5pt]
CA2 & each intervention moves the counter by a FIXED step (typically one), with a table of act to change & gives the bracketed counter readings the step was read off & \textcolor{black!55}{positive:}\ `stamp: 12 -> 13. cull: 13 -> 12. retool: 14 -> 14. One each way, and retool does not touch it'\newline \textcolor{black!55}{negative:}\ `acts move the counter' -- no step, no table, no bracket\\
\addlinespace[2.5pt]
CA3 & a NUMERIC PREDICTION for a later reading, stated BEFORE that moment arrives & names the figure predicted, the surface it will be read from, and when & \textcolor{black!55}{positive:}\ `ledger says 6 acts booked this pass, so I expect the counter to read 15 at the stall'\newline \textcolor{black!55}{negative:}\ `the counter will probably go up' -- no number, nothing checkable\\
\arrayrulecolor{black!25}\midrule[0.3pt]\arrayrulecolor{black}
XC1 & one law is expected to hold in a SECOND, DIFFERENT domain -- and names the act and the surface that would test it there & names the act, the surface and the expected direction before the probe is run & \textcolor{black!55}{positive:}\ `if the counter step is one per act everywhere, then stamping a consignment should move the audit counter by one as well -- I will stamp consignment-p and read the counter either side'\newline \textcolor{black!55}{negative:}\ `the same law probably holds elsewhere' -- no act, no surface, no expectation\\
\addlinespace[2.5pt]
XC2 & a mapping from NAMED acts to which hidden quantity each one moves -- an act-to-effect table covering three or more acts & gives the bracketed readings each row of the map was read off & \textcolor{black!55}{positive:}\ `stamp moves the first hidden slot, retool the second, cull the third; the counter brackets for each are 12->13, 13->13, 13->12'\newline \textcolor{black!55}{negative:}\ `each verb does something different' -- no verb, no effect, no reading\\
\addlinespace[2.5pt]
XC3 & the counter is a GLOBAL instrument: it counts individuals, not bodies, so it partitions everything standing in the drift into groups & gives the repeated probes across more than one pass that the claim rests on & \textcolor{black!55}{positive:}\ `the counter moved by one for each of my acts in passes 2 and 3 regardless of which body I touched -- it is counting the underlying things, not the bodies'\newline \textcolor{black!55}{negative:}\ `the counter is global' -- one probe, or none\\
\arrayrulecolor{black!25}\midrule[0.3pt]\arrayrulecolor{black}
RC1 & DISTINCT VISIBLE OBJECTS CAN BE ONE INDIVIDUAL -- one state, one turn -- across renderings and across locations & names the specific objects held to be one, and the readings that identified them & \textcolor{black!55}{positive:}\ `feeder-2 and the registry line are one thing: they moved together in three separate brackets and the counter only ever moved once for the pair'\newline \textcolor{black!55}{negative:}\ `objects whose labels are equal are one individual -- one state, one turn' -- the axiom stated in the abstract, with no object named and nothing observed\\
\addlinespace[2.5pt]
RC2 & posits a HIDDEN multi-part label, and the individual as the owner of one act per tick and of one queue & names the parts postulated and the observations that forced them & \textcolor{black!55}{positive:}\ `there must be a hidden four-part tag: three verbs each change exactly one part, which is why two bodies can agree on some readings and not others'\newline \textcolor{black!55}{negative:}\ `there is probably something hidden behind the objects' -- nothing postulated with structure, nothing forced by an observation\\
\addlinespace[2.5pt]
RC3 & replaces PER-OBJECT RATES with the hidden label parts and the rotation position as what the law is about & says what the old per-object variable failed to predict and what the new one does & \textcolor{black!55}{positive:}\ `per-printer mark rate predicted nothing across passes; once I read it as (which parts the tag holds, where in the rotation it sits) the same numbers fall out every pass'\newline \textcolor{black!55}{negative:}\ `a well-formed law is about label coordinates and counts of acts, not per-object rates' -- the replacement asserted, with no old variable and no new prediction\\
\bottomrule
\end{tabularx}
\end{table*}

%% file: sections/app_g_outerwilds.tex
\clearpage
\section{Coda: what \textit{Outer Wilds} asks of its players}
\label{app:outerwilds}

\providecommand{\owpix}[1]{\IfFileExists{figures/ow/#1.pdf}{%
  \raisebox{-0.85ex}{\includegraphics[height=3ex]{figures/ow/#1.pdf}}\,}{}}

\noindent\textit{Spoiler: this appendix discusses the puzzles and the ending of Outer Wilds.}

\subsection{The game}
\label{app:outerwilds-game}

\textit{Outer Wilds}, developed by Mobius Digital and published by Annapurna Interactive in
2019 \citep{mobius2019outerwilds}, places the player in a small solar system whose sun goes
supernova twenty-two minutes after the player wakes beside a campfire on the home planet.
Then the same twenty-two minutes begin again. The player has a wooden spaceship, a
spacesuit with a jetpack, a signalscope that picks up signals from across the system, a
launcher for a camera probe called the scout, and a translator for the writing of the
Nomai, an extinct species whose ruins cover the planets. Each loop restores every planet and
every object. The ship's log carries over: it records what the player has found, sorted by
place in one view and linked into a web of rumours in the other. Game facts below follow
the in-game texts as the community wiki documents them \citep{fandom2026outerwilds}. The
game began as a master's thesis on curiosity-driven exploration
\citep{usc2013outerwilds}, and Eurogamer and Polygon each named it their game of the year
for 2019 \citep{eurogamer2019outerwilds,polygon2019outerwilds}.

We read the game as a worked instance of the two properties of this paper
(Appendices~\ref{app:outerwilds-interconnected} and~\ref{app:outerwilds-open}), and its
players' accounts as a record of what those properties ask of a person
(Appendix~\ref{app:outerwilds-hard}).
Table~\ref{tab:outerwilds} sets the game's demands beside those of \benchname{}, and a critic's
account of its progression could describe one of our episodes: ``When you start a new loop, you
don't bring anything from the previous one with you except knowledge, much of which is
conveniently tracked in your spaceship's computer'' \citep{kotaku2019outerwilds}.

\begin{table*}[t]
\centering
\caption{The discovery demands placed on players in \textit{Outer Wilds} and on agents in
\benchname{}. Each row gives an element of the game, what a player must work out to use it,
the property it exercises, and the corresponding element of \benchname{}.}
\label{tab:outerwilds}
\footnotesize
\setlength{\tabcolsep}{4pt}
\renewcommand{\arraystretch}{1.3}
\begin{tabularx}{\textwidth}{@{}>{\raggedright\arraybackslash}p{3.2cm}
  >{\raggedright\arraybackslash}X
  >{\raggedright\arraybackslash}p{2.9cm}
  >{\raggedright\arraybackslash}p{4.4cm}@{}}
\toprule
Element of \textit{Outer Wilds} & What the player must work out & Property & In \benchname{} \\
\midrule
\owpix{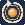}\owpix{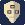}The loop
  & Where and when to go, since the planets change on a schedule and each loop restores them
  & Persistent exploration under a reset
  & \pix{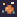}The pass: twenty-four ticks, after which the world resets \\
\owpix{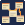}The ship's log
  & What the recorded facts imply together; the log links clues and records what was read
  & Persistent exploration under a reset; epistemically open
  & \pix{notes}The notebook, written by the agent and kept across passes \\
\owpix{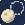}The quantum rules
  & That three rules taught on Giant's Deep, Ember Twin and Brittle Hollow compose one
    procedure at the shrine
  & Structurally interconnected
  & Hyperedges whose sources sit in three domains, as H1 over \dom{fen}{Fen},
    \dom{trellis}{Trellis} and \dom{ladder}{Ladder} \\
\owpix{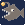}\owpix{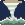}\owpix{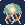}Anglerfish, cyclone, jellyfish
  & That a fact recorded on one planet opens a way on another
  & Structurally interconnected
  & Readings at one site that bear on another domain, as in the cross-domain bridge blocks
    XC1 to XC3 \\
\owpix{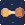}The Ash Twin towers
  & That a tower aligns with the centre of its astral body, a rule written in a forge on
    Brittle Hollow
  & Epistemically open
  & \pix{open}Representation changes RC1 to RC3, each licensed by joint evidence \\
\owpix{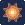}\owpix{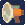}\owpix{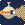}Dark Bramble
  & That duplicated signals map the seeds, and that the camera probe can serve as a beacon
  & Epistemically open
  & \pix{instrument}Instruments that report in their own words; what a reading is evidence
    for is the agent's inference \\
\owpix{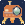}The ship
  & Flight itself, a skill of its own
  & Embodiment, set aside by \benchname{}
  & \pix{goto}One positioning verb: any site is one goto away \\
\owpix{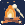}The ending
  & That the warp core, the Vessel and the coordinates of the Eye compose within one loop
  & All three
  & \pix{stamp}The standing order: one committed prediction of three readings \\
\bottomrule
\end{tabularx}
\end{table*}

\subsection{Clues that interlock across planets}
\label{app:outerwilds-interconnected}

The game is structurally interconnected by design. The thesis that began it ties each
central mystery to clues on several worlds: ``Each Curiosity is linked to three
Points-of-Interests, each of which is located on a different planet''
\citep{usc2013outerwilds}. The released game keeps this shape. The Quantum Moon wanders
between five orbits in the system and has a sixth location beside the Eye of the universe,
the object whose signal brought the Nomai to the system. A shrine on the moon takes the
player there. Its inscriptions ask the visitor to recall three rules, the last of them,
``Recall the rule of the sixth location'', beside a mural of a tower hanging above a black
hole. The rule of quantum imaging is taught in a tower on Giant's Deep, the rule of
quantum entanglement by a wandering shard in the caves of Ember Twin, and the rule of the
sixth location in a tower hanging from the crust of Brittle Hollow. Each rule answers a
local curiosity. Held together, the three form a procedure.

Smaller facts travel between planets in the same way. The anglerfish of Dark Bramble hunt
by sound, and the player learns that they are blind on Ember Twin. A rare cyclone that pushes
objects below the ocean current of Giant's Deep is explained beside working models of cyclones
on Brittle Hollow, and the insulation of the jellyfish that carry the player through the
electrified barrier around the core of Giant's Deep is recorded in a traveller's notes in Dark
Bramble.

The ending composes the whole system. The player removes the advanced warp core that powers
the loop from a Nomai facility inside Ash Twin, carries it to the Vessel on which the Nomai
arrived, now lost deep inside Dark Bramble, and enters the coordinates of the Eye that a
probe tracking module has recorded at the core of Giant's Deep. Each of the three goals is
reached through knowledge gathered on other planets. Polygon's citation for its game of the
year describes the experience plainly: ``It's only by piecing together the game's mechanics,
the Nomai's findings, and the Hearthians' current observations that I'm able to truly grasp
what is going on'' \citep{polygon2019outerwilds}.

\subsection{A representation the player builds}
\label{app:outerwilds-open}

The game is also epistemically open. It supplies facts in the words of whoever recorded
them and leaves their bearing to the player. The thesis states the principle: ``One
approach that seems to successfully encourage curiosity is to populate the world with
things that reveal affordances elsewhere in the world without telling players what to do''
\citep{usc2013outerwilds}. The ship's log follows it. In the writer's words its entries are
``only what we're 100\% certain the player knows from reading that piece of found text'',
with ``NO leaps of logic, no inferences'' \citep{gdc2021outerwilds}. The log links clues,
and every conclusion stays with the player.

Two cases show the representation work most clearly. Each warp tower on Ash Twin carries the
player to another body when that body passes overhead. One tower leads into the facility at
Ash Twin's own centre, and its receiver lies underground, so a player who takes a tower's
alignment point to be its receiver expects this tower to stay idle. A Nomai conversation in a
forge on Brittle Hollow records exactly that belief and its correction: a tower aligns with
the centre of its astral body. The twin planets share one centre, and the tower fires when
Ember Twin passes overhead. The way in opens once the player writes the rule over a different
variable, a revision of an existing concept that the game's own text prompts, and so a milder
demand than the Open conditions make.

Inside Dark Bramble, space folds. Glowing seeds open into chambers larger than the seeds
themselves, and a signal reaches the player from each of them. The signalscope, introduced in
the home village for a game of hide and seek, becomes the player's map of this topology. The
Nomai of one escape pod heard two identical beacons from the Vessel, followed the nearer, and
died beside a seed too small for any of them to enter, from which the beacon still sounded.
The scout fits. Fired through that seed, its signal reappears at an unlit seed in the
anglerfish nest, and that seed leads to the Vessel. The camera probe becomes a beacon.

\subsection{Progress stored in the player}
\label{app:outerwilds-reset}

The loop makes exploration persistent and keeps all of its progress in the player. The
game's writer takes up the question whether a first-time player could finish in a single
loop, and answers that it is possible in principle and extremely unlikely, since it would
take some advance knowledge of the game \citep{gdc2021outerwilds}. Removing the warp core
switches the loop off, so the last run is one life in which a death ends the game. The
coordinates of the Eye stay the same in every loop, so a player who has read them once can
enter them, and everything that run needs is already in the player.

\subsection{What players find hard}
\label{app:outerwilds-hard}

Players find the game demanding at the places this paper measures. One critic was
``especially stumped on a quantum conundrum involving a giant vanishing shrine, in which it
had to appear in a very specific place'' \citep{rps2019outerwilds}, the sixth-location
procedure above. Kotaku's verdict listed among its likes ``The unique satisfaction of
discovering clues either on purpose or by accident, getting lost'' and among its dislikes
``Controlling the janky spaceship, the confusing portal on Ash Twin, getting lost''
\citep{kotaku2019outerwilds}. Getting lost appears on both lists, and the confusing portal
is the warp tower above. Flight is a skill of its own: with the first version of the game's
opening, nearly every playtester flew into the sun shortly after take-off
\citep{usc2013outerwilds}. \benchname{} sets this embodied difficulty aside, since any site
is one \pix{goto}goto away (Section~\ref{sec:world}), and keeps the epistemic difficulty
whole.

Players also guard these discoveries for one another. Reviews stay free of spoilers so that
readers can find the discoveries on their own \citep{polygon2019outerwilds,kotaku2019outerwilds},
and the game's writer placed warnings in her own conference talk before its account of the
ending \citep{gdc2021outerwilds}. A received answer and a constructed one are different
possessions.

\subsection{What the game suggests for benchmark design}
\label{app:outerwilds-design}

In our reading, \textit{Outer Wilds} is epistemically open and structurally interconnected:
the player constructs the representation of each mechanism from clues scattered over several
planets, and the clues of each mystery jointly constrain one mechanism. Its players and
critics find it demanding and unforgettable, and the places that stump them are the places
where the two properties operate, the procedure the shrine composes from three planets and the
tower whose rule must be rewritten. We take this as the design's motivation and as a sign of
what a world with the two properties asks of a person. \benchname{} contains the two properties as
measurable objects, hyperedges whose sources sit in three domains and \pix{instrument}instruments
that report in their own words, beside the pass and its reset and a \pix{notes}notebook that
survives every reset. It also separates what the game presents together. The factorial prices
the construction of a representation and the composition of evidence one at a time, and the
standing order asks for the whole account at once, as the game's ending does.

The epigraph of this paper comes from the game. Solanum's two tenets, to seek out and to
understand, name what the ending asks of a player and what \benchname{} asks of an agent in one
committed prediction.